\documentclass{article} 
\usepackage{iclr2026_conference,times}

\usepackage{amsmath,amsfonts,bm}

\def\eqref#1{equation~\ref{#1}}

\def\1{\bm{1}}

\DeclareMathAlphabet{\mathsfit}{\encodingdefault}{\sfdefault}{m}{sl}
\SetMathAlphabet{\mathsfit}{bold}{\encodingdefault}{\sfdefault}{bx}{n}

\usepackage[pagebackref=false,breaklinks=true,letterpaper=true,colorlinks,citecolor=citecolor,bookmarks=false]{hyperref}
\usepackage{url}
\usepackage{graphicx}
\usepackage{booktabs}
\usepackage{amsmath}
\usepackage{subcaption}
\usepackage{amssymb}
\usepackage{multirow}
\usepackage{adjustbox}
\usepackage{colortbl}
\usepackage{makecell}
\usepackage{tabularx}
\usepackage{array}
\usepackage{pifont}
\usepackage{float}
\usepackage{xspace}
\usepackage{dsfont}
\usepackage{cleveref}
\usepackage{microtype}
\definecolor{citecolor}{HTML}{0071bc}

\definecolor{closed_models}{RGB}{240, 240, 240}
\definecolor{open_vlm}{RGB}{200, 255, 200}
\definecolor{open_models_gui_below_8B}{RGB}{185, 235, 255}
\definecolor{open_models_gui_over_8B}{RGB}{255, 219, 187}

\title{
  \raisebox{-.65ex}{\includegraphics[width=0.78cm]{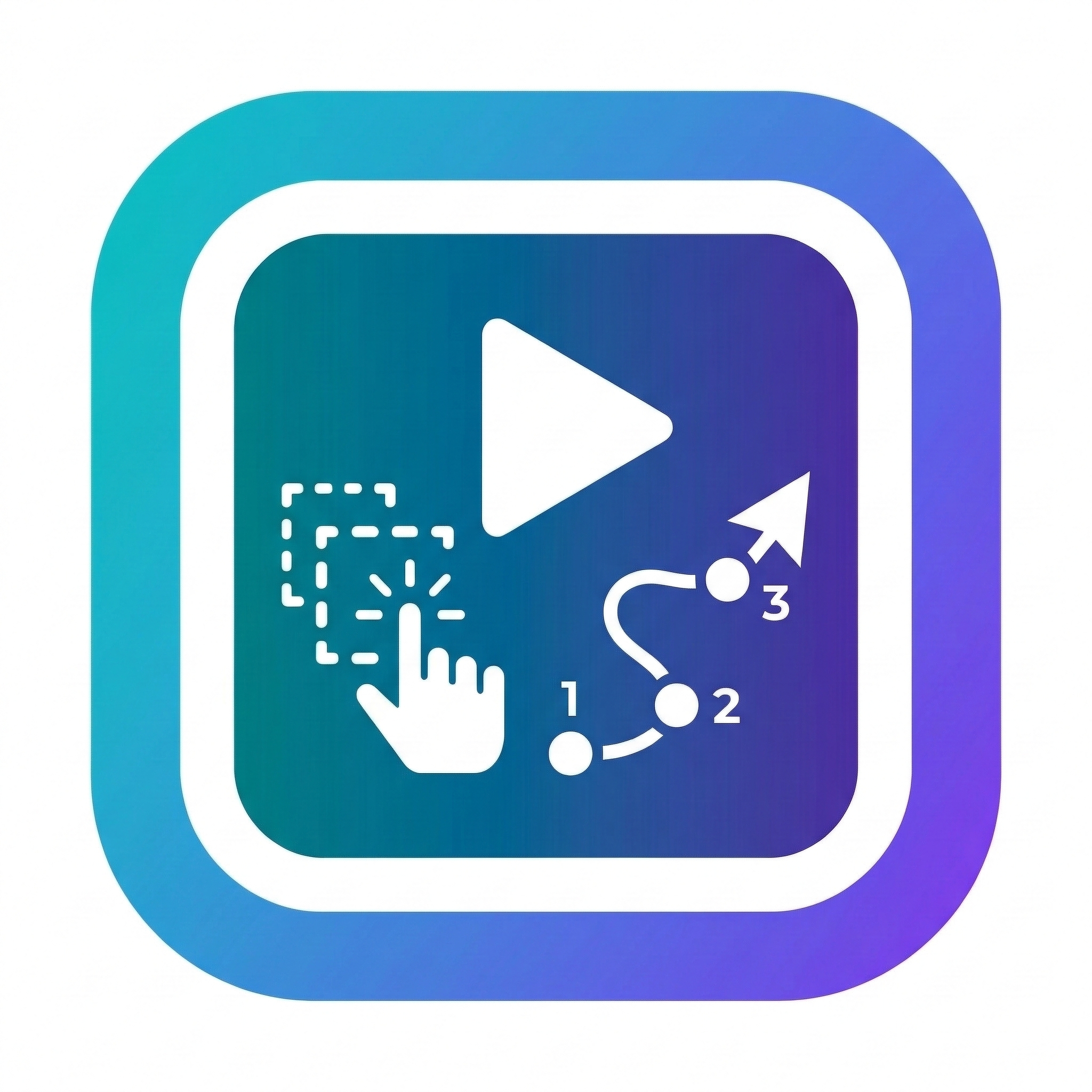}}\,%
  \textls[-5]{CUA-Suite: Massive Human-annotated Video Demonstrations for Computer-Use Agents}
}

\newcommand{\Comma}{\textsuperscript{,}}
\newcommand{\ServiceNow}{\textsuperscript{1}}
\newcommand{\UWaterloo}{\textsuperscript{2}}
\newcommand{\Mila}{\textsuperscript{3}}
\newcommand{\UMontreal}{\textsuperscript{4}}
\newcommand{\McGill}{\textsuperscript{5}}
\newcommand{\Oxford}{\textsuperscript{6}}
\newcommand{\NUS}{\textsuperscript{7}}

\newcommand{\FirstAuthor}{{\normalfont\textsuperscript{*}}}

\newcommand{\AuthorSep}{\hspace{1em}}

\author{
    \parbox{\textwidth}{
        \vspace{1.5ex}
        \centering
        Xiangru Jian\UWaterloo\Comma\FirstAuthor\AuthorSep
        Shravan Nayak\ServiceNow\Comma\Mila\Comma\UMontreal\Comma\FirstAuthor\AuthorSep
        Kevin Qinghong Lin\Oxford\AuthorSep
        Aarash Feizi\Mila\Comma\McGill\AuthorSep 
        Kaixin Li\NUS\AuthorSep 
        Patrice Bechard\ServiceNow\AuthorSep
        Spandana Gella\ServiceNow\Comma\McGill\AuthorSep
        Sai Rajeswar\ServiceNow\Comma\Mila
        \\[1.5ex]
        {
            \scriptsize\normalfont
            \ServiceNow ServiceNow\AuthorSep
            \UWaterloo University of Waterloo\AuthorSep
            \Mila Mila\AuthorSep
            \UMontreal Université de Montréal\AuthorSep
            \McGill McGill University\AuthorSep
            \Oxford University of Oxford\AuthorSep
            \NUS National University of Singapore \AuthorSep
            \FirstAuthor Equal Contribution
        }
    }
}

\newcommand{\dataset}{\textsc{CUA-Suite}}
\newcommand{\uivision}{\textsc{UI-Vision}}

\newcommand{\groundcua}{\textsc{GroundCUA}}
\newcommand{\actcua}{\textsc{VideoCUA}}

\iclrfinalcopy 
\begin{document}

\maketitle

\vspace{-25pt}
\begin{center}
    \raisebox{-0.3\height}{\includegraphics[width=1.25em,height=1.25em]{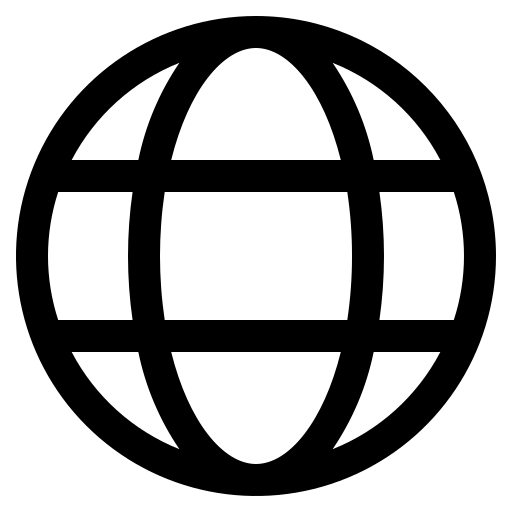}} \hspace{0.1em}
    \textbf{Project Page:} \href{https://cua-suite.github.io/}{\textcolor{blue}{\texttt{https://cua-suite.github.io}}}
\end{center}

\begin{abstract}
Computer-use agents (CUAs) hold great promise for automating complex desktop workflows, yet progress toward general-purpose agents is bottlenecked by the scarcity of continuous, high-quality human demonstration videos. Recent work emphasizes that continuous video, not sparse screenshots, is the critical missing ingredient for scaling these agents \citep{fdm1}. However, the largest existing open dataset, ScaleCUA \citep{liu2026scalecua}, contains only 2 million screenshots, equating to less than 20 hours of video. To address this bottleneck, we introduce \dataset{}, a large-scale ecosystem of expert video demonstrations and dense annotations for professional desktop computer-use agents. At its core is \actcua{}, which provides approximately 10,000 human-demonstrated tasks across 87 diverse applications with continuous 30\,fps screen recordings, kinematic cursor traces, and multi-layered reasoning annotations averaging 497 words per step, totaling approximately 55 hours and 6 million frames of expert video, more than 2.5$\times$ the largest existing open dataset. Unlike sparse datasets that capture only final click coordinates, these continuous video streams preserve the full temporal dynamics of human interaction, forming a superset of information that can be losslessly transformed into the formats required by existing agent frameworks. \dataset{} further provides two complementary resources: \uivision{} \citep{nayak2025uivisiondesktopcentricguibenchmark}, a rigorous benchmark for evaluating grounding and planning capabilities in CUAs, and \groundcua{} \citep{feizi2025groundingcomputeruseagents}, a large-scale grounding dataset with 56K annotated screenshots and over 3.6 million UI element annotations. Together, these resources provide dense, causal supervision in which every element on screen is labeled and every action is logged. Preliminary evaluation reveals that current foundation action models struggle substantially with professional desktop applications ($\approx$ 60\% task failure rate). Beyond benchmarking, \dataset{}'s rich multimodal corpus supports emerging research directions including generalist screen parsing, continuous spatial control, video-based reward modeling, and visual world models. All data and models are publicly released.
\end{abstract}

\section{Introduction}

\begin{figure}[t]
 \centering
 \vspace{-15pt}
 \includegraphics[width=1.0\linewidth]{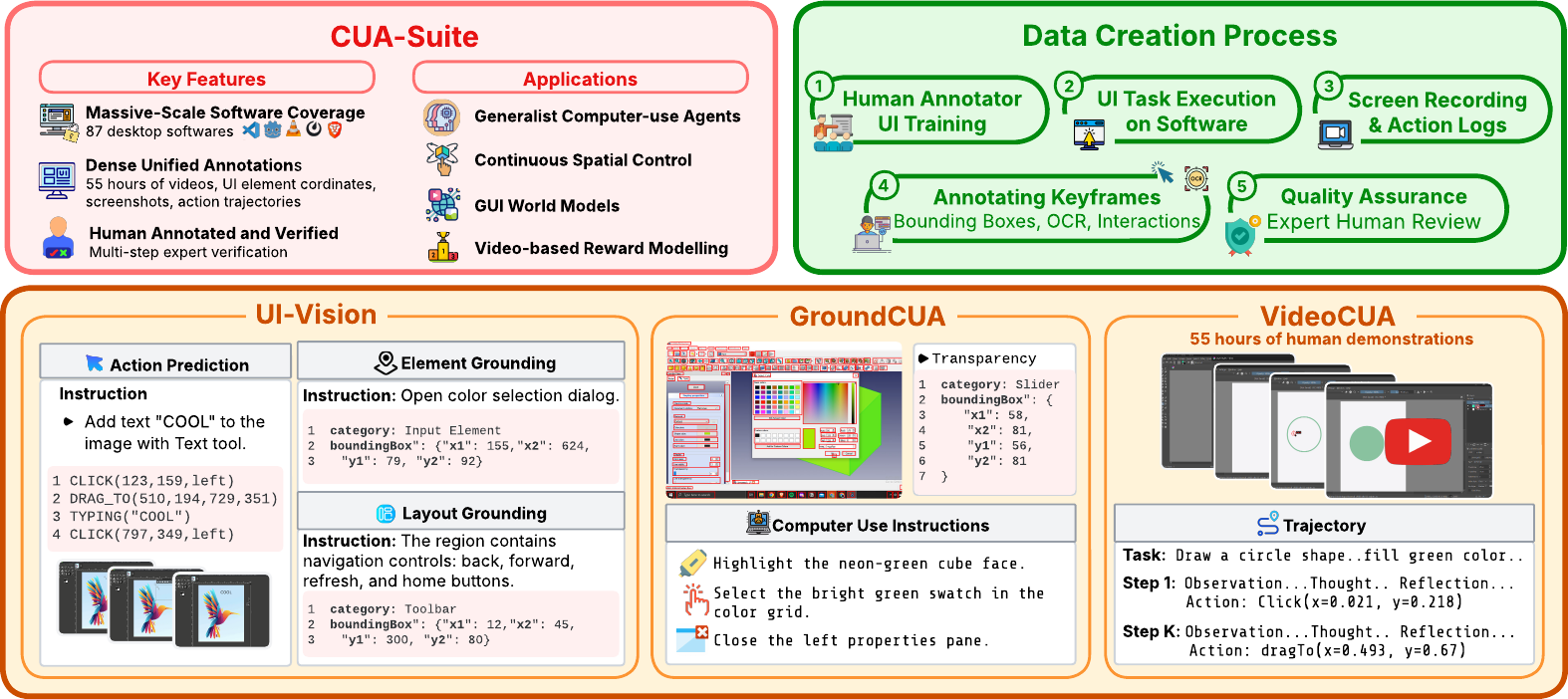}
 \caption{\textbf{\dataset{} Overview.} Human GUI trajectories are recorded across desktop platforms, expert-verified, and annotated with keyframes, bounding boxes, and interaction logs. The resulting \dataset{} comprises \uivision{}, a comprehensive benchmark to evaluate grounding and planning capabilities of computer-use agents; \groundcua{}, which consists of densely labeled UI screenshots with 3.6 million annotations; \actcua{}, which comprises 55 hours of video of humans performing computer-use tasks with detailed action trajectories.}
\label{fig:teaser} \vspace{-10pt}
\end{figure}

The vision of intelligent agents that can operate alongside humans at a computer, understand our goals, navigate our interfaces, and execute complex workflows on our behalf, has long captured the imagination of researchers and practitioners alike \citep{workarena2024,osworld,pang2025,zhang2025large,nguyen-etal-2025-gui}. Computer-use agents (CUAs) promise to achieve this vision and transform how we work: from automating repetitive data entry to orchestrating sophisticated 3D modeling pipelines, from streamlining scientific analyses to managing the ever-growing complexity of our day-to-day digital lives. In an era where digital literacy is a bottleneck \citep{wef2023future, nsc2025digitaldivide} and interface complexity grows exponentially \citep{buntcomplexity, MM-gi2000}, these agents offer a compelling vision: computers that transition from passive tools to active collaborators.

However, realizing this vision has proven difficult. Despite significant advances in vision-language models and foundation agents, today's CUAs remain surprisingly brittle \citep{nayak2025uivisiondesktopcentricguibenchmark,screenspot-pro}. They excel at simple web tasks but falter when using professional desktop applications, such as 3D modeling software, IDEs, and specialized tools that underpin modern knowledge work. This problem is exacerbated for popular but non-mainstream applications, such as open-source software, where models struggle to navigate unfamiliar interfaces \citep{nayak2025uivisiondesktopcentricguibenchmark}.

For CUAs to be truly useful, they must take a user's task, formulate a plan, and ground that plan in executable actions. The fundamental challenge is the lack of high-quality training data that encompasses rich, dense annotations for both planning and grounding. Recent works have attempted to fill this gap with automatically curated or synthesized datasets \citep{ariaui, wu2024osatlasfoundationactionmodel}, but these often suffer from noise inherent to automated generation. When human-curated datasets do exist, they typically cover only partial aspects of the problem, such as spatial grounding without temporal context \citep{seeclick,gou2024uground}. Moreover, even comprehensive human datasets such as OpenCUA \citep{OpenCUA2025} rely on action discretization, resulting in sparse screenshots that omit intermediate visual feedback between actions. Concurrent work has independently confirmed this limitation, arguing that screenshot-based agents are fundamentally unable to process high-framerate video, perform long-horizon tasks, or scale to competent agents \citep{fdm1}. For instance, ScaleCUA \citep{liu2026scalecua}, the largest existing open dataset, contains 2 million screenshots, which equates to less than 20 hours of video at 30\,fps. Such sparse data lacks the temporal continuity required to build visual world models or learn the continuous spatial control policies necessary for human-like cursor movement \citep{luo2025vimogenerativevisualgui, koh2026generativevisualcodemobile}. Bridging this gap requires richly annotated human data providing dense, multi-faceted feedback: continuous video trajectories, kinematic action traces, and precise UI grounding. Together, these signals enable models to capture the full spectrum of computer-use intelligence. 

This paper introduces \dataset{}, a large-scale ecosystem of expert video demonstrations and dense annotations that addresses the full stack of computer-use intelligence. At its core is \actcua{}, which provides approximately 55 hours and 6 million frames of full, uncut 30\,fps video recordings of human experts performing over 10,000 tasks across 87 professional desktop applications. Unlike sparse datasets that capture only final click coordinates, these continuous video streams preserve the full temporal dynamics of human interaction (Figure~\ref{fig:teaser}). Each video is further enriched with kinematic cursor traces and multi-layered reasoning annotations. The continuous video format enables future fine-tuning experiments comparing video-based and screenshot-based training signals. \dataset{} further provides two complementary resources: \uivision{} \citep{nayak2025uivisiondesktopcentricguibenchmark} and \groundcua{} \citep{feizi2025groundingcomputeruseagents}. \uivision{} is a rigorous benchmark for evaluating grounding and planning, specifically designed to expose model failures in diverse software applications. To address these challenges, \groundcua{} provides pixel-precise, human-curated annotations of UI elements across 87 applications, directly targeting the spatial grounding bottleneck.

By unifying all three resources (\actcua{}, \groundcua{}, and \uivision{}), \dataset{} provides dense, causal supervision, in which every element on screen is labeled and every action is logged. Table~\ref{tab:dataset_comparison} provides a systematic comparison of \actcua{} against existing datasets, highlighting its unique position at the intersection of continuous video, desktop focus, human curation, and rich reasoning annotations. This rich signal enables the training of foundation action models grounded in human-verified truth, and unlocks the potential to build visual world models for lookahead planning and continuous spatial control policies. In summary, our key contributions are:
\begin{itemize}
    \item \textbf{\actcua}: The largest open expert video corpus for desktop computer use, comprising approximately 55 hours and 6 million frames of 30\,fps recordings across 10,000 tasks and 87 applications, with kinematic cursor traces and multi-layered reasoning annotations.
    \item \textbf{The \dataset{} Framework}: The unification of continuous expert video demonstrations (\actcua) with pixel-precise \textit{grounding} (\groundcua) and rigorous evaluation (\uivision) into a single, comprehensive ecosystem for full-stack computer-use intelligence.
    \item \textbf{Fully open-source release}: We open-source all benchmarks, training data, and models to accelerate research in computer-use agents.
\end{itemize}

\section{Related Work}
Our work is situated at the intersection of visual grounding, agentic action prediction, and trajectory-based learning. While recent years have seen a proliferation of datasets for CUAs, a gap remains in resources that bridge high-fidelity video streams with dense, verified element-level supervision.

\noindent \textbf{GUI Visual Grounding Datasets.} Visual grounding is a prerequisite for reliable computer use. Most grounding datasets target mobile and web environments, leveraging standardized representations such as Android's View Hierarchy~\citep{deka2017rico,uibert,amex} or the HTML DOM~\citep{seeclick,webui}, with UGround~\citep{gou2024uground} scaling to 10M elements across 1.3M screenshots. However, these methods rely on accessibility trees that are often noisy or incomplete~\citep{muryn2025screen2axvisionbasedapproachautomatic} and fail to capture the pixel-level complexity of desktop applications. Desktop grounding remains considerably harder: OS-ATLAS~\citep{wu2024osatlasfoundationactionmodel} and JEDI~\citep{xie2025scaling} attempt to scale supervision through accessibility-tree traversal and synthetic interface generation, respectively, but automated methods often yield misaligned bounding boxes. Benchmarks such as ScreenSpot-Pro~\citep{screenspot-pro}, WinSpot~\citep{hui2025winclickguigroundingmultimodal}, and VenusBench-GD~\citep{zhou2025venusbenchgdcomprehensivemultiplatformgui} have exposed the severity of this gap, yet they cover only narrow slices of the desktop ecosystem and rely on semi-automated pipelines that limit their use as training data.

\noindent \textbf{Action Prediction and Agent Benchmarks.} Beyond static grounding, agents must reason about task logic and predict sequential actions. \textbf{Execution-Based Benchmarks.} Significant progress has been made in evaluating agents via execution feedback. MiniWoB++~\citep{miniwob++} and WebArena~\citep{webarena} serve as standard testbeds for web agents, while AndroidWorld~\citep{androidworld} and AITW~\citep{aitw} evaluate mobile agents on multi-step tasks. In the desktop domain, OSWorld~\citep{osworld} and Windows Agent Arena~\citep{bonatti2024windows} provide interactive environments for evaluating open-ended tasks. While these benchmarks excel at providing execution scores, they often lack the dense, offline supervision required to train Vision-Language-Action Models (VLAMs) from scratch, often relying on sparse reward signals. \textbf{Agent Architectures.} GUI agents have evolved from early visual encoders~\citep{pixel2act,cogagent} to reasoning-integrated architectures such as UI-TARS~\citep{uitars2025}, InfiGUI~\citep{liu2025infigui}, TongUI~\citep{zhang2025tonguiinternetscaletrajectoriesmultimodal}, and ScaleCUA~\citep{liu2026scalecua}, yet these remain trained on static screenshot-action pairs, limiting their understanding of temporal dynamics.

\noindent \textbf{Video-Centric and Trajectory Learning.} The emergence of models capable of processing long-context video has shifted focus toward learning from continuous observation rather than discrete states. \textbf{Learning from Observation.} Video data provides a rich temporal context that static screenshots miss. VideoGUI~\citep{videogui} utilizes instructional videos to benchmark GUI automation, while OmniACT~\citep{omniact} explores multimodal generalization. Recently, OpenCUA~\citep{OpenCUA2025} and Agent S~\citep{agashe2024agentsopenagentic} have highlighted the importance of diverse trajectory data for training generalist computer-use agents. To address the data bottleneck, several concurrent efforts propose scalable trajectory synthesis: GUICourse~\citep{guicourse} contributes a suite of datasets spanning 10M page-annotation pairs and over 80K navigation instructions for end-to-end agent training, AgentTrek~\citep{xu2025agenttrek} generates trajectories by mining and replaying web tutorials in real environments, and OS-Genesis~\citep{sun2024osgenesis} introduces reverse task synthesis where agents first explore GUI environments and retrospectively derive tasks from observed interactions. Despite these efforts, a critical limitation of existing video datasets is the granularity of their annotations. While datasets like VideoGUI provide high-level task descriptions, they lack frame-level grounding that links actions to specific UI elements. Table~\ref{tab:dataset_comparison} summarizes the landscape: no existing dataset simultaneously provides continuous 30\,fps video, desktop coverage, human-curated trajectories, and rich multi-layered reasoning annotations at scale.

\section{\dataset{}\,\raisebox{-2ex}{\includegraphics[width=1cm]{icon/cua-suite_v3.png}}\,}
\label{sec:method}

\subsection{Curation of \dataset{}}
\label{sec:dataset}
We introduce \dataset{}, a large-scale ecosystem of continuous expert video demonstrations and dense UI annotations for professional desktop applications. Where previous efforts often rely on synthetic accessibility trees for Desktop environments~\citep{ariaui, osworld}, recaption existing annotations~\citep{showui} or focus exclusively on web browsers~\citep{gou2024uground}, our approach centers on recording high-fidelity human behavior as continuous 30\,fps video. We prioritize professional-grade applications and dense, manual annotation to create a dataset for training agents on real-world workflows. This unified data engine underlies the \dataset{} ecosystem, supporting three complementary datasets: \actcua{} for training agents on complex workflow execution through continuous video trajectories, \groundcua{} for training agents on UI grounding, and \uivision{} for benchmarking visual perception and planning. Below, we describe our data collection pipeline, from application selection to dense annotation collection.

\noindent \textbf{Selecting Diverse Applications.}
To support general-purpose computer-use agents, we selected $87$ open-source applications across $12$ categories (\Cref{tab:platforms}). These applications range from software development (VS Code) and content creation (Blender, Inkscape, Krita) to finance and productivity (GnuCash, LibreOffice). By focusing on open-source applications with permissive licenses, we ensure the dataset can be freely released while encompassing a wide range of domains. These applications mirror the functionality of popular closed-source software (e.g., LibreOffice vs. Microsoft Office), making the dataset broadly applicable. Further details are provided in \Cref{app:platforms}.

\noindent \textbf{Expert-Driven Task Design.}
Instead of procedurally generating goals or using templates, we asked human experts to design tasks they would perform in a real work setting. The tasks range from simple actions (e.g., renaming a folder, creating a document) to complex workflows (e.g., editing a spreadsheet, running a simulation, applying subtitles to a video). We ensure that each task is well-defined and comprehensive. This approach ensures the collected trajectories represent coherent, goal-oriented behavior rather than random exploration. In total, annotators completed over $10,000$ task demonstrations across $87$ applications.

\noindent \textbf{Recording High-Fidelity Video Demonstrations.}
Annotators executed these tasks while our system captured continuous screen video at 30 frames per second, producing approximately 55 hours and 6 million frames of uncut expert demonstration footage across all tasks. Alongside the video stream, we logged every mouse click, drag, scroll, and keystroke with millisecond precision, yielding synchronized kinematic cursor traces. By preserving the complete visual state at every frame, the dataset encodes the full temporal dynamics of expert desktop interaction, including the intermediate cursor movements and visual feedback between actions that sparse screenshot-based datasets discard.

\noindent \textbf{Dense UI Annotation.}
From this continuous visual stream, we extract specific keyframes, i.e., snapshots of the interface immediately preceding state-changing user actions (e.g., clicks or text entry) to serve as the basis for grounding. This selection ensures that annotations correspond to the user's decision-making context. Annotators then manually label every visible UI element in these keyframes with bounding boxes. For each element, they provide a textual label. This label is the element's name when available, the displayed text for shorter strings, or a concise summary in the case of long passages such as source code or detailed descriptions. We also extract OCR using PaddleOCR \citep{cui2025paddleocrvlboostingmultilingualdocument} to extract raw text specifically for these longer segments. Additionally, approximately 50\% of elements are classified into one of eight high-level functional categories (see \Cref{tab:ui_categories}), adding a layer of semantic structure to the geometric ground truth.

\noindent \textbf{A Unified Foundation.}
This robust data engine serves as the single source of truth for the entire \dataset{}. The collected data is methodically processed to construct three complementary resources: \actcua{} for complex agentic execution through continuous video trajectories, \groundcua{} for fine-grained UI grounding, and \uivision{} for visual perception and planning evaluation. By grounding these complementary resources in a shared foundation of expert human behavior, \dataset{} provides a holistic platform for diagnosing and advancing the capabilities of computer-use agents. We envision this rich, multimodal corpus will serve as a catalyst for future research, supporting tasks beyond our current scope (see \Cref{sec:more_utility}) and enabling the community to build the next generation of generalist computer-use agents.


\subsection{\uivision\,\raisebox{-1.5ex}{\includegraphics[width=0.65cm]{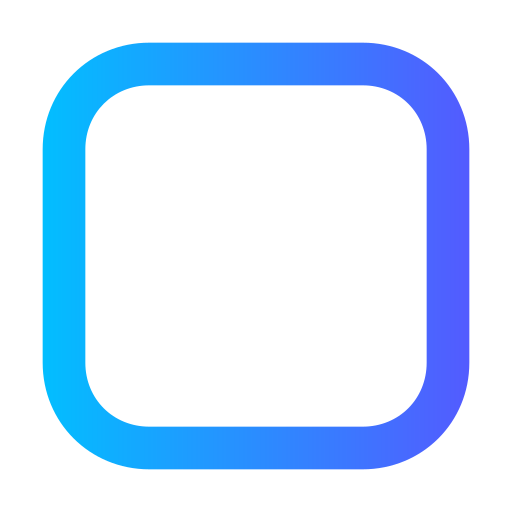}}\,}
\label{sec:uivision}


\begin{table}[t]
\centering
\small
\vspace{-30pt}
\caption{Element Grounding Performance on \uivision{}. Models are ranked in descending order by their average score. Results are drawn from their respective original publications, with the exception of Qwen3-VL, which is reported as evaluated in \citet{zhou2025maiuitechnicalreportrealworld}.}
\label{tab:model_performance}
\begin{tabular}{l c c c c}
\toprule
\textbf{Model} & \textbf{Basic} & \textbf{Functional} & \textbf{Spatial} & \textbf{Avg.} \\
\midrule
MAI-UI-32B \citep{zhou2025maiuitechnicalreportrealworld} & \textbf{59.1} & \textbf{57.1} & \textbf{26.9} & \textbf{47.7} \\
MAI-UI-8B \citep{zhou2025maiuitechnicalreportrealworld} & 51.7 & 49.6 & 22.5 & 41.3 \\
OpenCUA-72B \citep{OpenCUA2025} & -- & -- & -- & 37.3 \\
UI-Venus-Ground-72B \citep{gu2025uivenustechnicalreportbuilding} & 45.6 & 42.3 & 23.7 & 37.2 \\
PhiGround-7B + o3 \citep{zhang2025phi} & 44.2 & 43.8 & 20.5 & 36.2 \\
OpenCUA-32B \citep{OpenCUA2025} & -- & -- & -- & 33.3 \\
GUI-ARP-7B \citep{ye2025guiarpenhancinggroundingadaptive} & 39.6 & 35.4 & 18.6 & 31.2 \\
OpenCUA-7B \citep{OpenCUA2025} & -- & -- & -- & 29.7 \\
Qwen3-VL-32B \citep{bai2025qwen3vltechnicalreport} & 32.8 & 34.2 & 14.7 & 27.2 \\
PhiGround-7B \citep{zhang2025phi} & 36.8 & 37.1 & 7.6 & 27.2 \\
UI-Venus-Ground-7B \citep{gu2025uivenustechnicalreportbuilding} & 36.1 & 32.8 & 11.9 & 26.9 \\
InfiGUI-G1-7B \citep{liu2025infiguig1advancingguigrounding} & 36.2 & 31.9 & 11.5 & 26.5 \\
HyperClick \citep{zhang2025hyperclickadvancingreliablegui} & 35.3 & 32.1 & 11.0 & 26.1 \\
UI-TARS-72B \citep{uitars2025} & 31.4 & 30.5 & 14.7 & 25.5 \\
Qwen3-VL-8B \citep{bai2025qwen3vltechnicalreport} & 25.0 & 27.9 & 1.2 & 18.0 \\
UI-TARS-7B \citep{uitars2025} & 20.1 & 24.3 & 8.4 & 17.6 \\
\bottomrule
\end{tabular} \vspace{-1em}
\end{table}

\uivision{}~\citep{nayak2025uivisiondesktopcentricguibenchmark} is a desktop-centric benchmark for evaluating the visual perception and planning capabilities of computer-use agents. It comprises 450 high-quality task demonstrations originating from the \dataset{} spanning diverse applications and interaction patterns, and serves as the primary evaluation benchmark within the \dataset{} ecosystem.

The benchmark is designed to specifically test three fundamental agentic capabilities. \textbf{Element Grounding} evaluates the agent's ability to precisely localize UI elements given a textual query (e.g., "Click the Save button"), assessing the foundational visual understanding required to translate semantic intent into screen coordinates. \textbf{Layout Grounding} tests the agent's comprehension of the interface structure by requiring it to identify and group functionally related elements (e.g., "Select the navigation bar"), going beyond individual element recognition to evaluate holistic scene understanding. Finally, \textbf{Action Prediction} assesses the agent's planning capability by providing a high-level goal and the current screen state, and asking it to predict the next correct action (e.g., click, drag, type), connecting visual perception to executable decision-making. By leveraging the dense annotations and expert trajectories from \dataset{}, \uivision{} provides a multi-faceted diagnosis of where agents fail, i.e., whether in seeing the interface, understanding its structure, or planning the next move. We refer readers to \citet{nayak2025uivisiondesktopcentricguibenchmark} for a detailed discussion on benchmark creation and metric definitions.

\noindent \textbf{Results and Discussion.} 
Previous evaluations on \uivision{} identified visual grounding as the primary bottleneck limiting agent performance~\citep{nayak2025uivisiondesktopcentricguibenchmark}. Consequently, the analysis here focuses on re-evaluating the grounding capabilities of state-of-the-art multimodal models to assess recent progress and persistent challenges. (A detailed analysis of the Action Prediction task is provided in \Cref{sec:actcua}). The primary observation from \Cref{tab:model_performance} is that overall performance has nearly doubled in the year since the introduction of \uivision{}; the previous state-of-the-art, UI-TARS-72B, has been significantly outperformed by newer architectures, with MAI-UI-32B achieving a new high of 47.7\% in average accuracy. Despite this rapid progress, substantial scope for improvement remains. Breaking down the results by task type reveals that while models are excelling in the Basic and Functional categories, with top models approaching 60\% accuracy, the Spatial split remains stubbornly difficult across the board, indicating that reasoning about spatial relationships on the screen is still a major hurdle. This could be attributed to a lack of such training data in existing computer-use datasets. The results also demonstrate that scaling up model parameters yields consistent and predictable benefits for grounding. For instance, the OpenCUA family shows a clear upward trajectory, improving by 7.6 absolute points as capacity increases from 7B to 72B parameters. This scaling is likely due to improvements in reasoning capabilities, which help distinguish dense, visually similar UI regions. This is further evident in PhiGround-7B, where pairing the model with an o3 planner improves its average performance by an impressive 9.0 absolute points (from 27.2\% to 36.2\%), demonstrating that providing models with reasoned instructions mitigates execution errors.
\subsection{\groundcua\,\raisebox{-1.5ex}{\includegraphics[width=0.65cm]{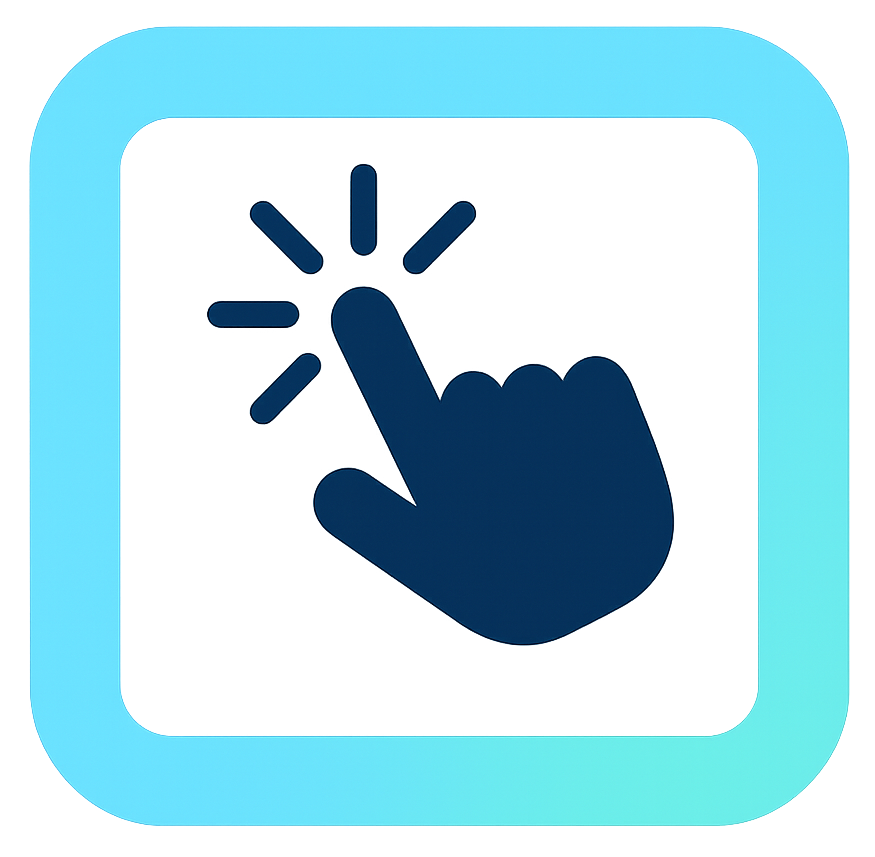}}\,}
\label{sec:groundcua}

To address the visual grounding bottleneck highlighted by evaluations on \uivision{}, \groundcua{}~\citep{feizi2025groundingcomputeruseagents} serves as a massive, high-quality training corpus for fine-grained UI understanding. \groundcua{} is built upon dense UI annotations sourced from \dataset{}. Key highlights of \groundcua{} compared to other datasets include its \textbf{Scale}, with 56K annotated screenshots and over 3.6 million element annotations; \textbf{Resolution and Density}, featuring high-resolution images with maximum annotation density that covers almost every visible element, including small icons and controls; \textbf{Expert Quality}, ensured by human-verified annotations for high accuracy and 50\% of elements categorized into fine-grained semantic types; and \textbf{Application Diversity}, spanning 87 diverse desktop applications for broad real-world coverage.

From these annotations, \groundcua{} yields a 700K instruction-tuning dataset designed to teach agents precise, application-specific grounding strategies. This data powers the training of \textbf{\textsc{GroundNext}}~\citep{feizi2025groundingcomputeruseagents}, a series of efficient vision-language models (3B and 7B). The models leverage a two-stage training recipe that includes supervised fine-tuning (SFT) followed by reinforcement learning (RL) to achieve state-of-the-art results on a variety of grounding datasets. Furthermore, \textsc{GroundNext-3B} achieves a score of 50.6 on OS-World Verified when paired with the o3 planner~\citep{o3}. This underscores its significant practical utility, positioning it as an effective and efficient solution for real-world agentic systems where inference speed and resource constraints are critical factors. Readers are referred to \citet{feizi2025groundingcomputeruseagents} for a detailed discussion of instruction-tuning dataset curation, \textsc{GroundNext} training, and comprehensive performance comparisons against alternative datasets.

\subsection{\actcua\,\raisebox{-1.5ex}{\includegraphics[width=0.65cm]{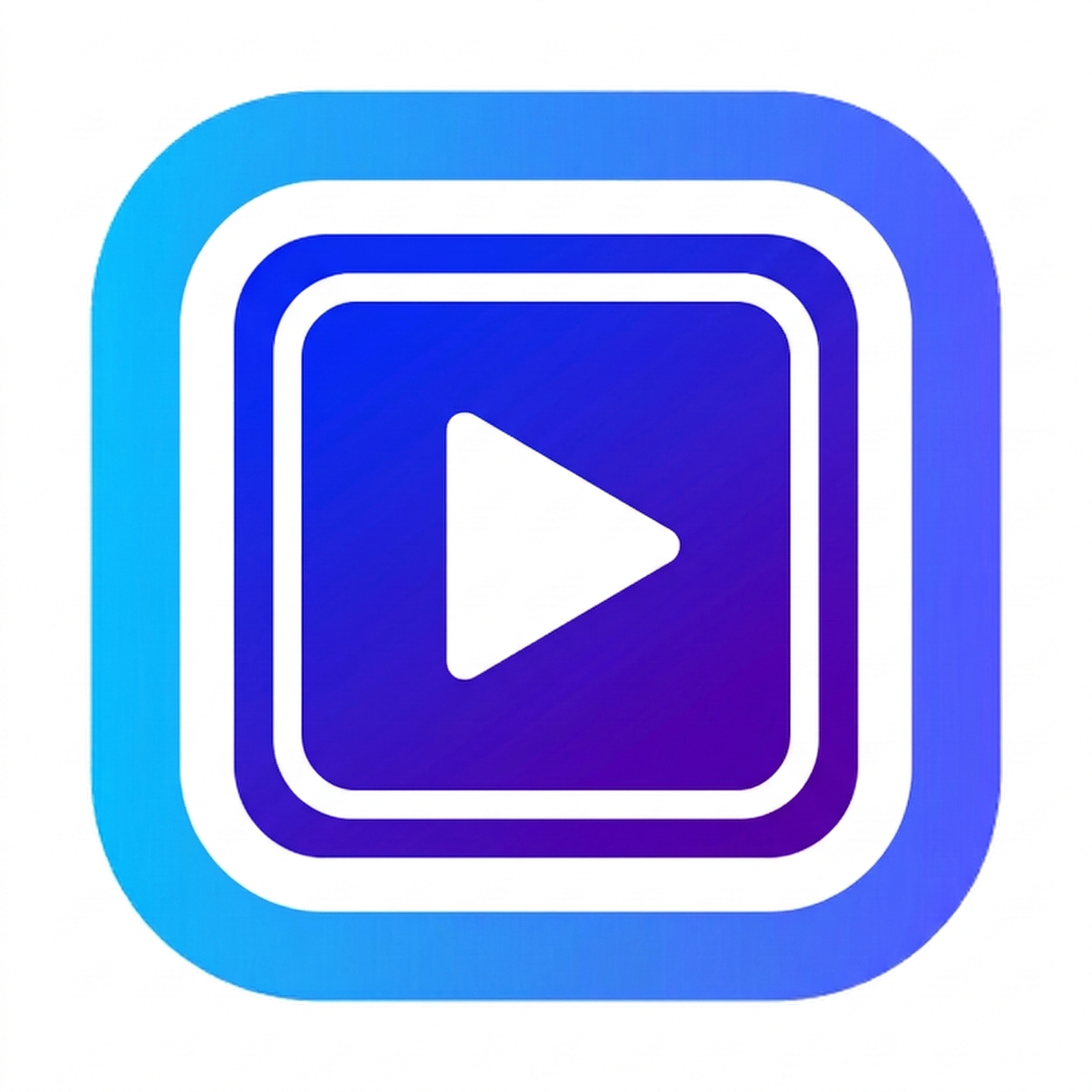}}\,}
\label{sec:actcua}

Frameworks such as OpenCUA~\citep{OpenCUA2025} have shown that scaling diverse interaction traces is essential for training generalist Computer-Use Agents (CUAs). However, progress is bottlenecked by the scarcity of high-quality human demonstrations in specialized domains, often necessitating reliance on synthetic data decomposition or noisy accessibility trees~\citep{xu2025agenttrek,sun2024osgenesis}. Existing large-scale trajectory datasets overwhelmingly target web or mobile platforms~\citep{mind2web,xu2025agenttrek,aitw,guiodyssey}, while desktop-oriented resources remain limited to small-scale evaluation benchmarks~\citep{osworld,videogui} rather than providing training-scale data for the diverse professional applications where agents struggle most (Table~\ref{tab:dataset_comparison}). \dataset{} directly addresses this bottleneck. Within \dataset{}, the video demonstration component is \textbf{\actcua{}}, comprising approximately 10,000 tasks across 87 professional desktop applications, totaling approximately 55 hours and 6 million frames of continuous 30\,fps expert video, more than 2.5$\times$ the approximately 2 million screenshots (${\sim}$18.5 hours at 30\,fps) in ScaleCUA~\citep{liu2026scalecua}, the largest existing open computer-use dataset~\citep{fdm1}. Table~\ref{tab:dataset_comparison} summarizes this comparison. \actcua{} is format-compatible with both OpenCUA~\citep{OpenCUA2025} and ScaleCUA~\citep{liu2026scalecua}, enabling seamless integration into their training pipelines as a high-quality expansion of desktop trajectory data. Beyond mere compatibility, we demonstrate the complete utility cycle: from \textit{annotation synthesis}, which enriches our raw video recordings with multi-layered reasoning annotations, to \textit{action prediction evaluation}, which reveals critical gaps in current foundation models when applied to professional desktop applications.

\begin{table}[t]
\centering
\vspace{-22pt}
\caption{\textbf{Comparison of \actcua{} with existing GUI trajectory and agent datasets.} \actcua{} is the only large-scale, human-curated dataset providing continuous 30\,fps expert video for professional desktop applications with long multi-layered CoT annotations. \textbf{Video}: continuous video recordings released with the dataset (not per-step screenshots). \textbf{CoT}: chain-of-thought annotations ($\ell$\,=\,long multi-layered, $s$\,=\,short). Though not released directly, \actcua{} has annotations dense enough for CoT generation with the code released. \textbf{Scale}: approximate data volume (SS\,=\,screenshots; h\,=\,video hours; fr.\,=\,video frames). OmniACT provides single-step tasks. OSWorld is an evaluation benchmark without training trajectories. ``--'' indicates not reported.}
\label{tab:dataset_comparison}
\vspace{0.5em}
\footnotesize
\renewcommand{\arraystretch}{1.2}
\setlength{\tabcolsep}{2pt}
\begin{tabular}{lcrrccccrr}
\toprule
\textbf{Dataset} & \textbf{Plat.} & \textbf{Tasks} & \textbf{\#Envs} & \textbf{Video} & \textbf{Desk.} & \textbf{Human} & \textbf{CoT} & \textbf{\#Act.} & \textbf{Scale} \\
\midrule
\multicolumn{10}{l}{\textit{Web}} \\
\rowcolor{closed_models!50}
Mind2Web~\citep{mind2web} & Web & 2,350 & 137 & \ding{55} & \ding{55} & \ding{51} & \ding{55} & 3 & ${\sim}$17K SS \\
\rowcolor{closed_models!50}
AgentTrek~\citep{xu2025agenttrek} & Web & 10,398 & 127 & \ding{55} & \ding{55} & \ding{55} & $s$ & 10 & ${\sim}$126K SS \\
\midrule
\multicolumn{10}{l}{\textit{Mobile}} \\
\rowcolor{open_vlm!50}
AITW~\citep{aitw} & Mob. & 715K & 357 & \ding{55} & \ding{55} & Mix. & \ding{55} & 4 & ${\sim}$4.6M SS \\
\rowcolor{open_vlm!50}
GUI-Odyssey~\citep{guiodyssey} & Mob. & 8,334 & 212 & \ding{55} & \ding{55} & Mix. & \ding{55} & 9 & ${\sim}$128K SS \\
\midrule
\multicolumn{10}{l}{\textit{Desktop \& Cross-platform}} \\
\rowcolor{open_models_gui_over_8B!50}
OmniACT~\citep{omniact} & D+W & 9,802 & 65 & \ding{55} & \ding{51} & \ding{51} & \ding{55} & 10 & ${\sim}$9.8K SS \\
\rowcolor{open_models_gui_over_8B!50}
OSWorld~\citep{osworld} & Desk. & 369 & 9 & \ding{55} & \ding{51} & \ding{51} & \ding{55} & {--} & Eval. \\
\rowcolor{open_models_gui_over_8B!50}
VideoGUI~\citep{videogui} & Desk. & 178 & 11 & \ding{51} & \ding{51} & Mix. & \ding{55} & 4 & ${\sim}$7h \\
\rowcolor{open_models_gui_over_8B!50}
GUI-World~\citep{guiworld} & Cross & 12,379 & {--} & \ding{51} & \ding{51} & Mix. & \ding{55} & {--} & ${\sim}$83K Frames \\
\rowcolor{open_models_gui_over_8B!50}
OpenCUA~\citep{OpenCUA2025} & Desk. & 22,625 & 140+ & \ding{55} & \ding{51} & \ding{51} & $\ell$ & 13 & ${\sim}$421K SS \\
\rowcolor{open_models_gui_over_8B!50}
ScaleCUA~\citep{liu2026scalecua} & Cross & ${\sim}$19K & {--} & \ding{55} & \ding{51} & Mix. & \ding{55} & {19} & ${\sim}$2M SS \\
\midrule
\rowcolor{open_models_gui_below_8B!50}
\textbf{\actcua{} (Ours)} & \textbf{Desk.} & \textbf{${\sim}$10K} & \textbf{87} & \textbf{\ding{51}} & \textbf{\ding{51}} & \textbf{\ding{51}} & $\boldsymbol{\ell}$ & \textbf{10} & \textbf{55h (6M fr.)} \\
\bottomrule
\end{tabular}\vspace{-0.5em}
\end{table}

\subsubsection{Multi-layered Reasoning Annotations}

A key challenge in leveraging video demonstration data for training action models is the \textit{semantic gap} between raw interaction logs and the structured reasoning required by modern vision-language agents. Raw recordings capture \textit{what} the user did (e.g., ``click at coordinates $(x, y)$''), but not \textit{why} they did it or \textit{what} visual elements they were targeting. To bridge this gap, we enrich each \actcua{} video with dense trajectory annotations that provide action-level intent descriptions for effective instruction following.

We address this by adopting the \textbf{Multi-layered reasoning trajectory synthesis pipeline} from OpenCUA~\citep{OpenCUA2025} that transforms our annotated recordings into richly described state-action sequences. For each keyframe $s_t$ in a task trajectory, we employ Claude-Sonnet-4.5 to generate four complementary annotation layers: (1) an \textbf{observation} providing detailed description of the current screen state, identifying relevant UI elements and their spatial arrangement; (2) a \textbf{thought} chain explaining the reasoning process connecting the high-level task goal to the immediate action choice; (3) an \textbf{action description} specifying the intended action in natural language grounded to visual elements rather than raw coordinates; and (4) a \textbf{reflection} analyzing the outcome and enabling self-correction signals for training. This synthesis process converts each trajectory step into the format $\tau_t = (s_t, o_t, r_t, d_t, a_t, s_{t+1}, \text{ref}_t)$, where $o_t$ is the observation, $r_t$ is the reasoning chain, $d_t$ is the action description, $a_t$ is the executable \texttt{pyautogui} code, and $\text{ref}_t$ is the reflection. This rich structure enables training on both \textit{action prediction} (predicting $a_t$ from $s_t$ and $d_t$) and \textit{reasoning generation} (producing $r_t$ given $s_t$ and the task instruction).

Our synthesized trajectories average \textbf{496.7 words per step} (157.4 observation, 194.3 reasoning, 17.7 action description, 127.4 reflection), providing the dense supervision signal that recent work identifies as critical for training capable GUI agents~\citep{xu2024aguvisunifiedpurevision,uitars2025}.
We provide complete annotated trajectory examples in Appendix~\ref{sec:actcua_examples}.

\subsubsection{Action Prediction Evaluation}
\label{sec:eval_protocol}

To understand how well current foundation action models generalize to complex desktop applications our \dataset{} can offer, we conduct a preliminary evaluation on \textbf{256 sampled tasks} spanning 87 applications. Rather than serving as a definitive benchmark, this evaluation is designed to reveal the landscape of model capabilities and limitations across diverse desktop domains.

\noindent \textbf{Trajectory Preparation.} Raw video recordings are converted into state-action trajectories compatible with the above-mentioned format. Each trajectory step contains a screenshot, the corresponding \texttt{pyautogui} action code, and (for CoT-enriched trajectories) the four annotation layers described above. All coordinates are stored in normalized form $(x, y) \in [0, 1]^2$ and converted to absolute pixels at the model's input resolution during inference. To mitigate information leakage from cursor position and to handle redundant \texttt{moveTo}$\to$\texttt{click} pairs, we apply keyframe extraction and action filtering detailed in Appendix~\ref{sec:eval_preprocessing}. After filtering, the evaluation comprises \textbf{1,999 coordinate-based predictions} across click, doubleClick, rightClick, and dragTo actions.

\noindent \textbf{Evaluation Setting.} We evaluate under a \textbf{task-level action prediction} setting that reflects realistic agent deployment: the model receives (1) the overall task instruction (e.g., ``Copy file `Presentation3.pptx' to a new folder and then ...''), (2) the last $k{=}5$ ground-truth actions as history context formatted with both natural language descriptions and \texttt{pyautogui} code, and (3) the current keyframe screenshot. The model must predict the next action \textit{without} receiving step-level instructions, i.e., it must infer what to do from the task goal and action history alone, which is the realistic case of how CUAs work in practice. 

\noindent \textbf{Evaluation Scope and Metrics.} Due to the difficulty of automatically verifying non-coordinate actions (e.g., keyboard shortcuts, text entry) whose correctness depends on application state rather than spatial proximity, the automated evaluation focuses exclusively on \textbf{coordinate-based actions}. For these, we report the mean and median pixel distance between predicted and ground-truth coordinates, as well as success rates at 20- and 50-pixel thresholds. Pixel distance provides a continuous measure of prediction quality, whereas threshold-based success rates indicate the proportion of predictions that lie within a practically useful proximity to the target. For a holistic assessment that \textbf{covers all action types}, we complement these automated metrics with a \textbf{human-evaluation study} (Section~\ref{sec:human_eval}) that includes non-coordinate actions and assesses the semantic correctness of coordinate predictions.

\subsubsection{Model Performance}

We evaluate OpenCUA-7B and OpenCUA-32B~\citep{OpenCUA2025}, the two publicly available foundation action models at different scales. Table~\ref{tab:large_scale_results} presents the results. Both models exhibit limited accuracy under task-level prediction. Across 256 sampled tasks (up to 1,999 predictions), the 32B model achieves \textbf{37.7\%} @50px, while the 7B model reaches only \textbf{16.5\%}. Scaling from 7B to 32B yields consistent improvement: +21.2 percentage points at @50px, with mean pixel distance decreasing from 387.5 to 274.2. However, absolute performance remains modest. Per-application performance varies widely (from 3.6\% to 73.3\% @50px) with specialized creative tools and canvas-based applications posing the greatest challenge, while web-like interfaces achieve higher accuracy (detailed analysis in Appendix~\ref{sec:app_level_analysis}, error distance distribution in Appendix~\ref{sec:error_distance_analysis}).

\begin{table}[t]
\centering
\vspace{-25pt}
\caption{\textbf{Action Prediction Results.} OpenCUA models evaluated on 256 sampled \actcua{} tasks (87 apps) under task-level instruction with 5-step action history context.}
\label{tab:large_scale_results}
\vspace{0.5em}
\small
\setlength{\tabcolsep}{5pt}
\begin{tabular}{lrcccc}
\toprule
\textbf{Model} & \textbf{Preds} & \textbf{Mean Px$\downarrow$} & \textbf{Med.\ Px$\downarrow$} & \textbf{@20px$\uparrow$} & \textbf{@50px$\uparrow$} \\
\midrule
OpenCUA-7B  & 1,946 & 387.5 & 236.0 & 7.9\% & 16.5\% \\
OpenCUA-32B & 1,999 & 274.2 & 97.0  & 22.0\% & 37.7\% \\
\bottomrule
\end{tabular}\vspace{-15pt}
\end{table}

\begin{figure}[t]
    \centering
    \vspace{-20pt}
    \begin{subfigure}[b]{0.48\textwidth}
        \includegraphics[width=\textwidth]{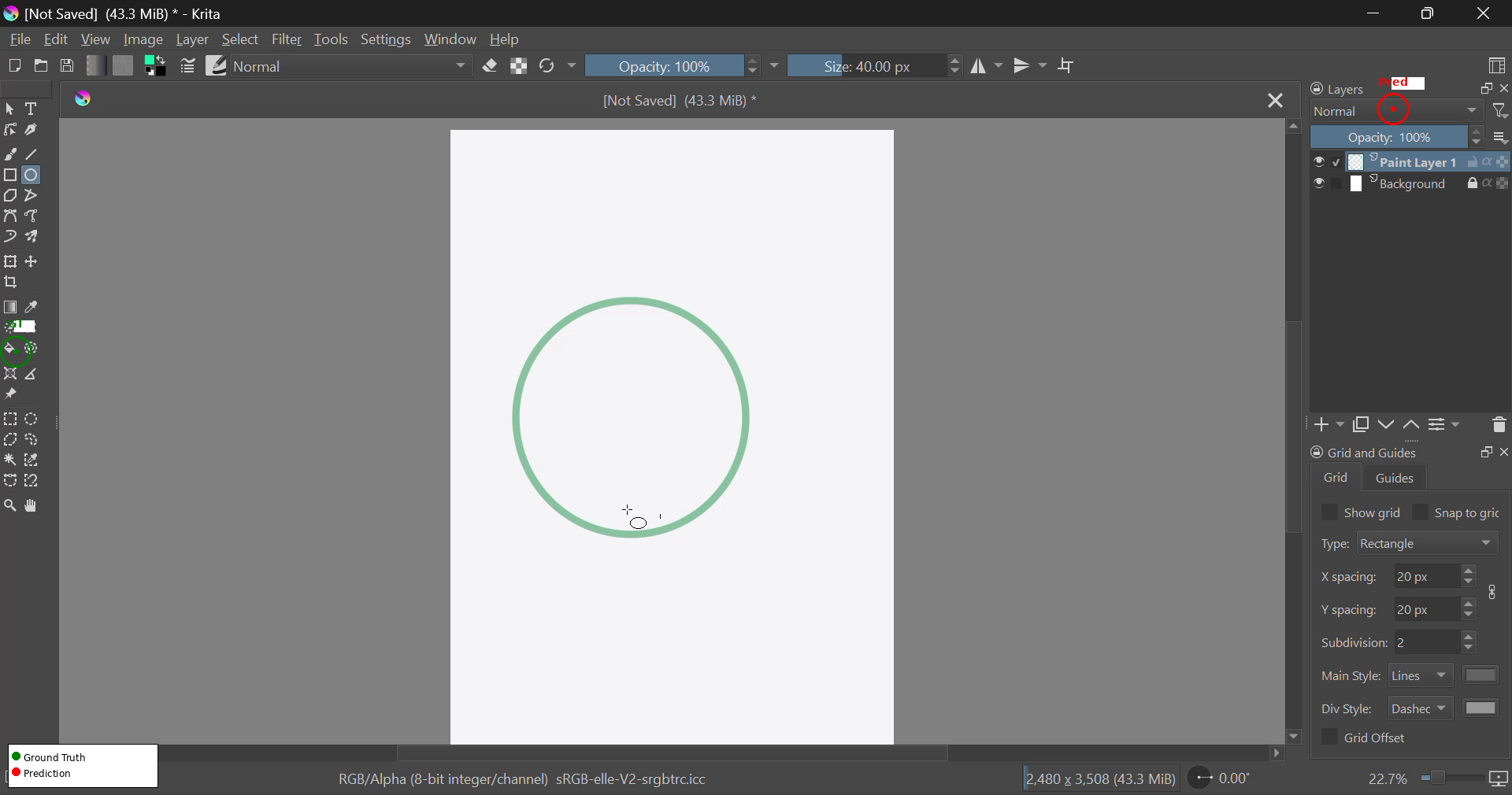}
        \caption{\textbf{Krita}: Cross-panel confusion.}
        \label{fig:fail_krita}
    \end{subfigure}
    \hfill
    \begin{subfigure}[b]{0.48\textwidth}
        \includegraphics[width=\textwidth]{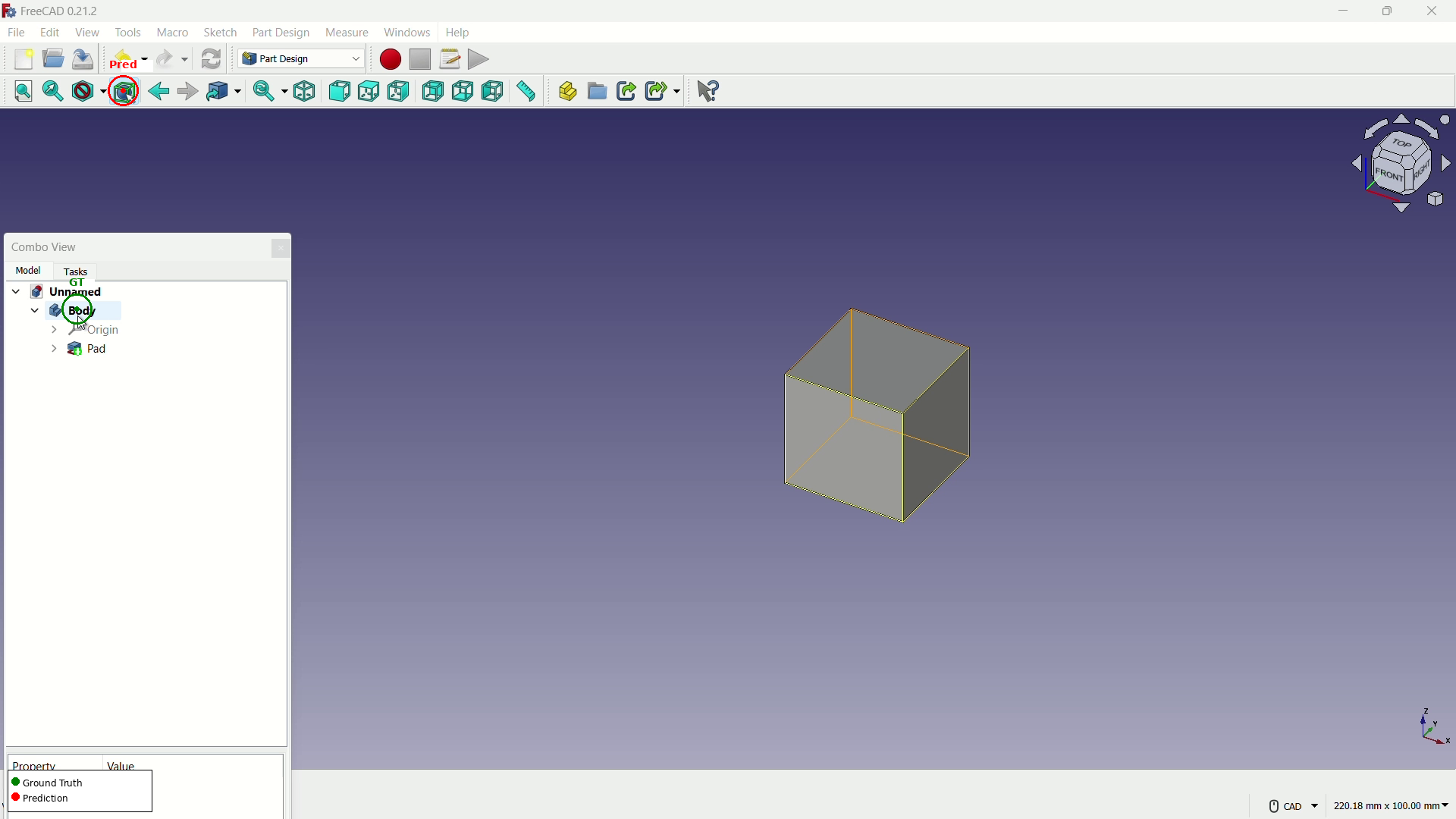}
        \caption{\textbf{FreeCAD}: Tree vs.\ toolbar.}
        \label{fig:fail_freecad}
    \end{subfigure}

    \vspace{0.5em}

    \begin{subfigure}[b]{0.48\textwidth}
        \includegraphics[width=\textwidth]{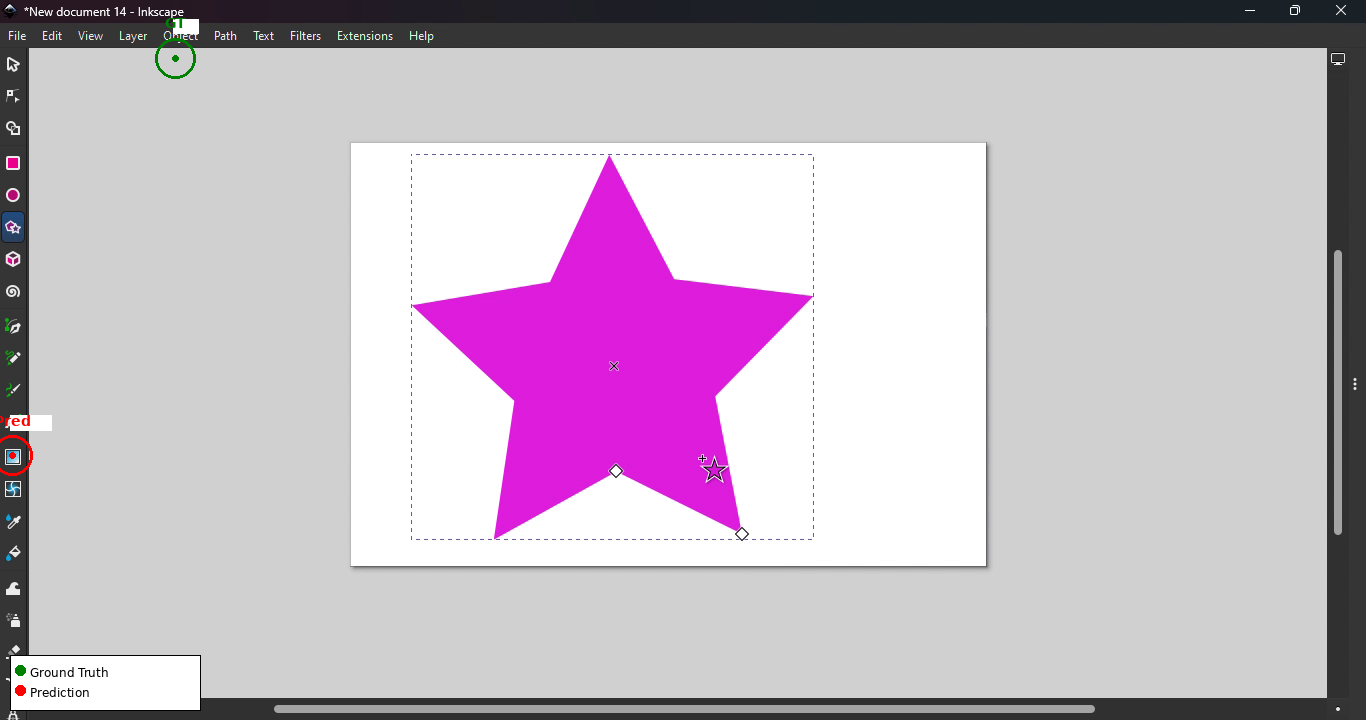}
        \caption{\textbf{Inkscape}: Menu vs.\ tool sidebar.}
        \label{fig:fail_inkscape}
    \end{subfigure}
    \hfill
    \begin{subfigure}[b]{0.48\textwidth}
        \includegraphics[width=\textwidth]{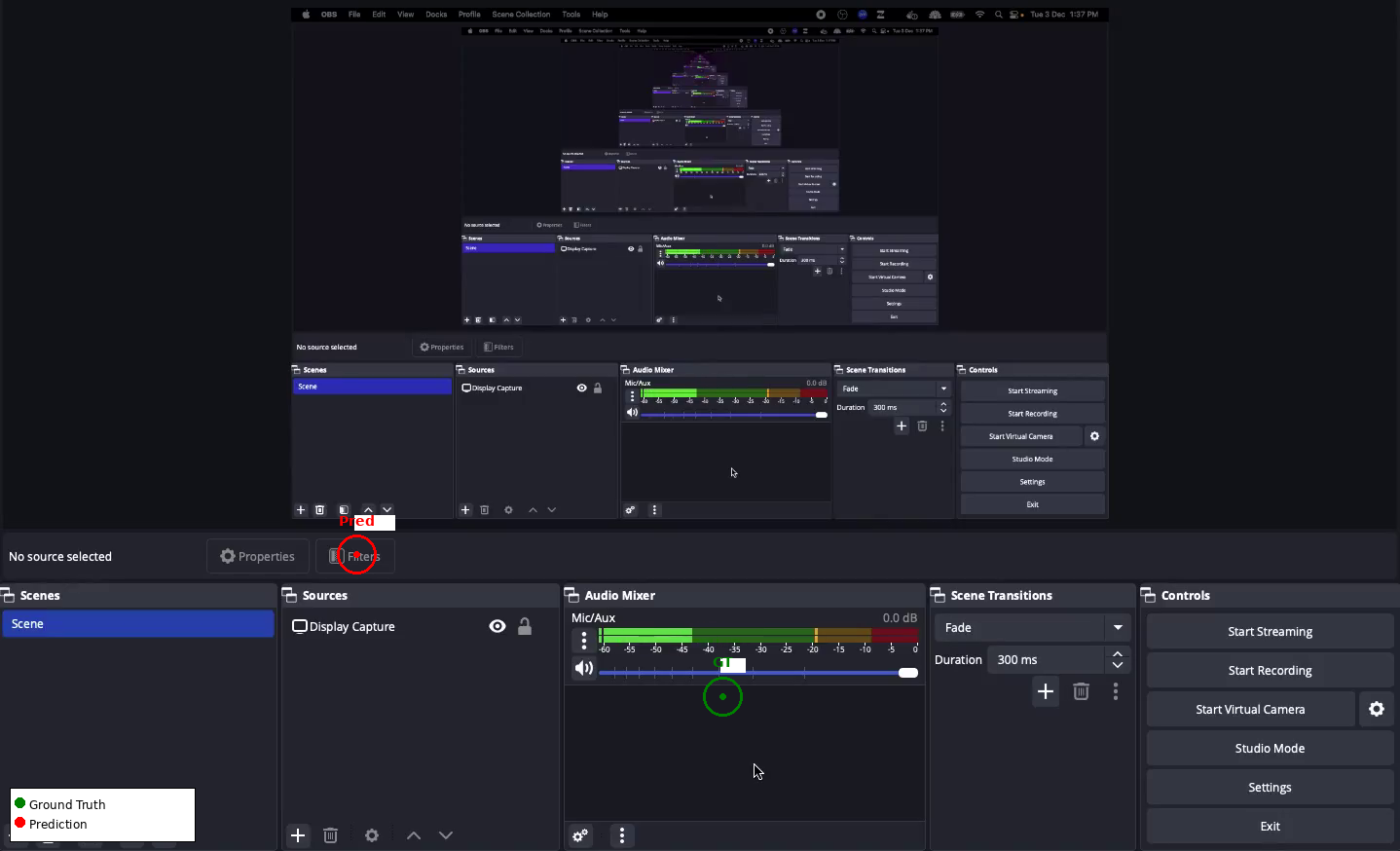}
        \caption{\textbf{OBS Studio}: Multi-panel layout.}
        \label{fig:fail_obs}
    \end{subfigure}

    \caption{\textbf{Representative Prediction Failures.} Green circles indicate ground truth and red circles show model predictions. These examples illustrate desktop-specific challenges for CUAs.}
    \label{fig:failure_cases} \vspace{-1.3em}
\end{figure}


\noindent \textbf{Qualitative Error Analysis.} Figure~\ref{fig:failure_cases} illustrates representative failure modes. In Krita, the model targets the Layers panel on the far right rather than a tool icon in the left sidebar, i.e., a cross-panel error spanning nearly the full screen width. In FreeCAD, the model clicks a toolbar icon rather than right-clicking the target node in the model tree, confusing two visually distinct but spatially proximate UI regions. In Inkscape, the model selects a sidebar tool instead of the intended menu bar item, and in OBS Studio, it clicks the wrong panel in a dense multi-panel layout. These failures share a common pattern: the model struggles to disambiguate among visually similar interactive elements distributed across the complex, multi-panel interfaces characteristic of professional desktop applications, which is precisely what our \actcua{} is able to offer.

\subsubsection{Human Evaluation of Predicted Trajectories}
\label{sec:human_eval}

The automated pixel distance metrics cannot capture whether a prediction constitutes a semantically correct action, as a prediction 30 pixels from the ground truth may still click the correct button, while one 10 pixels away may land on an adjacent element. To complement the distance-based evaluation, we conduct a human evaluation that directly assesses the functional correctness of the 32B model's predicted actions across all action types.

\noindent \textbf{Annotation Protocol.} We sample 49 tasks using stratified sampling by step count with diversity-based priority, yielding 576 annotated steps. For each predicted action, a human annotator judges: (1) \textbf{grounding correctness}: whether the predicted coordinates target the intended UI element (``yes'', ``no'', or ``n/a'' for non-coordinate actions); and (2) \textbf{action correctness}: whether the predicted action type and parameters match the intended interaction. A step is \textit{correct} only if both the action is correct and the grounding is correct (or not applicable). For multi-step compound actions (e.g., drag sequences, multi-keystroke text entry), annotators assess whether the consecutive steps collectively achieve the intended outcome.

\noindent \textbf{Results.} The 32B model achieves a combined stepwise accuracy of \textbf{57.6\%} (332/576). Decomposing this reveals a notable asymmetry: action correctness reaches \textbf{85.9\%} (495/576), while grounding accuracy among coordinate-based steps is only \textbf{52.4\%} (195/372). This gap indicates that the model frequently identifies the correct action type and intent but fails to localize the target UI element precisely. Non-coordinate steps (keyboard shortcuts, text entry) achieve higher accuracy at 67.6\% (138/204), consistent with these actions not requiring spatial grounding. Per-task accuracy ranges from 0\% to 100\% (mean 56.2\%), with 9 of 49 tasks achieving more than 80\% and 7 falling below 30\%, confirming that model reliability is highly application-dependent.

\subsubsection{Implications for Foundation Action Models}

Both automated and human evaluations converge on the same conclusion: current foundation action models struggle substantially with professional desktop applications, achieving only 37.7\% @50px and 57.6\% human-verified stepwise accuracy. The wide per-application variance confirms that the core difficulty lies in the diverse visual vocabularies and interaction patterns of professional desktop software, where existing training data is scarce. \textbf{\actcua{}} directly targets this domain gap through \textbf{domain coverage} of 87 professional applications where models struggle most, \textbf{video scale} of approximately 55 hours of continuous 30\,fps recordings that capture the full temporal dynamics of expert interaction, \textbf{annotation density} of $\sim$497 words per step providing rich supervisory signal for action planning, and \textbf{action diversity} spanning complex interaction primitives (drags, fine-grained mouse control) that web-centric datasets underrepresent.

\section{Universality and Application Potential of \dataset{}}
\label{sec:more_utility}

A primary design principle of \dataset{} is \textbf{universality}: by capturing continuous video streams, kinematic cursor trajectories, and expert-verified dense annotations, our data constitutes a \textit{superset} of the information required by current paradigms. Concretely, \dataset{} can be losslessly transformed into the specialized formats of every leading agent framework (such as screenshot-action pairs, state-action-next-state triplets, or continuous kinematic traces) without discarding any recorded signal. This generality means that as new training paradigms emerge, \dataset{} remains directly applicable without re-collection. We highlight four application frontiers below, with detailed descriptions in Appendix~\ref{sec:utility_details}. \textbf{(1) Generalist Screen Parsing.} Current MLLMs struggle to reliably parse desktop screenshots into structured UI elements~\citep{omniparser}. \dataset{} provides dense, human-verified bounding-box annotations covering all interactable elements, including canvas-based and custom-drawn widgets missed by DOM-based approaches, together with functional descriptions that enable semantics-aware element captioning. Recent works, such as ScreenParse~\citep{gurbuz2026movingsparsegroundingcomplete}, have already explored this possibility on a subset of data in GroundCUA. With denser, richer annotations in \dataset{}, we believe it can enable even more powerful and robust models for screen parsing. \textbf{(2) Continuous Spatial Control.} GUI-Cursor~\citep{zhao2025learningguigroundingspatial} argues that discrete coordinate prediction is insufficient for high-precision tasks, proposing continuous cursor movements from visual feedback instead. \dataset{} captures intermediate cursor movements and complete video context, preserving the kinematic priors (e.g., Fitts's Law deceleration) needed to train imitation learning or offline RL policies for feedback-driven navigation. \textbf{(3) Visual World Models.} Training GUI world models such as those based on textual planning~\citep{yourllm}, visual generation~\citep{luo2025vimogenerativevisualgui,neuralos,guan2026computerusingworldmodel} or code rendering~\citep{code2world} requires dense $(s_t, a_t, s_{t+1})$ triplets that most sparse-screenshot datasets cannot provide. Our 30fps video recordings, paired with timestamped actions, provide exactly these causal pairs, enabling action-conditioned video generation and visual lookahead planning for desktop workflows. \textbf{(4) Video-Based Reward Modeling.} Evaluating whether an agent has successfully completed a task remains a key bottleneck for scaling CUA training with reinforcement learning. ExeVRM~\citep{song2026exevrm} demonstrates that video-based reward models, which assess task completion from execution video rather than agent internals, can outperform proprietary models. \dataset{}'s continuous expert video recordings with task-level annotations provide an ideal source of positive demonstrations for training such reward models, while the dense step-level annotations enable fine-grained, step-wise reward signals beyond binary task-level labels.

\section{Conclusion}

We presented \dataset{}, a comprehensive ecosystem for training and evaluating desktop computer-use agents. At its core, \actcua{} provides the largest open expert video corpus for desktop computer use, comprising approximately 55 hours of continuous 30\,fps recordings across 10,000 tasks and 87 professional applications, with kinematic cursor traces and multi-layered reasoning annotations. This is complemented by pixel-precise grounding (\groundcua{}) and rigorous evaluation (\uivision{}) within a single coherent ecosystem. Our evaluations reveal that spatial grounding remains the primary bottleneck for current foundation action models in desktop environments, with performance varying widely across application domains. Beyond its immediate use as a benchmark and training corpus, \dataset{}'s continuous video streams, kinematic traces, and dense annotations support emerging paradigms such as generalist screen parsing, continuous spatial control, and visual world models. We will release all data, benchmarks, and models to the community, and hope \dataset{} will serve as a foundation for building the next generation of general-purpose computer-use agents.

\newpage

\bibliography{iclr2026_conference, related_datasets}

@techreport{wef2023future,
  title       = {The Future of Jobs Report 2025},
  author      = {{World Economic Forum}},
  year        = {2025},
  institution = {World Economic Forum},
  address     = {Geneva, Switzerland},
  url         = {https://reports.weforum.org/docs/WEF_Future_of_Jobs_Report_2025.pdf}
}

@techreport{nsc2025digitaldivide,
  title       = {Closing the Digital Divide in the United States},
  author      = {{National Skills Coalition}},
  year        = {2025},
  institution = {National Skills Coalition},
  type        = {Factsheet},
  url         = {https://nationalskillscoalition.org/wp-content/uploads/2025/02/NSC-National_factsheet_digdivide.pdf}
}

@inproceedings{buntcomplexity,
author = {Bunt, Andrea and Conati, Cristina and McGrenere, Joanna},
title = {Supporting interface customization using a mixed-initiative approach},
year = {2007},
isbn = {1595934812},
publisher = {Association for Computing Machinery},
address = {New York, NY, USA},
url = {https://doi.org/10.1145/1216295.1216317},
doi = {10.1145/1216295.1216317},
booktitle = {Proceedings of the 12th International Conference on Intelligent User Interfaces},
pages = {92–101},
numpages = {10},
location = {Honolulu, Hawaii, USA},
series = {IUI '07}
}

@inproceedings{MM-gi2000,
title = {Are We All in the Same "Bloat"?},
author = {Joanna McGrenere and Gale Moore},
booktitle = {Proceedings of the Graphics Interface 2000 Conference, May 15-17, 2000, Montr{'{e}}al, Qu{'{e}}bec, Canada},
year = {2000},
month = {May},
pages = {187--196},
url = {http://graphicsinterface.org/wp-content/uploads/gi2000-25.pdf}
}

@misc{feizi2025groundingcomputeruseagents,
      title={Grounding Computer Use Agents on Human Demonstrations}, 
      author={Aarash Feizi and Shravan Nayak and Xiangru Jian and Kevin Qinghong Lin and Kaixin Li and Rabiul Awal and Xing Han Lù and Johan Obando-Ceron and Juan A. Rodriguez and Nicolas Chapados and David Vazquez and Adriana Romero-Soriano and Reihaneh Rabbany and Perouz Taslakian and Christopher Pal and Spandana Gella and Sai Rajeswar},
      year={2025},
      eprint={2511.07332},
      archivePrefix={arXiv},
      primaryClass={cs.LG},
      url={https://arxiv.org/abs/2511.07332}, 
}

@misc{koh2026generativevisualcodemobile,
      title={Generative Visual Code Mobile World Models}, 
      author={Woosung Koh and Sungjun Han and Segyu Lee and Se-Young Yun and Jamin Shin},
      year={2026},
      eprint={2602.01576},
      archivePrefix={arXiv},
      primaryClass={cs.LG},
      url={https://arxiv.org/abs/2602.01576}, 
}

@article{neuralos,
  title={Neuralos: Towards simulating operating systems via neural generative models},
  author={Rivard, Luke and Sun, Sun and Guo, Hongyu and Chen, Wenhu and Deng, Yuntian},
  journal={arXiv preprint arXiv:2507.08800},
  year={2025}
}

@article{yourllm,
  title={Is your llm secretly a world model of the internet? model-based planning for web agents},
  author={Gu, Yu and Zhang, Kai and Ning, Yuting and Zheng, Boyuan and Gou, Boyu and Xue, Tianci and Chang, Cheng and Srivastava, Sanjari and Xie, Yanan and Qi, Peng and others},
  journal={arXiv preprint arXiv:2411.06559},
  year={2024}
}

@misc{o3,
  author       = "{OpenAI}",
  title        = {Introducing {OpenAI} o3 and o4-mini},
  year         = {2025},
  month        = {April},
  howpublished = {\url{https://openai.com/index/introducing-o3-and-o4-mini/}}
}

@misc{zhou2025maiuitechnicalreportrealworld,
      title={MAI-UI Technical Report: Real-World Centric Foundation GUI Agents}, 
      author={Hanzhang Zhou and Xu Zhang and Panrong Tong and Jianan Zhang and Liangyu Chen and Quyu Kong and Chenglin Cai and Chen Liu and Yue Wang and Jingren Zhou and Steven Hoi},
      year={2025},
      eprint={2512.22047},
      archivePrefix={arXiv},
      primaryClass={cs.CV},
      url={https://arxiv.org/abs/2512.22047}, 
}

@misc{ye2025guiarpenhancinggroundingadaptive,
      title={GUI-ARP: Enhancing Grounding with Adaptive Region Perception for GUI Agents}, 
      author={Xianhang Ye and Yiqing Li and Wei Dai and Miancan Liu and Ziyuan Chen and Zhangye Han and Hongbo Min and Jinkui Ren and Xiantao Zhang and Wen Yang and Zhi Jin},
      year={2025},
      eprint={2509.15532},
      archivePrefix={arXiv},
      primaryClass={cs.CV},
      url={https://arxiv.org/abs/2509.15532}, 
}

@misc{zhang2025hyperclickadvancingreliablegui,
      title={HyperClick: Advancing Reliable GUI Grounding via Uncertainty Calibration}, 
      author={Shaojie Zhang and Pei Fu and Ruoceng Zhang and Jiahui Yang and Anan Du and Xiuwen Xi and Shaokang Wang and Ying Huang and Bin Qin and Zhenbo Luo and Jian Luan},
      year={2025},
      eprint={2510.27266},
      archivePrefix={arXiv},
      primaryClass={cs.CV},
      url={https://arxiv.org/abs/2510.27266}, 
}

@misc{liu2025infiguig1advancingguigrounding,
      title={InfiGUI-G1: Advancing GUI Grounding with Adaptive Exploration Policy Optimization}, 
      author={Yuhang Liu and Zeyu Liu and Shuanghe Zhu and Pengxiang Li and Congkai Xie and Jiasheng Wang and Xavier Hu and Xiaotian Han and Jianbo Yuan and Xinyao Wang and Shengyu Zhang and Hongxia Yang and Fei Wu},
      year={2025},
      eprint={2508.05731},
      archivePrefix={arXiv},
      primaryClass={cs.AI},
      url={https://arxiv.org/abs/2508.05731}, 
}

@misc{gu2025uivenustechnicalreportbuilding,
      title={UI-Venus Technical Report: Building High-performance UI Agents with RFT}, 
      author={Zhangxuan Gu and Zhengwen Zeng and Zhenyu Xu and Xingran Zhou and Shuheng Shen and Yunfei Liu and Beitong Zhou and Changhua Meng and Tianyu Xia and Weizhi Chen and Yue Wen and Jingya Dou and Fei Tang and Jinzhen Lin and Yulin Liu and Zhenlin Guo and Yichen Gong and Heng Jia and Changlong Gao and Yuan Guo and Yong Deng and Zhenyu Guo and Liang Chen and Weiqiang Wang},
      year={2025},
      eprint={2508.10833},
      archivePrefix={arXiv},
      primaryClass={cs.CV},
      url={https://arxiv.org/abs/2508.10833}, 
}

@misc{bai2025qwen3vltechnicalreport,
      title={Qwen3-VL Technical Report}, 
      author={Shuai Bai and Yuxuan Cai and Ruizhe Chen and Keqin Chen and Xionghui Chen and Zesen Cheng and Lianghao Deng and Wei Ding and Chang Gao and Chunjiang Ge and Wenbin Ge and Zhifang Guo and Qidong Huang and Jie Huang and Fei Huang and Binyuan Hui and Shutong Jiang and Zhaohai Li and Mingsheng Li and Mei Li and Kaixin Li and Zicheng Lin and Junyang Lin and Xuejing Liu and Jiawei Liu and Chenglong Liu and Yang Liu and Dayiheng Liu and Shixuan Liu and Dunjie Lu and Ruilin Luo and Chenxu Lv and Rui Men and Lingchen Meng and Xuancheng Ren and Xingzhang Ren and Sibo Song and Yuchong Sun and Jun Tang and Jianhong Tu and Jianqiang Wan and Peng Wang and Pengfei Wang and Qiuyue Wang and Yuxuan Wang and Tianbao Xie and Yiheng Xu and Haiyang Xu and Jin Xu and Zhibo Yang and Mingkun Yang and Jianxin Yang and An Yang and Bowen Yu and Fei Zhang and Hang Zhang and Xi Zhang and Bo Zheng and Humen Zhong and Jingren Zhou and Fan Zhou and Jing Zhou and Yuanzhi Zhu and Ke Zhu},
      year={2025},
      eprint={2511.21631},
      archivePrefix={arXiv},
      primaryClass={cs.CV},
      url={https://arxiv.org/abs/2511.21631}, 
}

@article{bonatti2024windows,
  author      = {Bonatti, Rogerio and Zhao, Dan and Bonacci, Francesco and Dupont, Dillon and Abdali, Sara and Li, Yinheng and Wagle, Justin and Koishida, Kazuhito and Bucker, Arthur and Jang, Lawrence and Hui, Zack},
  title       = {Windows Agent Arena: Evaluating Multi-Modal OS Agents at Scale},
  institution = {Microsoft},
  year        = {2024},
  month       = {September}
}

@misc{workarena2024,
  title        = {WorkArena: How Capable Are Web Agents at Solving Common Knowledge Work Tasks?},
  author       = {Alexandre Drouin and Maxime Gasse and Massimo Caccia and Issam H. Laradji and Manuel Del Verme and Tom Marty and Léo Boisvert and Megh Thakkar and Quentin Cappart and David Vazquez and Nicolas Chapados and Alexandre Lacoste},
  year         = {2024},
  eprint       = {2403.07718},
  archivePrefix= {arXiv},
  primaryClass = {cs.LG}
}

@article{ariaui,
  title   = {Aria-UI: Visual Grounding for GUI Instructions},
  author  = {Yuhao Yang and Yue Wang and Dongxu Li and Ziyang Luo and Bei Chen and Chao Huang and Junnan Li},
  year    = {2024},
  journal = {arXiv preprint arXiv:2412.16256}
}

@misc{xu2024aguvisunifiedpurevision,
      title={Aguvis: Unified Pure Vision Agents for Autonomous GUI Interaction}, 
      author={Yiheng Xu and Zekun Wang and Junli Wang and Dunjie Lu and Tianbao Xie and Amrita Saha and Doyen Sahoo and Tao Yu and Caiming Xiong},
      year={2024},
      eprint={2412.04454},
      archivePrefix={arXiv},
      primaryClass={cs.CL},
      url={https://arxiv.org/abs/2412.04454}, 
}

@article{gou2024uground,
    title={Navigating the Digital World as Humans Do: Universal Visual Grounding for GUI Agents},
    author={Boyu Gou and Ruohan Wang and Boyuan Zheng and Yanan Xie and Cheng Chang and Yiheng Shu and Huan Sun and Yu Su},
    journal={arXiv preprint arXiv:2410.05243},
    year={2024},
    url={https://arxiv.org/abs/2410.05243},
}

@article{cogagent,
  title={Cogagent: A visual language model for gui agents},
  author={Hong, Wenyi and Wang, Weihan and Lv, Qingsong and Xu, Jiazheng and Yu, Wenmeng and Ji, Junhui and Wang, Yan and Wang, Zihan and Dong, Yuxiao and Ding, Ming and others},
  journal={arXiv preprint arXiv:2312.08914},
  year={2023}
}

@article{mind2web,
  title={Mind2web: Towards a generalist agent for the web},
  author={Deng, Xiang and Gu, Yu and Zheng, Boyuan and Chen, Shijie and Stevens, Sam and Wang, Boshi and Sun, Huan and Su, Yu},
  journal={Advances in Neural Information Processing Systems},
  volume={36},
  year={2024}
}

@article{aitw,
  title={Android in the wild: A large-scale dataset for android device control},
  author={Rawles, Christopher and Li, Alice and Rodriguez, Daniel and Riva, Oriana and Lillicrap, Timothy},
  journal={arXiv preprint arXiv:2307.10088},
  year={2023}
}

@article{osworld,
  title={OSWorld: Benchmarking Multimodal Agents for Open-Ended Tasks in Real Computer Environments},
  author={Xie, Tianbao and Zhang, Danyang and Chen, Jixuan and Li, Xiaochuan and Zhao, Siheng and Cao, Ruisheng and Hua, Toh Jing and Cheng, Zhoujun and Shin, Dongchan and Lei, Fangyu and others},
  journal={arXiv preprint arXiv:2404.07972},
  year={2024}
}

@inproceedings{deka2017rico,
  title={Rico: A mobile app dataset for building data-driven design applications},
  author={Deka, Biplab and Huang, Zifeng and Franzen, Chad and Hibschman, Joshua and Afergan, Daniel and Li, Yang and Nichols, Jeffrey and Kumar, Ranjitha},
  booktitle={Proceedings of the 30th annual ACM symposium on user interface software and technology},
  pages={845--854},
  year={2017}
}

@article{webarena,
  title={Webarena: A realistic web environment for building autonomous agents},
  author={Zhou, Shuyan and Xu, Frank F and Zhu, Hao and Zhou, Xuhui and Lo, Robert and Sridhar, Abishek and Cheng, Xianyi and Bisk, Yonatan and Fried, Daniel and Alon, Uri and others},
  journal={arXiv preprint arXiv:2307.13854},
  year={2023}
}

@article{seeclick,
  title={Seeclick: Harnessing gui grounding for advanced visual gui agents},
  author={Cheng, Kanzhi and Sun, Qiushi and Chu, Yougang and Xu, Fangzhi and Li, Yantao and Zhang, Jianbing and Wu, Zhiyong},
  journal={arXiv preprint arXiv:2401.10935},
  year={2024}
}

@article{pixel2act,
  title={From pixels to ui actions: Learning to follow instructions via graphical user interfaces},
  author={Shaw, Peter and Joshi, Mandar and Cohan, James and Berant, Jonathan and Pasupat, Panupong and Hu, Hexiang and Khandelwal, Urvashi and Lee, Kenton and Toutanova, Kristina N},
  journal={Advances in Neural Information Processing Systems},
  volume={36},
  year={2024}
}

@inproceedings{miniwob++,
  title={World of bits: An open-domain platform for web-based agents},
  author={Shi, Tianlin and Karpathy, Andrej and Fan, Linxi and Hernandez, Jonathan and Liang, Percy},
  booktitle={International Conference on Machine Learning},
  pages={3135--3144},
  year={2017},
  organization={PMLR}
}

@article{uibert,
  title={Uibert: Learning generic multimodal representations for ui understanding},
  author={Bai, Chongyang and Zang, Xiaoxue and Xu, Ying and Sunkara, Srinivas and Rastogi, Abhinav and Chen, Jindong and others},
  journal={arXiv preprint arXiv:2107.13731},
  year={2021}
}

@article{guicourse,
  title={GUICourse: From General Vision Language Models to Versatile GUI Agents},
  author={Chen, Wentong and Cui, Junbo and Hu, Jinyi and Qin, Yujia and Fang, Junjie and Zhao, Yue and Wang, Chongyi and Liu, Jun and Chen, Guirong and Huo, Yupeng and others},
  journal={arXiv preprint arXiv:2406.11317},
  year={2024}
}

@article{amex,
  title={AMEX: Android Multi-annotation Expo Dataset for Mobile GUI Agents},
  author={Chai, Yuxiang and Huang, Siyuan and Niu, Yazhe and Xiao, Han and Liu, Liang and Zhang, Dingyu and Gao, Peng and Ren, Shuai and Li, Hongsheng},
  journal={arXiv preprint arXiv:2407.17490},
  year={2024}
}

@article{guiworld,
  title={GUI-WORLD: A Dataset for GUI-oriented Multimodal LLM-based Agents},
  author={Chen, Dongping and Huang, Yue and Wu, Siyuan and Tang, Jingyu and Chen, Liuyi and Bai, Yilin and He, Zhigang and Wang, Chenlong and Zhou, Huichi and Li, Yiqiang and others},
  journal={arXiv preprint arXiv:2406.10819},
  year={2024}
}

@article{guiodyssey,
  title={GUI Odyssey: A Comprehensive Dataset for Cross-App GUI Navigation on Mobile Devices},
  author={Lu, Quanfeng and Shao, Wenqi and Liu, Zitao and Meng, Fanqing and Li, Boxuan and Chen, Botong and Huang, Siyuan and Zhang, Kaipeng and Qiao, Yu and Luo, Ping},
  journal={arXiv preprint arXiv:2406.08451},
  year={2024}
}

@inproceedings{videogui,
  title={VideoGUI: A Benchmark for GUI Automation from Instructional Videos.},
  author={Lin, Kevin Qinghong and Li, Linjie and Gao, Difei and Wu, Qinchen and Yan, Mingyi and Yang, Zhengyuan and Wang, Lijuan and Shou, Mike Zheng},
  booktitle={NeurIPS},
  year={2024}
}

@article{androidworld,
  title={AndroidWorld: A dynamic benchmarking environment for autonomous agents},
  author={Rawles, Christopher and Clinckemaillie, Sarah and Chang, Yifan and Waltz, Jonathan and Lau, Gabrielle and Fair, Marybeth and Li, Alice and Bishop, William and Li, Wei and Campbell-Ajala, Folawiyo and others},
  journal={arXiv preprint arXiv:2405.14573},
  year={2024}
}

@article{omniparser,
  title={Omniparser for pure vision based gui agent},
  author={Lu, Yadong and Yang, Jianwei and Shen, Yelong and Awadallah, Ahmed},
  journal={arXiv preprint arXiv:2408.00203},
  year={2024}
}

@article{omniact,
  title={OmniACT: A Dataset and Benchmark for Enabling Multimodal Generalist Autonomous Agents for Desktop and Web},
  author={Kapoor, Raghav and Butala, Yash Parag and Russak, Melisa and Koh, Jing Yu and Kamble, Kiran and Alshikh, Waseem and Salakhutdinov, Ruslan},
  journal={arXiv preprint arXiv:2402.17553},
  year={2024}
}

@misc{agashe2024agentsopenagentic,
      title={Agent S: An Open Agentic Framework that Uses Computers Like a Human}, 
      author={Saaket Agashe and Jiuzhou Han and Shuyu Gan and Jiachen Yang and Ang Li and Xin Eric Wang},
      year={2024},
      eprint={2410.08164},
      archivePrefix={arXiv},
      primaryClass={cs.AI},
      url={https://arxiv.org/abs/2410.08164}, 
}

@inproceedings{webui,
  title={Webui: A dataset for enhancing visual ui understanding with web semantics},
  author={Wu, Jason and Wang, Siyan and Shen, Siman and Peng, Yi-Hao and Nichols, Jeffrey and Bigham, Jeffrey P},
  booktitle={Proceedings of the 2023 CHI Conference on Human Factors in Computing Systems},
  pages={1--14},
  year={2023}
}

@article{uitars2025,
  author    = {Yujia Qin and Yining Ye and Junjie Fang and Haoming Wang and Shihao Liang and Shizuo Tian and Junda Zhang and Jiahao Li and Yunxin Li and Shijue Huang and Wanjun Zhong and Kuanye Li and Jiale Yang and Yu Miao and Woyu Lin and Longxiang Liu and Xu Jiang and Qianli Ma and Jingyu Li and Xiaojun Xiao and Kai Cai and Chuang Li and Yaowei Zheng and Chaolin Jin and Chen Li and Xiao Zhou and Minchao Wang and Haoli Chen and Zhaojian Li and Haihua Yang and Haifeng Liu and Feng Lin and Tao Peng and Xin Liu and Guang Shi},
  title     = {UI-TARS: Pioneering Automated GUI Interaction with Native Agents},
  journal   = {arXiv preprint arXiv:2501.12326},
  url       = {https://github.com/bytedance/UI-TARS},
  year      = {2025}
}

@inproceedings{screenspot-pro,
author = {Li, Kaixin and Meng, Ziyang and Lin, Hongzhan and Luo, Ziyang and Tian, Yuchen and Ma, Jing and Huang, Zhiyong and Chua, Tat-Seng},
title = {ScreenSpot-Pro: GUI Grounding for Professional High-Resolution Computer Use},
year = {2025},
isbn = {9798400720352},
publisher = {Association for Computing Machinery},
address = {New York, NY, USA},
url = {https://doi.org/10.1145/3746027.3755688},
doi = {10.1145/3746027.3755688},
booktitle = {Proceedings of the 33rd ACM International Conference on Multimedia},
pages = {8778–8786},
numpages = {9},
location = {Dublin, Ireland},
series = {MM '25}
}

@article{xie2025scaling,
  title={Scaling Computer-Use Grounding via User Interface Decomposition and Synthesis},
  author={Xie, Tianbao and Deng, Jiaqi and Li, Xiaochuan and Yang, Junlin and Wu, Haoyuan and Chen, Jixuan and Hu, Wenjing and Wang, Xinyuan and Xu, Yuhui and Wang, Zekun and others},
  journal={arXiv preprint arXiv:2505.13227},
  year={2025}
}

@misc{nayak2025uivisiondesktopcentricguibenchmark,
      title={UI-Vision: A Desktop-centric GUI Benchmark for Visual Perception and Interaction}, 
      author={Shravan Nayak and Xiangru Jian and Kevin Qinghong Lin and Juan A. Rodriguez and Montek Kalsi and Rabiul Awal and Nicolas Chapados and M. Tamer Özsu and Aishwarya Agrawal and David Vazquez and Christopher Pal and Perouz Taslakian and Spandana Gella and Sai Rajeswar},
      year={2025},
      eprint={2503.15661},
      archivePrefix={arXiv},
      primaryClass={cs.CV},
      url={https://arxiv.org/abs/2503.15661}, 
}

@article{zhang2025phi,
  title={Phi-Ground Tech Report: Advancing Perception in GUI Grounding},
  author={Zhang, Miaosen and Xu, Ziqiang and Zhu, Jialiang and Dai, Qi and Qiu, Kai and Yang, Yifan and Luo, Chong and Chen, Tianyi and Wagle, Justin and Franklin, Tim and others},
  journal={arXiv preprint arXiv:2507.23779},
  year={2025}
}

@misc{OpenCUA2025, 
  title={OpenCUA: Open Foundations for Computer-Use Agents}, 
  author={Wang, Xinyuan and Wang, Bowen and Lu, Dunjie and Yang, Junlin and Xie, Tianbao and Wang, Junli and Deng, Jiaqi and Guo, Xiaole and Xu, Yiheng and Wu, Chen Henry and Shen, Zhennan and Li, Zhuokai and Li, Ryan and Li, Xiaochuan and Chen, Junda and Zheng, Boyuan and Li, Peihang and Lei, Fangyu and Cao, Ruisheng and Fu, Yeqiao and Shin, Dongchan and Shin, Martin and Hu, Jiarui and Wang, Yuyan and Chen, Jixuan and Ye, Yuxiao and Zhang, Danyang and Wang, Yipu and Wang, Heng and Yang, Diyi and Zhong, Victor and Charles, Y. and Yang, Zhilin and Yu, Tao}, 
  year={2025}, 
  url={https://opencua.xlang.ai/} 
}

@article{liu2025infigui,
  title={Infigui-r1: Advancing multimodal gui agents from reactive actors to deliberative reasoners},
  author={Liu, Yuhang and Li, Pengxiang and Xie, Congkai and Hu, Xavier and Han, Xiaotian and Zhang, Shengyu and Yang, Hongxia and Wu, Fei},
  journal={arXiv preprint arXiv:2504.14239},
  year={2025}
}

@misc{wu2024osatlasfoundationactionmodel,
      title={OS-ATLAS: A Foundation Action Model for Generalist GUI Agents}, 
      author={Zhiyong Wu and Zhenyu Wu and Fangzhi Xu and Yian Wang and Qiushi Sun and Chengyou Jia and Kanzhi Cheng and Zichen Ding and Liheng Chen and Paul Pu Liang and Yu Qiao},
      year={2024},
      eprint={2410.23218},
      archivePrefix={arXiv},
      primaryClass={cs.CL},
      url={https://arxiv.org/abs/2410.23218}, 
}

@misc{muryn2025screen2axvisionbasedapproachautomatic,
      title={Screen2AX: Vision-Based Approach for Automatic macOS Accessibility Generation}, 
      author={Viktor Muryn and Marta Sumyk and Mariya Hirna and Sofiya Garkot and Maksym Shamrai},
      year={2025},
      eprint={2507.16704},
      archivePrefix={arXiv},
      primaryClass={cs.LG},
      url={https://arxiv.org/abs/2507.16704}, 
}

@misc{zhao2025learningguigroundingspatial,
      title={Learning GUI Grounding with Spatial Reasoning from Visual Feedback}, 
      author={Yu Zhao and Wei-Ning Chen and Huseyin Atahan Inan and Samuel Kessler and Lu Wang and Lukas Wutschitz and Fangkai Yang and Chaoyun Zhang and Pasquale Minervini and Saravan Rajmohan and Robert Sim},
      year={2025},
      eprint={2509.21552},
      archivePrefix={arXiv},
      primaryClass={cs.CV},
      url={https://arxiv.org/abs/2509.21552}, 
}

@misc{luo2025vimogenerativevisualgui,
      title={ViMo: A Generative Visual GUI World Model for App Agents}, 
      author={Dezhao Luo and Bohan Tang and Kang Li and Georgios Papoudakis and Jifei Song and Shaogang Gong and Jianye Hao and Jun Wang and Kun Shao},
      year={2025},
      eprint={2504.13936},
      archivePrefix={arXiv},
      primaryClass={cs.HC},
      url={https://arxiv.org/abs/2504.13936}, 
}

@inproceedings{
xu2025agenttrek,
title={AgentTrek: Agent Trajectory Synthesis via Guiding Replay with Web Tutorials},
author={Yiheng Xu and Dunjie Lu and Zhennan Shen and Junli Wang and Zekun Wang and Yuchen Mao and Caiming Xiong and Tao Yu},
booktitle={The Thirteenth International Conference on Learning Representations},
year={2025},
url={https://openreview.net/forum?id=EEgYUccwsV}
}

@article{sun2024osgenesis,
  title={OS-Genesis: Automating GUI Agent Trajectory Construction via Reverse Task Synthesis},
  author={Sun, Qiushi and Cheng, Kanzhi and Ding, Zichen and Jin, Chuanyang and Wang, Yian and Xu, Fangzhi and Wu, Zhenyu and Jia, Chengyou and Chen, Liheng and Liu, Zhoumianze and others},
  journal={arXiv preprint arXiv:2412.19723},
  year={2024}
}

@inproceedings{
liu2026scalecua,
title={Scale{CUA}: Scaling Open-Source Computer Use Agents with Cross-Platform Data},
author={Zhaoyang Liu and JingJing Xie and Zichen Ding and Zehao Li and Bowen Yang and Zhenyu Wu and Xuehui Wang and Qiushi Sun and Shi Liu and Weiyun Wang and Shenglong Ye and Qingyun Li and Zeyue Tian and Gen Luo and Xiangyu Yue and Biqing Qi and Kai Chen and Bowen Zhou and Yu Qiao and Qifeng Chen and Wenhai Wang},
booktitle={The Fourteenth International Conference on Learning Representations},
year={2026},
url={https://openreview.net/forum?id=yBFUqdJFZn}
}

@misc{zhou2025venusbenchgdcomprehensivemultiplatformgui,
      title={VenusBench-GD: A Comprehensive Multi-Platform GUI Benchmark for Diverse Grounding Tasks}, 
      author={Beitong Zhou and Zhexiao Huang and Yuan Guo and Zhangxuan Gu and Tianyu Xia and Zichen Luo and Fei Tang and Dehan Kong and Yanyi Shang and Suling Ou and Zhenlin Guo and Changhua Meng and Shuheng Shen},
      year={2025},
      eprint={2512.16501},
      archivePrefix={arXiv},
      primaryClass={cs.CV},
      url={https://arxiv.org/abs/2512.16501}, 
}

@misc{zhang2025tonguiinternetscaletrajectoriesmultimodal,
      title={TongUI: Internet-Scale Trajectories from Multimodal Web Tutorials for Generalized GUI Agents}, 
      author={Bofei Zhang and Zirui Shang and Zhi Gao and Wang Zhang and Rui Xie and Xiaojian Ma and Tao Yuan and Xinxiao Wu and Song-Chun Zhu and Qing Li},
      year={2025},
      eprint={2504.12679},
      archivePrefix={arXiv},
      primaryClass={cs.CV},
      url={https://arxiv.org/abs/2504.12679}, 
}

@misc{hui2025winclickguigroundingmultimodal,
      title={WinClick: GUI Grounding with Multimodal Large Language Models}, 
      author={Zheng Hui and Yinheng Li and Dan zhao and Tianyi Chen and Colby Banbury and Kazuhito Koishida},
      year={2025},
      eprint={2503.04730},
      archivePrefix={arXiv},
      primaryClass={cs.CL},
      url={https://arxiv.org/abs/2503.04730}, 
}

@misc{cui2025paddleocrvlboostingmultilingualdocument,
      title={PaddleOCR-VL: Boosting Multilingual Document Parsing via a 0.9B Ultra-Compact Vision-Language Model}, 
      author={Cheng Cui and Ting Sun and Suyin Liang and Tingquan Gao and Zelun Zhang and Jiaxuan Liu and Xueqing Wang and Changda Zhou and Hongen Liu and Manhui Lin and Yue Zhang and Yubo Zhang and Handong Zheng and Jing Zhang and Jun Zhang and Yi Liu and Dianhai Yu and Yanjun Ma},
      year={2025},
      eprint={2510.14528},
      archivePrefix={arXiv},
      primaryClass={cs.CV},
      url={https://arxiv.org/abs/2510.14528}, 
}

@inproceedings{showui,
  title={Showui: One vision-language-action model for gui visual agent},
  author={Lin, Kevin Qinghong and Li, Linjie and Gao, Difei and Yang, Zhengyuan and Wu, Shiwei and Bai, Zechen and Lei, Stan Weixian and Wang, Lijuan and Shou, Mike Zheng},
  booktitle={Proceedings of the Computer Vision and Pattern Recognition Conference},
  pages={19498--19508},
  year={2025}
}

@article{code2world,
  title={Code2World: A GUI World Model via Renderable Code Generation},
  author={Zheng, Yuhao and Zhong, Li'an and Wang, Yi and Dai, Rui and Liu, Kaikui and Chu, Xiangxiang and Lv, Linyuan and Torr, Philip and Lin, Kevin Qinghong},
  journal={arXiv preprint arXiv:2602.09856},
  year={2026}
}

@article{
zhang2025large,
title={Large Language Model-Brained {GUI} Agents: A Survey},
author={Chaoyun Zhang and Shilin He and Jiaxu Qian and Bowen Li and Liqun Li and Si Qin and Yu Kang and Minghua Ma and Guyue Liu and Qingwei Lin and Saravan Rajmohan and Dongmei Zhang and Qi Zhang},
journal={Transactions on Machine Learning Research},
issn={2835-8856},
year={2025},
url={https://openreview.net/forum?id=xChvYjvXTp},
note={}
}

@misc{fdm1,
      title={The First Fully General Computer Action Model},
      author={Neel Redkar and Yudhister Kumar and Devansh Pandey and Galen Mead},
      year={2026},
      howpublished={\url{https://si.inc/posts/fdm1/}},
      note={Standard Intelligence Blog, February 2026},
}

@inproceedings{nguyen-etal-2025-gui,
    title = "{GUI} Agents: A Survey",
    author = "Nguyen, Dang  and
      Chen, Jian  and
      Wang, Yu  and
      Wu, Gang  and
      Park, Namyong  and
      Hu, Zhengmian  and
      Lyu, Hanjia  and
      Wu, Junda  and
      Aponte, Ryan  and
      Xia, Yu  and
      Li, Xintong  and
      Shi, Jing  and
      Chen, Hongjie  and
      Lai, Viet Dac  and
      Xie, Zhouhang  and
      Kim, Sungchul  and
      Zhang, Ruiyi  and
      Yu, Tong  and
      Tanjim, Mehrab  and
      Ahmed, Nesreen K.  and
      Mathur, Puneet  and
      Yoon, Seunghyun  and
      Yao, Lina  and
      Kveton, Branislav  and
      Kil, Jihyung  and
      Nguyen, Thien Huu  and
      Bui, Trung  and
      Zhou, Tianyi  and
      Rossi, Ryan A.  and
      Dernoncourt, Franck",
    editor = "Che, Wanxiang  and
      Nabende, Joyce  and
      Shutova, Ekaterina  and
      Pilehvar, Mohammad Taher",
    booktitle = "Findings of the Association for Computational Linguistics: ACL 2025",
    month = jul,
    year = "2025",
    address = "Vienna, Austria",
    publisher = "Association for Computational Linguistics",
    url = "https://aclanthology.org/2025.findings-acl.1158/",
    doi = "10.18653/v1/2025.findings-acl.1158",
    pages = "22522--22538",
    ISBN = "979-8-89176-256-5"
}

@misc{gurbuz2026movingsparsegroundingcomplete,
      title={Moving Beyond Sparse Grounding with Complete Screen Parsing Supervision}, 
      author={A. Said Gurbuz and Sunghwan Hong and Ahmed Nassar and Marc Pollefeys and Peter Staar},
      year={2026},
      eprint={2602.14276},
      archivePrefix={arXiv},
      primaryClass={cs.CV},
      url={https://arxiv.org/abs/2602.14276}, 
}

@misc{song2026exevrm,
      title={Video-Based Reward Modeling for Computer-Use Agents}, 
      author={Linxin Song and Jieyu Zhang and Huanxin Sheng and Taiwei Shi and Gupta Rahul and Yang Liu and Ranjay Krishna and Jian Kang and Jieyu Zhao},
      year={2026},
      eprint={2603.10178},
      archivePrefix={arXiv},
      primaryClass={cs.CV},
      url={https://arxiv.org/abs/2603.10178}, 
}

@misc{pang2025,
      title={Paper2Poster: Towards Multimodal Poster Automation from Scientific Papers}, 
      author={Wei Pang and Kevin Qinghong Lin and Xiangru Jian and Xi He and Philip Torr},
      year={2025},
      eprint={2505.21497},
      archivePrefix={arXiv},
      primaryClass={cs.CV},
      url={https://arxiv.org/abs/2505.21497}, 
}

@misc{guan2026computerusingworldmodel,
      title={Computer-Using World Model}, 
      author={Yiming Guan and Rui Yu and John Zhang and Lu Wang and Chaoyun Zhang and Liqun Li and Bo Qiao and Si Qin and He Huang and Fangkai Yang and Pu Zhao and Lukas Wutschitz and Samuel Kessler and Huseyin A Inan and Robert Sim and Saravan Rajmohan and Qingwei Lin and Dongmei Zhang},
      year={2026},
      eprint={2602.17365},
      archivePrefix={arXiv},
      primaryClass={cs.SE},
      url={https://arxiv.org/abs/2602.17365}, 
}
\bibliographystyle{iclr2026_conference}

\newpage

\appendix

\section*{\LARGE Appendix}
\vspace{8pt} 

\noindent\textbf{\Large Table of Contents}

\begin{flushright}
    \textbf{Page}
\end{flushright}

\noindent
\renewcommand{\arraystretch}{1.3} 
\begin{tabularx}{\linewidth}{Xr} 
    \textbf{A. \dataset{} Creation} \dotfill & \pageref{app:dataset_creation} \\
    \hspace{2em} A.1 Platforms \dotfill & \pageref{app:platforms} \\
    \hspace{2em} A.2 Human Annotation \dotfill & \pageref{app:human_annotation} \\
    \textbf{B. Additional \actcua{} Evaluation Analysis} \dotfill & \pageref{sec:actcua_eval_details} \\
    \hspace{2em} B.1 Evaluation Preprocessing \dotfill & \pageref{sec:eval_preprocessing} \\
    \hspace{2em} B.2 Application-Level Analysis \dotfill & \pageref{sec:app_level_analysis} \\
    \hspace{2em} B.3 Error Distance Analysis \dotfill & \pageref{sec:error_distance_analysis} \\
    \textbf{C. Application Potential of \dataset{}} \dotfill & \pageref{sec:utility_details} \\
    \hspace{2em} C.1 Advancing Generalist Screen Parsing \dotfill & \pageref{app:screen_parsing} \\
    \hspace{2em} C.2 Learning Continuous Spatial Control \dotfill & \pageref{app:spa_ctrl} \\
    \hspace{2em} C.3 Enabling Visual World Models \dotfill & \pageref{app:world_model} \\
    \hspace{2em} C.4 Video-Based Reward Modeling \dotfill & \pageref{app:reward_model} \\
    \textbf{D. \actcua{} Trajectory Examples} \dotfill & \pageref{sec:actcua_examples} \\
    \hspace{2em} D.1 Krita (Digital Art) \dotfill & \pageref{sec:actcua_example_92083} \\
    \hspace{2em} D.2 GIMP (Image Editing) \dotfill & \pageref{sec:actcua_example_75737} \\



    




    
\end{tabularx}

\newpage

\section{\dataset{} Creation}
\label{app:dataset_creation}

\subsection{Platforms}
\label{app:platforms}

\rowcolors{2}{black!5}{white}

\begin{table*}[h]
\caption{Categories of desktop applications and their corresponding applications.}
\label{tab:platforms}
\centering
\renewcommand{\arraystretch}{1.4}
\setlength{\tabcolsep}{8pt}
\begin{tabularx}{\linewidth}{lX}
\toprule
\textbf{Category} & \textbf{Platforms} \\
\midrule
Education & Anki, Zotero, Calibre, OpenBoard, Mendeley \\
Browsers & Brave, Chromium, Mozilla Firefox, DuckDuckGo \\
Development & VS Code, Atom, Eclipse, NetBeans, PyCharm, IntelliJ IDEA, Brackets, Geany, Bluefish, KDevelop, Komodo Edit, Code::Blocks, Qt Creator, Arduino IDE \\
Productivity & LibreOffice Calc, LibreOffice Draw, LibreOffice Impress, LibreOffice Writer, draw.io, Joplin, OpenProject, Affine, PDFedit, OnlyOffice Calendar, OnlyOffice Document Editor, OnlyOffice Forms, OnlyOffice PDF Forms, OnlyOffice Presentation, OnlyOffice Spreadsheet, Nextcloud, Gnumeric, Simplenote, WeKan \\
Graphics and Design & Blender, GIMP, Inkscape, Krita, darktable, FontForge, Scribus, WordPress \\
Video and Audio Production & OpenShot, OBS Studio, Lightworks, Shotcut, Natron, OpenToonz, Audacity, MuseScore \\
Communication & Element, Signal, Mastodon, Lemmy, Matrix, Zulip, Jitsi \\
Entertainment & VLC Media Player, Kodi, Emby \\
System Utilities & Ubuntu Terminal, Conky, Bash, 7-Zip, Flameshot, Nemo, gedit \\
Security & Bitwarden, Cryptomator \\
Finance and Business Analytics & GnuCash, Frappe Books, Metabase \\
Scientific & RStudio, Veusz, GNU Octave, GrassGIS, QGIS, FreeCAD, Spyder \\
\bottomrule
\end{tabularx}

\end{table*}

\begin{table*}[t]
\caption{UI element categories in \dataset{} with descriptions and representative examples.}
\label{tab:ui_categories}
\centering
\small
\setlength{\tabcolsep}{6pt}
\renewcommand{\arraystretch}{1.1}
\begin{tabular}{p{3cm} p{10cm}}
\toprule
\textbf{Category} & \textbf{Description and Common UI Elements} \\
\midrule
\textbf{Input Element} & Interactive fields where users enter or modify data, like text boxes, checkboxes, radio buttons, etc. \\

\textbf{Sidebar} & Vertical or horizontal panels that provide quick access to tools or navigation.  
Examples include tool palettes, folder trees, settings sidebars. \\

\textbf{Information Display} & Regions that primarily present textual or numerical information.  
Examples include labels, console outputs, document text, and code blocks. \\

\textbf{Button} & Clickable controls that trigger an action like submit button, “OK/Cancel” buttons, play/pause buttons \\

\textbf{Navigation} & Elements that help users move within or across applications.  
Examples: tabs, back/forward arrows etc. \\

\textbf{Visual Elements} & Non-textual graphical elements that convey information or functionality.  Examples include icons, thumbnails, images, charts, and progress bars. \\

\textbf{Menu} & Structured lists of commands or options, often hierarchical.  
Examples: file menu, context menu, dropdown menus. \\

\textbf{Others} & Elements not covered by the above categories, often decorative or container elements like spacers.\\
\bottomrule
\end{tabular}

\end{table*}

We select 87 platforms, focusing on open-source software with permissive licenses. These applications span 12 diverse categories, detailed in \Cref{tab:platforms}. Our selection is motivated by the under-representation of such platforms in existing datasets and the flexibility provided by permissive licensing, which enables dataset release with minimal restrictions. We further show that this choice does not compromise generalization \citep{feizi2025groundingcomputeruseagents}, as the open-source software usually shares UI elements and layout with its closed-source counterparts. For instance, LibreOffice and Microsoft Office Suite share many interface elements, layout, and functionality. This ensures broader applicability of \dataset{}.

\subsection{Human Annotation}
\label{app:human_annotation}

We collaborated with a professional data-labeling vendor specializing in dataset curation for AI applications. The annotation effort spanned three phases, beginning with a pilot study where we worked closely with the annotation team to refine task instructions and provide iterative feedback. The annotation team consisted of around 70 individuals, organized into multiple tiers of annotators, quality assurance specialists, and project managers. The majority of the team was based in India and Latin America, with participants aged 20–35 and a balanced gender distribution. All annotators held at least a bachelor’s degree in technical fields such as Computer Science or Engineering and had prior experience in data labeling and user interface research.

Annotators underwent a training process to become familiar with the platforms and annotation guidelines. They were compensated hourly, with each task requiring on average 60–90 minutes to complete, including quality checks. The process began with the creation of computer-use tasks for 87 software applications (see Table~\ref{tab:platforms}). Annotators then executed these tasks while screen recordings were collected. From these recordings, we extracted keyframes corresponding to major user interactions. Each keyframe was annotated using a custom tool, where annotators drew bounding boxes around all visible interface elements. For each bounding box, annotators assigned a label corresponding to the element’s name, or, in the case of textual elements, the text was also provided in addition to the element name. For long text segments such as source code or lengthy descriptions, annotators provided a concise summary that captured the main theme. To supplement these summaries, we also applied OCR using PaddleOCR~\citep{cui2025paddleocrvlboostingmultilingualdocument} to extract the full text when available. In addition, 50\% element was assigned to one of eight high-level categories. We applied rigorous quality assurance at multiple stages. Annotations were reviewed by dedicated quality specialists, cross-checked by the authors, and validated using custom evaluation scripts. This pipeline allowed us to construct a large-scale dataset of grounded user interface interactions with high diversity and reliable annotation quality.

\section{Additional \actcua{} Evaluation Analysis}
\label{sec:actcua_eval_details}

This appendix presents evaluation preprocessing details and supplementary analyses for the action prediction experiments described in Section~\ref{sec:actcua}.

\subsection{Evaluation Preprocessing}
\label{sec:eval_preprocessing}

\noindent \textbf{Keyframe Extraction.} Standard processed screenshots are captured at or slightly before the action timestamp, which may reveal the cursor's ground-truth target position. To mitigate this information leakage, we extract \textbf{keyframes} at the temporal midpoint between consecutive actions: for action $a_t$ with timestamp $\tau_t$, the keyframe is captured at $(\tau_{t-1} + \tau_t) / 2$. This ensures the cursor has not yet reached the target location, providing a fairer evaluation of spatial grounding.

\noindent \textbf{moveTo Handling.} Desktop interactions frequently consist of \texttt{moveTo}$\to$\texttt{click} pairs targeting the same coordinates. Since the \texttt{moveTo} action is a preparatory cursor movement rather than a semantically distinct interaction, we exclude \texttt{moveTo} steps from both evaluation and action history. Furthermore, for \texttt{click} actions that directly follow a \texttt{moveTo}, we use the \texttt{moveTo} step's keyframe instead of the \texttt{click} step's keyframe, as the latter may still capture the cursor near the target position.

\subsection{Application-Level Analysis}
\label{sec:app_level_analysis}

We analyze per-application performance using the 32B model results. Table~\ref{tab:app_performance} presents the most challenging and most successful applications.

\begin{table}[h]
\centering
\caption{\textbf{Per-Application Performance (OpenCUA-32B).} Applications exhibiting the lowest and highest @50px success rates (minimum 3 coordinate predictions). Performance varies widely, from under 5\% to over 70\%, reflecting the diversity of desktop application interfaces.}
\label{tab:app_performance}
\vspace{0.5em}
\begin{tabular}{llccc}
\toprule
\textbf{Application} & \textbf{Domain} & \textbf{Preds} & \textbf{Mean Px} & \textbf{@50px} \\
\midrule
\multicolumn{5}{l}{\textit{Lowest @50px}} \\
Darktable & Photo Editing & 28 & 259.0 & 3.6\% \\
OnlyOffice Forms & Forms & 17 & 232.6 & 11.8\% \\
Krita & Digital Art & 31 & 417.8 & 12.9\% \\
Kodi & Media Center & 41 & 591.9 & 14.6\% \\
GnuCash & Accounting & 52 & 359.2 & 21.2\% \\
FreeCAD & 3D CAD & 31 & 289.4 & 22.6\% \\
QGIS & GIS & 31 & 224.9 & 22.6\% \\
\midrule
\multicolumn{5}{l}{\textit{Highest @50px}} \\
OnlyOffice Spreadsheet & Spreadsheet & 15 & 123.9 & 73.3\% \\
IntelliJ IDEA & IDE & 21 & 127.0 & 66.7\% \\
LibreOffice Calc & Spreadsheet & 24 & 223.4 & 66.7\% \\
VLC Media Player & Media Player & 21 & 80.2 & 66.7\% \\
DuckDuckGo & Web Browser & 14 & 100.1 & 64.3\% \\
Metabase & Dashboard & 24 & 180.8 & 62.5\% \\
Brave & Web Browser & 21 & 122.5 & 61.9\% \\

\bottomrule
\end{tabular}
\end{table}

\noindent \textbf{Challenging Applications.} Applications with specialized visual interfaces consistently show the lowest success rates. Creative tools such as Darktable (3.6\% @50px) and Krita (12.9\%) feature dense toolbars, non-standard widget layouts, and domain-specific iconography. Canvas-based tools such as FreeCAD (22.6\%) and QGIS (22.6\%) pose similar challenges due to their complex, multi-panel interfaces. Media applications such as Kodi (14.6\%) also exhibit lower accuracy, likely due to overlay-based controls and dynamic content regions.

\noindent \textbf{Relatively Successful Applications.} Applications with layouts resembling web interfaces, such as web browsers (DuckDuckGo 64.3\%, Brave 61.9\%), spreadsheet editors (OnlyOffice Spreadsheet 73.3\%, LibreOffice Calc 66.7\%), and IDEs (IntelliJ IDEA 66.7\%), exhibit higher success rates. These applications typically feature standard toolbar arrangements and clear visual hierarchies, which may better align with the models' training distribution.

\subsection{Error Distance Analysis}
\label{sec:error_distance_analysis}

To characterize the distribution of prediction errors, we analyze the pixel distances of predictions that exceed the 50-pixel threshold. Table~\ref{tab:error_distribution} presents the distribution for both models.

\begin{table}[h]
\centering
\caption{\textbf{Distribution of Prediction Distances $>$50px.} Both models show a substantial proportion of high-distance predictions, with roughly one-third exceeding 500 pixels.}
\label{tab:error_distribution}
\vspace{0.5em}
\begin{tabular}{lcccc}
\toprule
& \multicolumn{2}{c}{\textbf{OpenCUA-32B}} & \multicolumn{2}{c}{\textbf{OpenCUA-7B}} \\
\cmidrule(lr){2-3} \cmidrule(lr){4-5}
\textbf{Distance Range} & \textbf{Count} & \textbf{\%} & \textbf{Count} & \textbf{\%} \\
\midrule
50--100px   & 252 & 20.2\% & 222 & 13.7\% \\
100--200px  & 252 & 20.2\% & 325 & 20.0\% \\
200--300px  & 150 & 12.0\% & 246 & 15.1\% \\
300--500px  & 205 & 16.5\% & 294 & 18.1\% \\
$>$500px    & 386 & 31.0\% & 537 & 33.1\% \\
\midrule
\textbf{Total ($>$50px)} & \textbf{1,245} & --- & \textbf{1,624} & --- \\
\bottomrule
\end{tabular}
\end{table}

\noindent \textbf{Distance Distribution.} Among predictions exceeding 50 pixels, approximately 20\% fall within 50--100 pixels of the target for the 32B model, representing cases where the predicted coordinates are in the general vicinity of the target. At the other end, 31.0\% (32B) and 33.1\% (7B) exceed 500 pixels, corresponding to predictions that land far from the intended target on screen. The 32B model produces fewer total high-distance predictions (1,245 vs.\ 1,624) and a higher fraction in the 50--100px range (20.2\% vs.\ 13.7\%), suggesting that scaling helps shift the distribution toward smaller distances rather than solely eliminating errors.

\noindent \textbf{Within-Task Variability.} A notable pattern is that many tasks exhibit mixed prediction quality: 128 of 256 evaluated tasks contain both predictions within 20 pixels of the target and predictions exceeding 500 pixels. This suggests that the model can accurately locate some UI elements within a task, whereas it predicts coordinates far from the target for others, rather than consistently performing well or poorly across all steps of a given task.

\section{Application Potential of \dataset{}}
\label{sec:utility_details}

This appendix expands on the four application frontiers introduced in Section~\ref{sec:more_utility}, providing detailed descriptions of how \dataset{} supports each use case.

\subsection{Advancing Generalist Screen Parsing}\label{app:screen_parsing}

The efficacy of vision-based agents is fundamentally limited by their ability to parse unstructured screenshots into structured elements. As highlighted by OmniParser~\citep{omniparser}, current Multimodal Large Language Models (MLLMs) struggle to reliably identify interactable regions and understand the functional semantics of icons without external modules.

\dataset{} provides the ideal supervision to train robust, desktop-centric screen parsers:
\begin{itemize}
    \item \textbf{Panoptic Element Coverage:} Unlike web-based datasets that rely on DOM trees, which often miss canvas-based or custom-drawn widgets, our dataset features dense, human-verified annotations that cover \textit{all} interactable elements on the screen. This provides the dense supervisory signal required to fine-tune the detection modules of architectures such as OmniParser for complex desktop GUIs.
    \item \textbf{Semantics-Aware Captioning:} Beyond mere localization, our metadata includes functional descriptions for icons and widgets. This rich semantic layer can be used to train the captioning components of screen parsers, enabling them to distinguish visually similar yet functionally distinct controls (e.g., a ``play'' button in a music player from a ``run'' button in an IDE).
\end{itemize}

\subsection{Learning Continuous Spatial Control} \label{app:spa_ctrl}

Recent work, such as GUI-Cursor~\citep{zhao2025learningguigroundingspatial}, argues that treating GUI interaction as a discrete coordinate-prediction task (i.e., teleporting the mouse) is insufficient for high-precision tasks. Instead, they propose learning continuous relative cursor movements from visual feedback. Training such policies requires dense kinematic trajectory data, which are absent from most step-by-step navigation benchmarks.

Our dataset is uniquely positioned to support this paradigm shift:
\begin{itemize}
    \item \textbf{Kinematic Trajectories:} Unlike standard datasets that only record the final click coordinates, our action logs capture the intermediate cursor movements (via \texttt{MOVE\_TO} events) and complete video context. This preserves the human kinematic priors of mouse movement, such as deceleration near targets (Fitts's Law).
    \item \textbf{Policy Training:} This continuous data allows for the training of imitation learning (IL) or offline reinforcement learning (RL) policies that mimic human navigation patterns. By leveraging our data, models such as GUI-Cursor can be trained to perform iterative, feedback-driven navigation, thereby significantly improving robustness to the spatial hallucinations common in zero-shot VLMs.
\end{itemize}

\subsection{Enabling Visual World Models} \label{app:world_model}

A rapidly emerging frontier in autonomous agents is the development of \textit{World Models}---systems capable of predicting future states given current observations and actions. As demonstrated by recent work such as ViMo (Visual World Model for App Agents), a key limitation in current research is the lack of datasets that pair fine-grained actions with high-fidelity future frames~\citep{luo2025vimogenerativevisualgui}. Most datasets provide sparse screenshots that omit the immediate visual feedback of an action (e.g., a button press or a menu opening).

\dataset{} uniquely enables the training of high-fidelity GUI world models:
\begin{itemize}
    \item \textbf{Dense Causal Pairs:} Our complete video recordings provide the perfect $(s_t, a_t, s_{t+1})$ triplets required for training video prediction models. Unlike sparse datasets, we capture the immediate visual consequences of user actions at 30fps.
    \item \textbf{Action-Conditioned Video Generation:} By conditioning generative video models on our verified annotations, researchers can train models to simulate the visual interface dynamics of complex software. This enables agents to perform \textit{visual lookahead planning} (i.e., simulating the outcome of a click before executing it), which is crucial for error prevention in mission-critical desktop workflows.
\end{itemize}

\subsection{Video-Based Reward Modeling} \label{app:reward_model}

A critical challenge in scaling reinforcement learning for computer-use agents is the reliable assessment of task completion. Traditional approaches depend on hand-crafted reward functions, environment-specific heuristics, or access to agent internals, none of which scale across diverse applications. ExeVRM~\citep{song2026exevrm} introduces a method-agnostic alternative: training vision-language models to evaluate task completion directly from execution videos, i.e., sequences of keyframes captured during agent trajectories. Their approach achieves 84.7\% accuracy and 87.7\% recall on their ExeVR-Bench benchmark spanning Ubuntu, macOS, Windows, and Android, outperforming proprietary models.

\dataset{} is well suited to support and advance this paradigm:
\begin{itemize}
    \item \textbf{Expert Positive Demonstrations:} Our approximately 10,000 complete expert task demonstrations, each captured as continuous 30\,fps video paired with a natural-language task instruction, provide a large-scale source of verified positive examples (successful task completions) for training video-based reward models. The diversity across 87 professional applications ensures broad coverage of the desktop domain where reward modeling is most challenging.
    \item \textbf{Step-Level Reward Signals:} ExeVRM relies on step-level annotations for synthesizing hard negative samples via adversarial instruction translation. \dataset{}'s multi-layered reasoning annotations, including per-step observations, thought chains, action descriptions, and reflections, provide exactly this granularity, enabling the construction of fine-grained reward signals that go beyond binary task-level labels.
    \item \textbf{Temporal Richness:} The spatiotemporal token pruning mechanism in ExeVRM is designed to identify decisive UI changes in long, high-resolution videos. Our continuous recordings preserve the full temporal dynamics, including subtle intermediate states (loading indicators, hover effects, transient feedback) that are critical for accurate reward assessment yet absent from sparse-screenshot datasets.
\end{itemize}


\section{\actcua{} Trajectory Examples}
\label{sec:actcua_examples}

This appendix presents complete Chain-of-Thought (CoT) annotated trajectories from two representative \actcua{} tasks spanning different application domains. Each step shows the screenshot observed by the agent, the generated observation of the current screen state, the reasoning chain (thought), the executed action as \texttt{pyautogui} code, and the post-action reflection. For coordinate-based actions, the target location is marked on the screenshot with a colored crosshair: {\color{red}red} for clicks, {\color{blue}blue} for cursor moves, {\color{green!70!black}green} for drags, {\color{orange}orange} for right-clicks, and {\color{violet}purple} for mouse-down events.

\subsection{Krita (Digital Art)}
\label{sec:actcua_example_92083}

\noindent\textbf{Task Instruction:} \textit{On this Krita platform, please Draw a circle shape and add a inner shadow using the Multiply blending mode after you filled the circle with the color green}

\noindent\textbf{Total Steps:} 20

\noindent\textbf{Coordinate Legend:} {\color{red}$\boldsymbol{\oplus}$ click} \quad {\color{blue}$\boldsymbol{\oplus}$ moveTo} \quad {\color{green!70!black}$\boldsymbol{\oplus}$ dragTo} \quad {\color{orange}$\boldsymbol{\oplus}$ rightClick} \quad {\color{violet}$\boldsymbol{\oplus}$ mouseDown}

\vspace{0.8em}
\noindent\rule{\textwidth}{0.4pt}

\noindent\textbf{Step 0} \hfill {\small\texttt{pyautogui.moveTo(x=0.0208, y=0.2198)}}

\vspace{0.3em}

\noindent
\begin{center}
  \includegraphics[width=\textwidth]{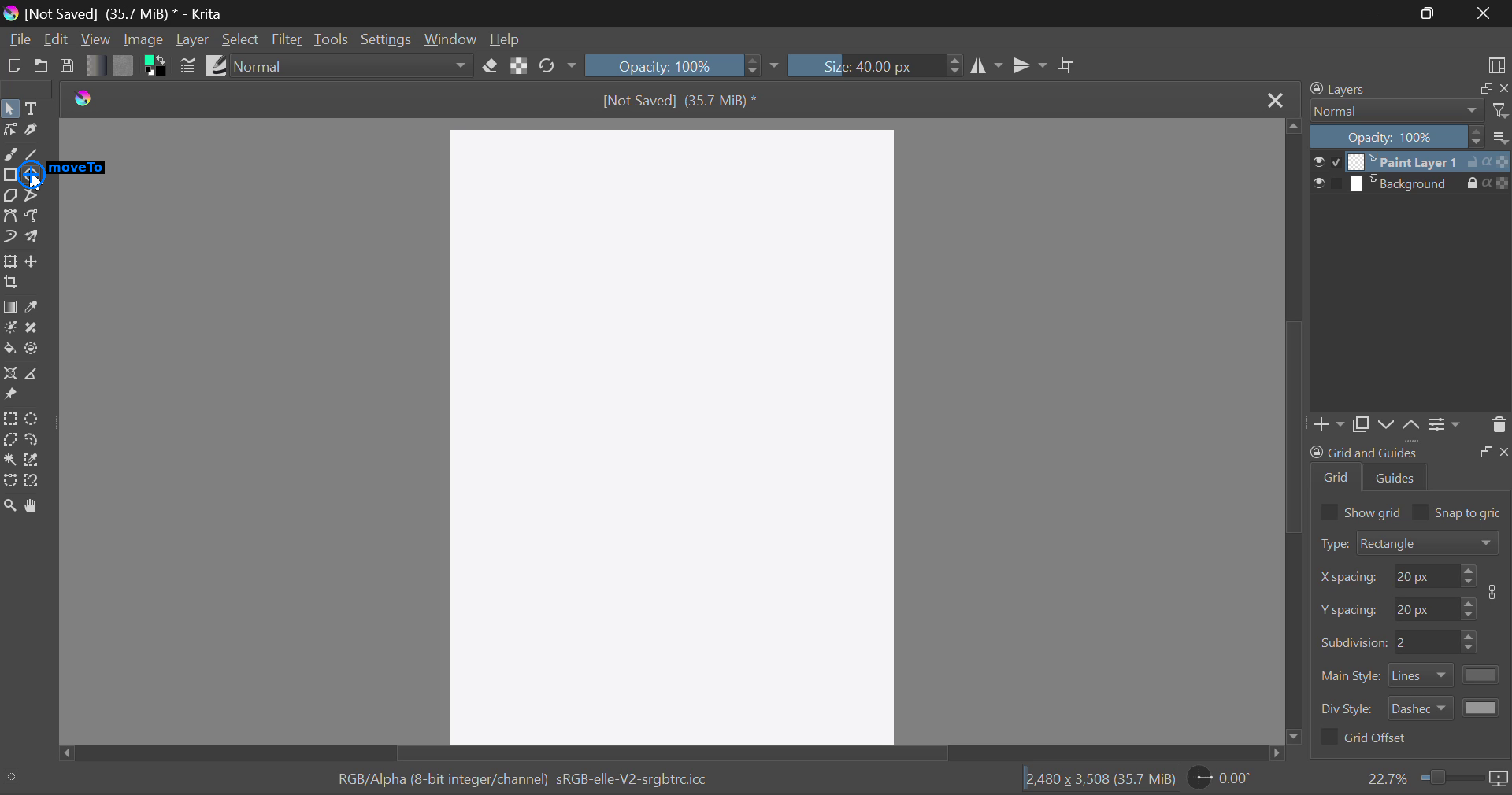}
\end{center}

\noindent\textbf{Observation:} The application is Krita, a digital painting software. The main canvas area is centered and displays a blank white canvas. On the left side, there is a vertical toolbar containing various drawing and editing tools. The toolbar includes icons for selection tools, brushes, shapes, and other utilities. Near the top of this toolbar are selection tools, followed by brush tools, and further down are shape tools. The right side panel shows the Layers section with "Paint Layer 1" and "Background" listed, along with opacity settings at 100\% and blending mode set to "Normal". Below that is a "Grid and Guides" section with grid configuration options. The top menu bar contains File, Edit, View, Image, Layer, Select, Filter, Tools, Settings, Window, and Help menus. The document shows "[Not Saved] (35.7 MiB)" in the title bar and tab.

\vspace{0.3em}

\noindent\textbf{Thought:} This is the beginning of the task, so no previous actions have been taken. I need to draw a circle shape, fill it with green, and add an inner shadow using Multiply blending mode. The first step is to select the appropriate tool for creating a circle. In Krita's left toolbar, I can see various tools arranged vertically. To draw a circle, I need to select the ellipse or circle shape tool. Looking at the toolbar structure, there are selection tools at the top, then drawing tools, and shape tools below. The ellipse tool would typically be among the shape tools in the middle-lower section of the toolbar. By selecting the ellipse/circle tool, I'll be able to draw a circular shape on the canvas. This is the necessary first step before I can fill it with green color and apply the inner shadow effect with Multiply blending mode.

\vspace{0.3em}

\noindent\textbf{Reflection:} Comparing the two full screenshots, there is no visible change between them. The mouse was supposed to move to the ellipse shape tool in the left vertical toolbar, but the screenshots appear identical with no indication that any tool has been selected or highlighted. Looking at the image patch showing the toolbar area, I can see various tools including what appears to be selection tools and shape tools. The action intended to move to coordinates (0.0208, 0.2198) which should be in the left toolbar area, but there's no visual feedback showing that the ellipse tool was actually selected or even hovered over. The action appears to have been a mouse movement only without a click, which explains why no tool selection occurred. However, even a hover action should typically show some visual feedback in most applications. The lack of any visible change suggests the action may not have been executed properly or the coordinates may not have targeted the correct tool. For the task of drawing a circle, the user needs to actually click on the ellipse/circle shape tool, not just move the mouse to it.

\vspace{0.8em}
\noindent\rule{\textwidth}{0.4pt}

\noindent\textbf{Step 1} \hfill {\small\texttt{pyautogui.click(x=0.0208, y=0.2178)}}

\vspace{0.3em}

\noindent
\begin{center}
  \includegraphics[width=\textwidth]{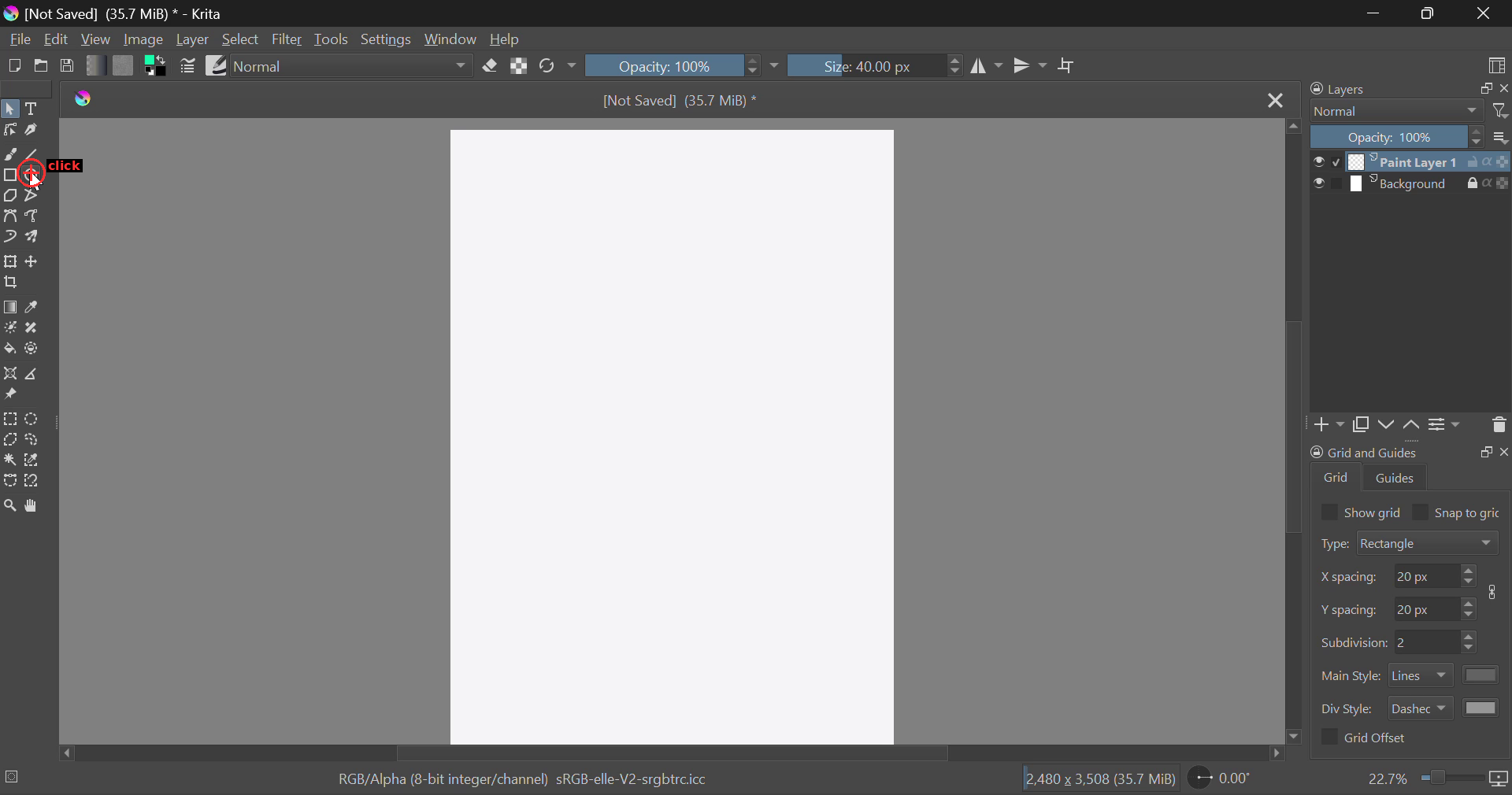}
\end{center}

\noindent\textbf{Observation:} The current state shows Krita is open with a blank white canvas in the center. The left vertical toolbar contains various drawing and shape tools arranged vertically. From top to bottom, the toolbar includes selection tools, drawing tools like brushes and pencils, and shape tools in the lower-middle section. Among the shape tools, I can see icons for rectangles, ellipses, and other geometric shapes. The canvas is ready for drawing, with "Paint Layer 1" visible and selected in the Layers panel on the right side. The document is marked as "[Not Saved]" in the title bar. The right panel displays layer information showing "Paint Layer 1" and "Background" layer, along with Grid and Guides settings in the lower portion. The blend mode is set to "Normal" and opacity is at 100\%.

\vspace{0.3em}

\noindent\textbf{Thought:} The previous action only moved the mouse to the ellipse tool location without clicking, which is why there's no visible change in the interface. Simply hovering over a tool doesn't activate it in Krita - I need to actually click on it to select it. Without the click, the ellipse tool remains unselected and I cannot proceed with the task of drawing a circle, filling it with green, and adding an inner shadow with Multiply blending mode.

My task requires me to: select the ellipse tool, draw a circle, fill it with green color, add an inner shadow effect, and set the blending mode to Multiply. Since the tool wasn't actually selected in the previous step, I need to click on the ellipse shape tool now to activate it. The ellipse tool is located in the left toolbar among the shape tools. Once clicked, it will become active and allow me to draw the circular shape needed for this task. This click action will properly select the tool and enable me to proceed with drawing the circle on the canvas.

\vspace{0.3em}

\noindent\textbf{Reflection:} The visual changes between the two screenshots show that the ellipse shape tool has been successfully selected. In the left toolbar, the ellipse tool icon now appears highlighted/selected (it has a lighter background compared to the first screenshot). Additionally, a small ellipse shape has appeared on the canvas in the upper-left area, which indicates that the tool is now active and ready to be used for drawing. The status bar at the bottom also shows 'Basic-1' which suggests a shape layer or tool state has been activated. This step was necessary and effective because it properly activated the ellipse tool, which is required to draw the circle shape as specified in the task. The action successfully clicked on the ellipse shape tool, enabling the next steps of drawing a circle, filling it with green, and adding an inner shadow with Multiply blending mode.

\vspace{0.8em}
\noindent\rule{\textwidth}{0.4pt}

\noindent\textbf{Step 2} \hfill {\small\texttt{pyautogui.moveTo(x=0.3359, y=0.3782)}}

\vspace{0.3em}

\noindent
\begin{center}
  \includegraphics[width=\textwidth]{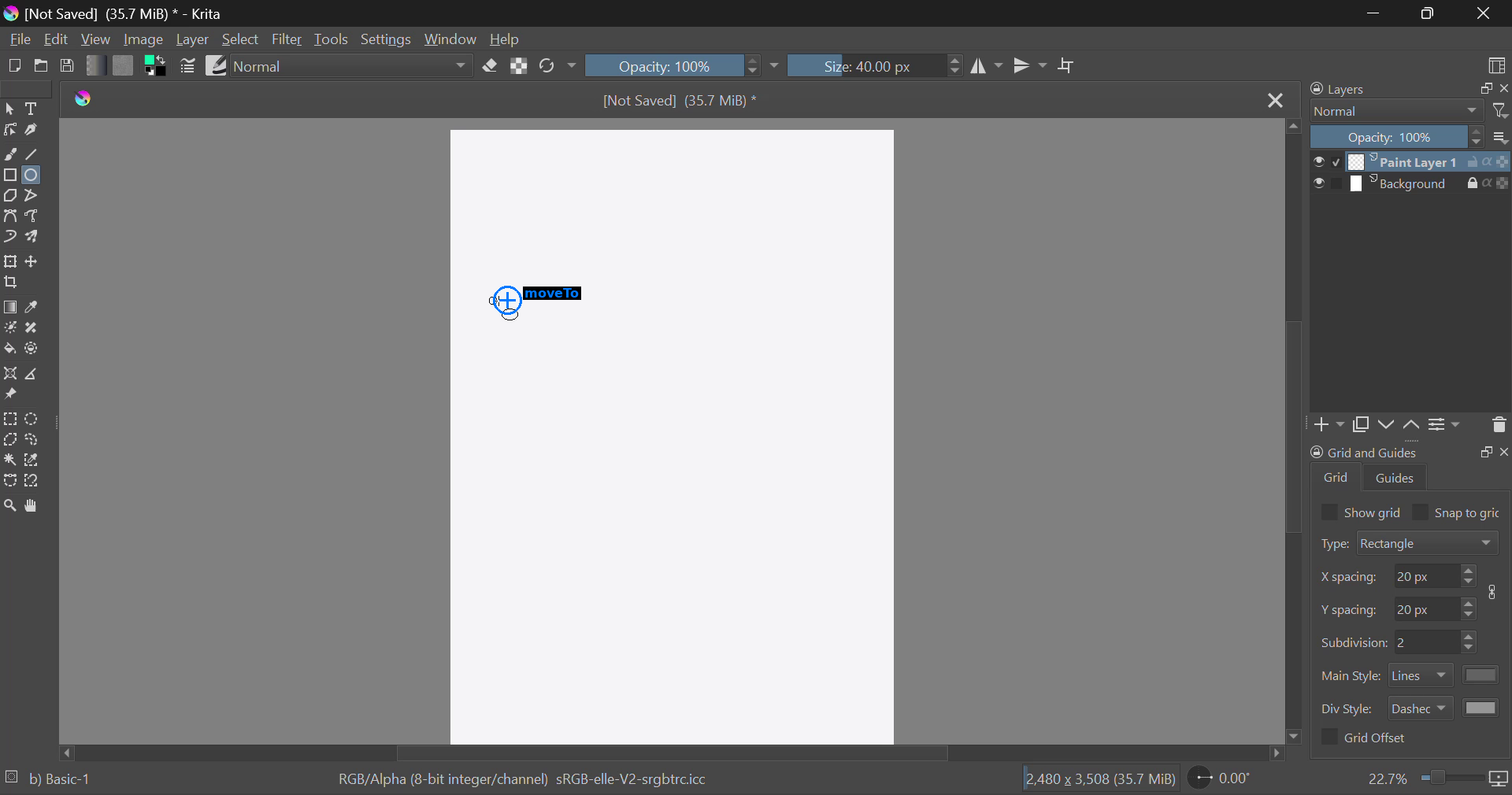}
\end{center}

\noindent\textbf{Observation:} The Krita application is open with a blank white canvas displayed in the center of the workspace. A small red circle outline is visible on the canvas in the upper-left area, indicating that the ellipse shape tool has been activated. The left toolbar shows various drawing and shape tools, with the ellipse tool appearing selected with a lighter background. The top toolbar displays the blend mode set to "Normal" and opacity at 100\%, with a brush size of 40.00 px. The right panel shows the Layers panel with "Paint Layer 1" currently active and visible, along with a "Background" layer below it. Below the Layers panel is a "Grid and Guides" section showing configuration options including Rectangle type, 20 px spacing for both X and Y, subdivision of 2, and line styles set to Lines and Dasher. The status bar at the bottom shows "Basic-1" and displays canvas dimensions of 2,480 x 3,508 at 22.7\% zoom.

\vspace{0.3em}

\noindent\textbf{Thought:} The ellipse tool has been successfully selected and is now active, as shown by the small red circle outline on the canvas and the highlighted tool in the left toolbar. This confirms the tool is ready for drawing operations.

Assessing my progress toward the complete task:
1. Select the ellipse tool - Completed successfully
2. Draw a circle shape - Need to start this step now
3. Fill the circle with green color - Pending
4. Add an inner shadow effect - Pending
5. Set blending mode to Multiply - Pending

The small circle currently visible on the canvas appears to be a preliminary shape from the tool activation. To properly complete the task, I need to draw a new circle of appropriate size. With the ellipse tool selected, the standard workflow is to click at a starting point on the canvas and then drag to create the circle shape.

I need to position at a suitable starting point on the canvas to begin drawing the circle. The canvas has ample white space available for drawing. A logical starting position would be in the upper-left to center area of the canvas, which provides enough room to drag and create a visible, properly-sized circle. This position will serve as the anchor point from which I'll drag to define the circle's dimensions. After creating the circle shape, I'll proceed with filling it with green color, adding the inner shadow effect, and setting the blending mode to Multiply as required.

\vspace{0.3em}

\noindent\textbf{Reflection:} The visual changes between the two screenshots show that the mouse cursor has moved from the upper-left area of the canvas (where a small red circle outline was visible) to a new position slightly below and to the right on the white canvas. The small red circle outline that was previously visible in the upper-left area is no longer present in the second screenshot, indicating the mouse has successfully moved to a new location. This step is necessary and effective because it positions the cursor at an appropriate starting point on the canvas to begin drawing the circle shape. The ellipse tool is still active (as shown in the left toolbar), and the cursor is now positioned in a suitable area of the canvas with enough space to drag and create a properly-sized circle. This is a logical preparatory step before the actual drawing action (click and drag) that will create the circle shape.

\vspace{0.8em}
\noindent\rule{\textwidth}{0.4pt}

\noindent\textbf{Step 3} \hfill {\small\texttt{pyautogui.mouseDown(button='left')}}

\vspace{0.3em}

\noindent
\begin{center}
  \includegraphics[width=\textwidth]{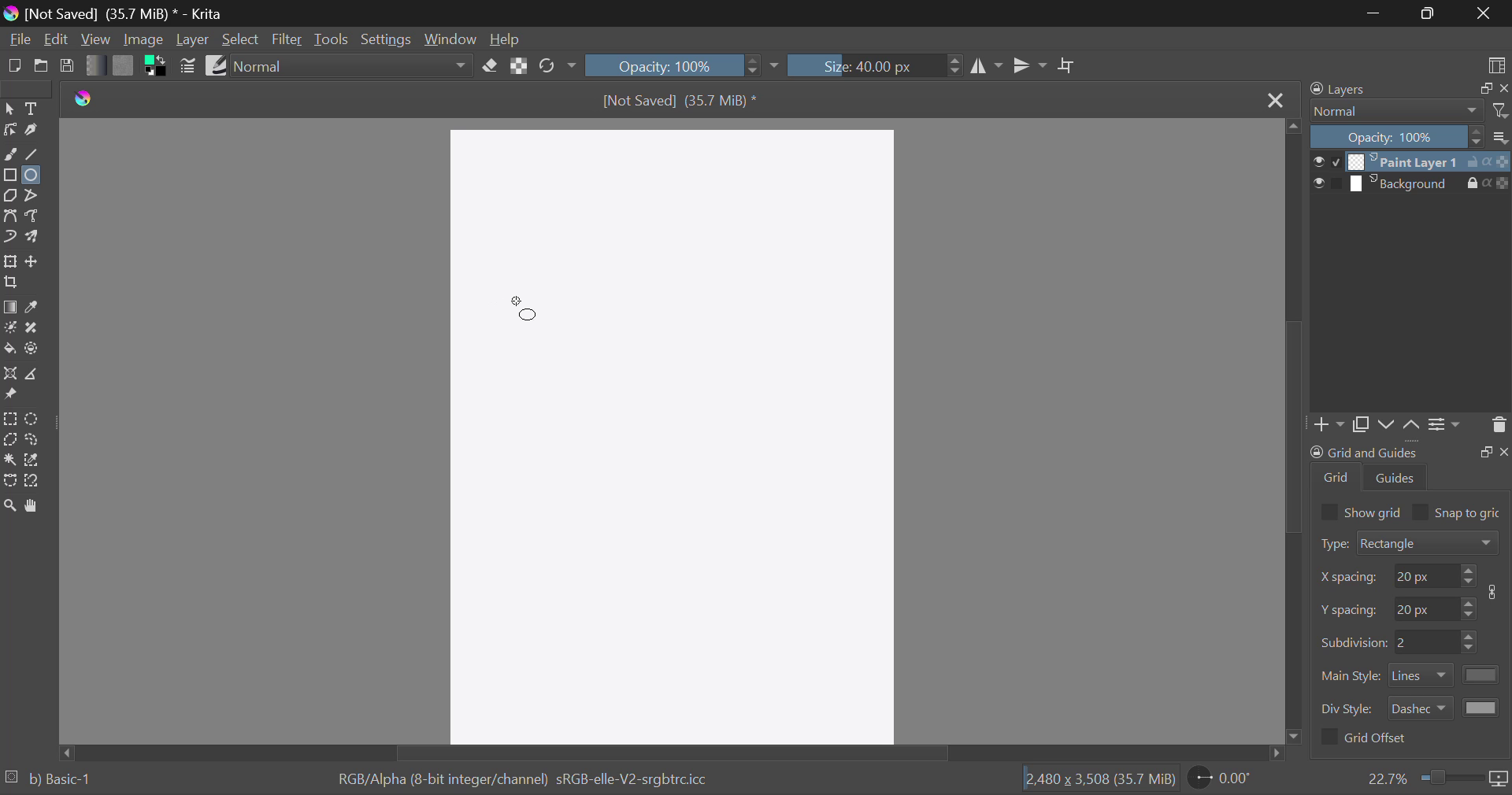}
\end{center}

\noindent\textbf{Observation:} The Krita application is open with a white canvas displayed in the center of the workspace. On the canvas, there is a small ellipse outline positioned in the upper-left to center area. The left vertical toolbar shows various drawing tools, with the ellipse tool currently selected and highlighted. The top toolbar displays "Normal" as the blending mode, "Opacity: 100\%", and "Size: 40.00 px". The right panel shows the Layers section with "Paint Layer 1" active and "Background" layer below it, both with Normal blending mode. Below the Layers panel is a "Grid and Guides" section showing grid configuration options. The title bar indicates the document is not saved, showing "[Not Saved] (35.7 MiB)".

\vspace{0.3em}

\noindent\textbf{Thought:} The cursor is now positioned on the canvas where a small ellipse outline is visible, indicating the ellipse tool is ready for use. The tool has been successfully selected from the left toolbar as shown by its highlighted state.

Assessing my progress toward the complete task:
1. Select the ellipse tool - Completed successfully
2. Draw a circle shape - In progress, need to execute the drawing action
3. Fill the circle with green color - Not yet started
4. Add an inner shadow effect - Not yet started
5. Set blending mode to Multiply - Not yet started

To draw a circle with the ellipse tool in Krita, I need to perform a click-and-drag operation. This involves pressing the left mouse button down at the current cursor position to establish the starting anchor point, then dragging to define the circle's dimensions, and finally releasing to complete the shape. The cursor is positioned in a suitable area of the canvas with adequate space to create a properly-sized circle.

The next logical step is to press down the left mouse button at the current cursor position. This will initiate the drawing process by anchoring the starting point of the circle. After this, I will need to drag the mouse to expand the circle to the desired size and then release the button to finalize the shape. Once the circle is created, I can proceed with the remaining steps: filling it with green color, adding the inner shadow effect, and changing the blending mode to Multiply.

\vspace{0.3em}

\noindent\textbf{Reflection:} Between the two screenshots, there is a visible change: a circle outline has been drawn on the canvas. In the first screenshot, there was only a small ellipse cursor indicator near the top-left area. In the second screenshot, a complete circle outline is now visible in the center of the canvas, with dimension information displayed in the top-left corner showing 'Width: 1,282 px Height: 1,314 px'. The mouse cursor has also moved to a different position near the bottom-right of the circle. This indicates that the mouseDown action successfully initiated the circle drawing process, and the user (or subsequent action) completed the drag operation to create the circle shape. The step is necessary and effective as it is the required action to begin drawing a circle with the ellipse tool in Krita.

\vspace{0.8em}
\noindent\rule{\textwidth}{0.4pt}

\noindent\textbf{Step 4} \hfill {\small\texttt{pyautogui.dragTo(x=0.4927, y=0.6703, button='left')}}

\vspace{0.3em}

\noindent
\begin{center}
  \includegraphics[width=\textwidth]{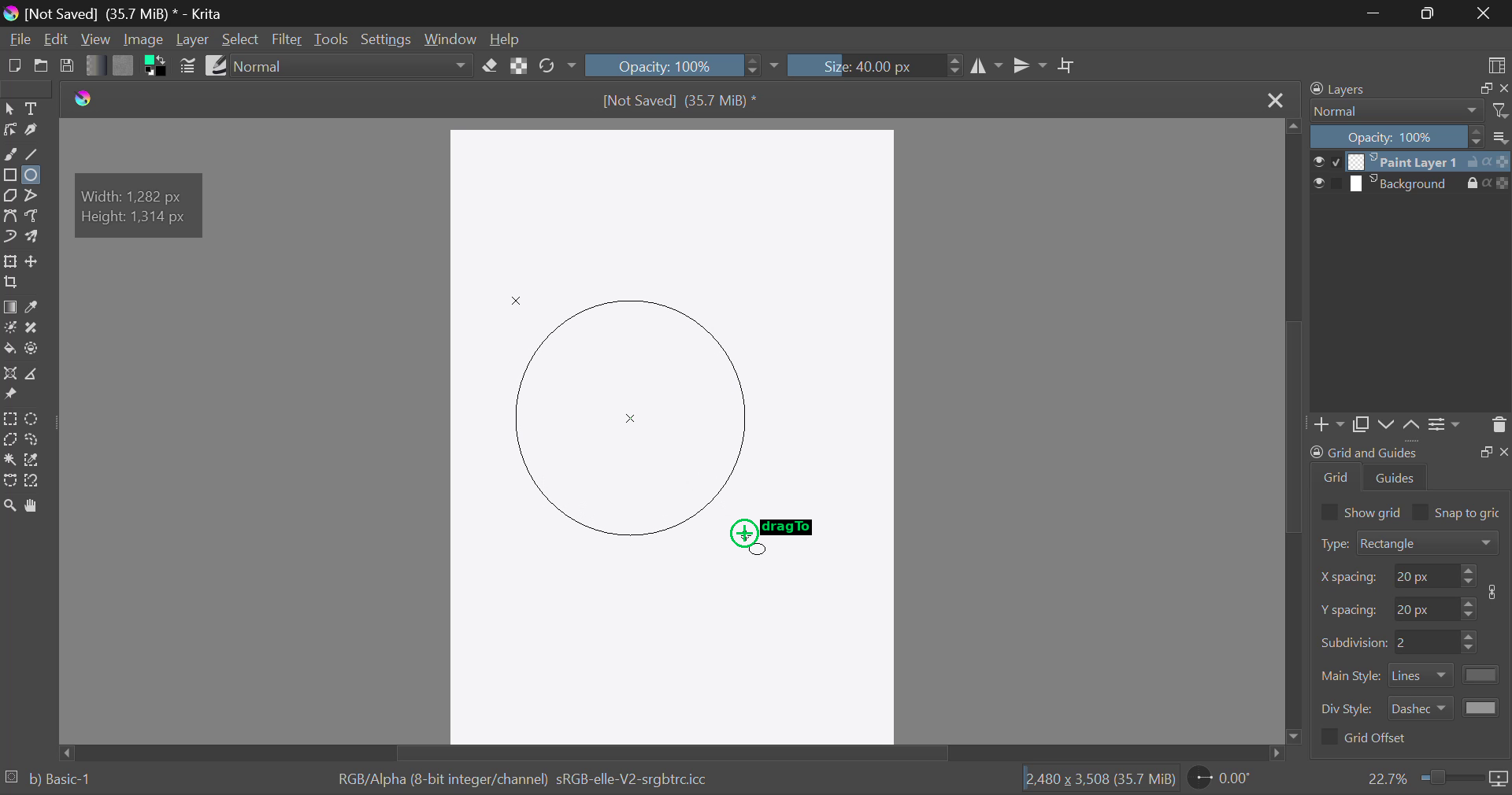}
\end{center}

\noindent\textbf{Observation:} The Krita application is currently active with a white canvas displayed in the center. A circle outline has been drawn on the canvas, positioned slightly above center and to the left. The dimension information in the top-left corner shows "Width: 1,282 px Height: 1,314 px". 

On the left side, there is a vertical toolbar containing various drawing and selection tools. The ellipse tool appears to be selected in this toolbar. At the top of the window, there is a menu bar with options including File, Edit, View, Image, Layer, Select, Filter, Tools, Settings, Window, and Help. Below the menu bar is a toolbar showing brush settings including "Normal" blending mode, "Opacity: 100\%", and "Size: 40.00 px".

On the right side of the interface, there is a Layers panel showing "Paint Layer 1" and "Background" layers. The panel indicates "Normal" blending mode and "Opacity: 100\%". Below that is a "Grid and Guides" panel with options for grid display, snap to grid, grid type set to "Rectangle", X spacing at 20 px, Y spacing at 20 px, subdivision at 2, main style set to "Lines", and div style set to "Dasher".

The canvas shows a circle outline with small crosshair markers visible. The document is labeled as "[Not Saved] (35.7 MiB)" in the title bar.

\vspace{0.3em}

\noindent\textbf{Thought:} The circle outline is now visible on the canvas, which confirms that the mouseDown action successfully initiated the drawing process. The dimensions displayed indicate the circle is being formed with a width of 1,282 px and height of 1,314 px. Since the mouse button is currently held down from the previous action, I need to continue the drag operation to finalize the circle's size and shape.

Assessing my progress:
1. Select the ellipse tool - Completed successfully
2. Draw a circle shape - Currently in progress, the outline is visible but the drag operation needs to be completed
3. Fill the circle with green color - Not yet started
4. Add an inner shadow effect - Not yet started
5. Set blending mode to Multiply - Not yet started

To complete the circle drawing, I need to drag to a position that will create a well-proportioned circle. The current outline shows the circle is being formed, and I need to drag to a point that maintains the circular shape. Looking at the canvas layout and the current circle position, dragging toward the lower right area of the circle will help complete the shape with appropriate dimensions.

After completing this drag operation, I will need to release the mouse button to finalize the circle shape. Once the circle is created, I can proceed with the remaining tasks: filling it with green color, adding an inner shadow effect, and changing the blending mode to Multiply.

\vspace{0.3em}

\noindent\textbf{Reflection:} The visual changes between the two screenshots show that the circle shape has been successfully completed. In the first screenshot, there was a red circle indicator showing the drag action in progress at the lower right area of the circle outline. In the second screenshot, the red circle indicator has moved to a different position (lower right, outside the main circle), and the circle shape appears to be fully formed with clean, complete outlines. The circle is now a complete ellipse shape on the canvas. This step was necessary and effective as it completed the drag operation that was initiated in the previous step, finalizing the circle's dimensions and shape. The action successfully moved the mouse to create a properly sized circle, which is the first required element for the task of drawing a circle and adding an inner shadow with green fill.

\vspace{0.8em}
\noindent\rule{\textwidth}{0.4pt}

\noindent\textbf{Step 5} \hfill {\small\texttt{pyautogui.mouseUp(button='left')}}

\vspace{0.3em}

\noindent
\begin{center}
  \includegraphics[width=\textwidth]{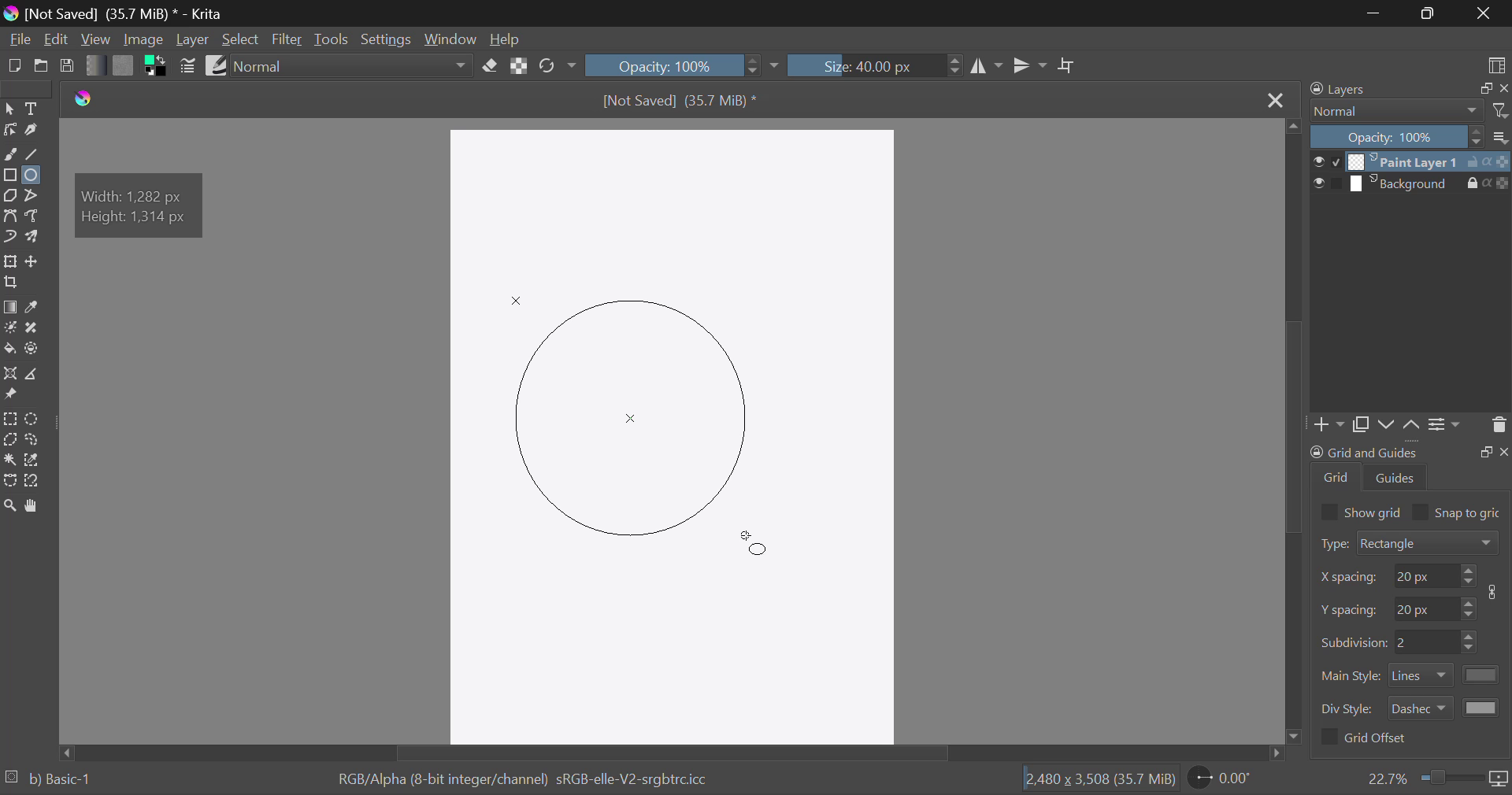}
\end{center}

\noindent\textbf{Observation:} The Krita application is currently active with a white canvas displayed in the center. A circle shape outline is visible on the canvas, positioned slightly left of center and occupying a significant portion of the canvas area. The circle appears to be a complete, clean outline with dimensions shown in a tooltip (Width: 1,282 px, Height: 1,314 px) in the upper left area of the canvas. The left toolbar shows various drawing tools, with the ellipse tool appearing to be selected. The right panel displays layer information showing "Paint Layer 1" and "Background" layers. The blending mode dropdown is currently set to "Normal" with 100\% opacity visible in the top toolbar. At the bottom of the screen, the canvas dimensions are shown as 2,480 x 3,508 pixels. There are no pop-ups, error messages, or loading states visible.

\vspace{0.3em}

\noindent\textbf{Thought:} The circle shape has been successfully drawn on the canvas with a complete outline visible. The drag operation from the previous steps has created a well-formed circle with the dimensions displayed. Since the mouse button was held down during the dragging process to create the circle, I now need to release it to finalize the shape creation.

Assessing my progress toward completing the task:
1. Select the ellipse tool - Completed successfully
2. Draw a circle shape - The outline is complete and visible, but the drawing action needs to be finalized
3. Fill the circle with green color - Not yet started
4. Add an inner shadow effect - Not yet started
5. Set blending mode to Multiply - Not yet started

The next logical step is to release the left mouse button to complete the circle drawing action. This will finalize the shape and allow me to proceed with filling it with green color and adding the inner shadow effect with Multiply blending mode. Once the mouse button is released, the circle will be established as a complete shape object on the canvas, enabling me to access its properties for applying fill colors and effects.

\vspace{0.3em}

\noindent\textbf{Reflection:} The visual changes between the two screenshots show that the circle has been finalized and filled with a green color. In the first screenshot, the circle had a thin black outline with construction markers (small x marks) visible, indicating it was still in the drawing/editing state. In the second screenshot, the circle now has a solid green stroke/outline with no construction markers visible, indicating the shape has been completed and committed. The file size also increased from 35.7 MiB to 43.3 MiB, confirming that the action was successfully applied. This step was necessary and effective as it completed the circle drawing action by releasing the mouse button, which finalized the shape and allowed Krita to render it properly. This is a required step before proceeding to fill the circle and add effects.

\vspace{0.8em}
\noindent\rule{\textwidth}{0.4pt}

\noindent\textbf{Step 6} \hfill {\small\texttt{pyautogui.moveTo(x=0.0125, y=0.4436)}}

\vspace{0.3em}

\noindent
\begin{center}
  \includegraphics[width=\textwidth]{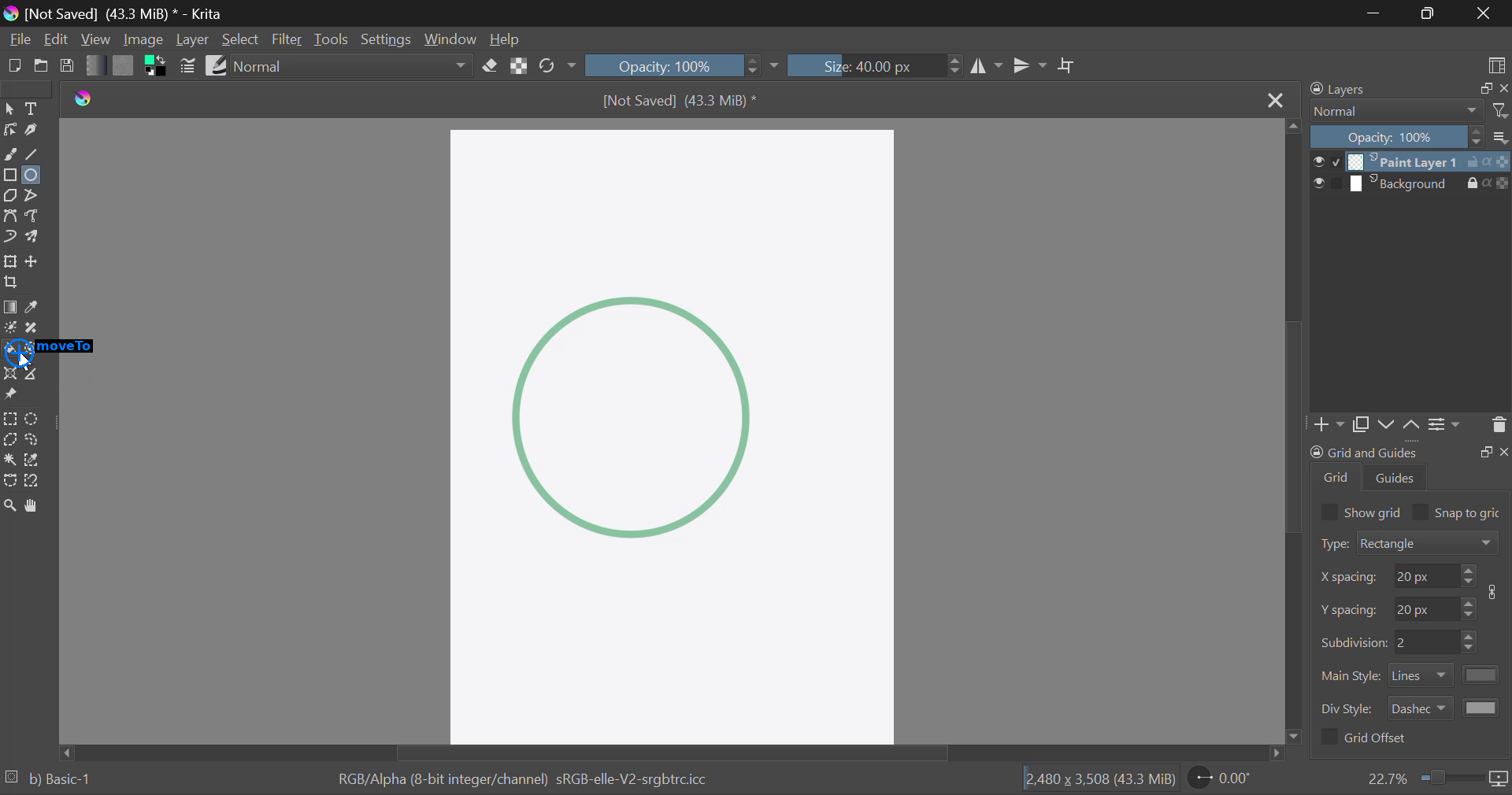}
\end{center}

\noindent\textbf{Observation:} The Krita application is currently open with a white canvas displaying a green circle outline in the center. The circle has a green stroke but no fill color yet. The left vertical toolbar contains various drawing and editing tools arranged vertically. From top to bottom, there are selection tools, shape tools, painting tools, and various other editing tools. Near the middle-lower section of the toolbar, there appears to be a fill bucket tool and other color-related tools. The right side panel shows the Layers panel with "Paint Layer 1" and "Background" layers visible, and below that is a "Grid and Guides" section. The top menu bar contains File, Edit, View, Image, Layer, Select, Filter, Tools, Settings, Window, and Help options. The current blending mode is set to "Normal" and opacity is at 100\%.

\vspace{0.3em}

\noindent\textbf{Thought:} The circle has been successfully drawn and finalized with a green stroke outline visible on the canvas. Now I need to fill the interior of this circle with green color before adding the inner shadow effect with Multiply blending mode. To accomplish this, I need to select the fill bucket tool from the left toolbar. Looking at the toolbar structure, I can see various tools arranged vertically. The fill bucket tool is typically located in the middle-lower section of the toolbar among the painting and color tools. Once I select the fill tool, I'll be able to click inside the circle to fill it with green color, which appears to already be set as the current color based on the green stroke. After filling the circle, I can then proceed to add the inner shadow effect and configure its blending mode to Multiply as required by the task.

\vspace{0.3em}

\noindent\textbf{Reflection:} Comparing the two full screenshots, there is no visible change in the interface. The circle with green stroke remains on the canvas, and the toolbar on the left appears unchanged. The action intended to move the mouse to the fill bucket tool in the left vertical toolbar. However, examining the coordinate (0.0125, 0.4436) and the position of the red circle in the first screenshot's image patch, the mouse appears to be positioned at the very edge of the toolbar, possibly outside the clickable area of any tool. The fill bucket tool should be more centrally positioned within the toolbar. Since no tool appears to be selected or highlighted, and there's no visual change indicating successful navigation to the fill bucket tool, this action is incorrect. The coordinates used were too far to the left edge, missing the actual tool buttons in the toolbar.

\vspace{0.8em}
\noindent\rule{\textwidth}{0.4pt}

\noindent\textbf{Step 7} \hfill {\small\texttt{pyautogui.click(x=0.0109, y=0.4416)}}

\vspace{0.3em}

\noindent
\begin{center}
  \includegraphics[width=\textwidth]{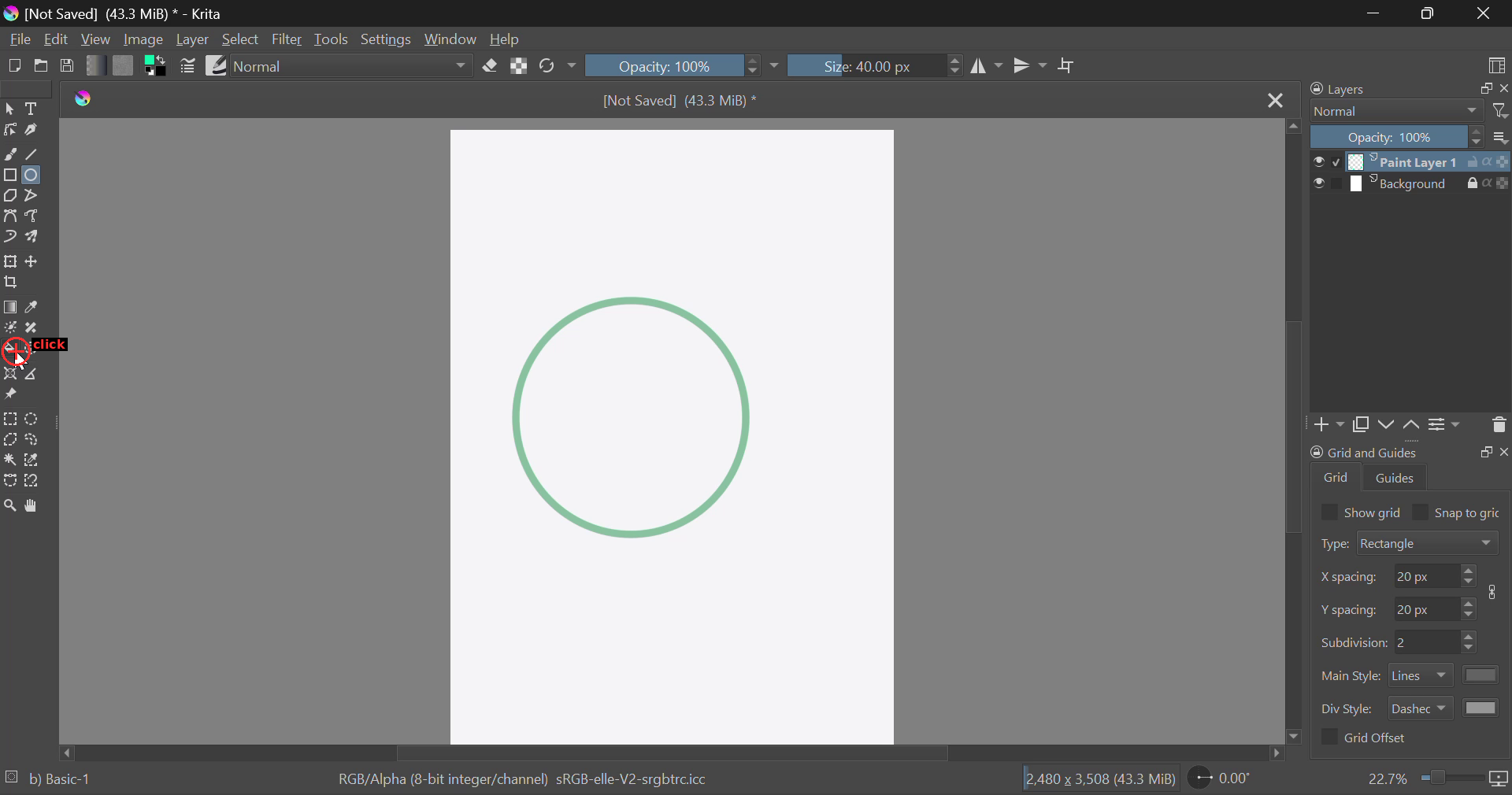}
\end{center}

\noindent\textbf{Observation:} The Krita application is open with a white canvas displaying a green circle outline in the center. The circle has been drawn but not yet filled with color. On the left side of the screen is a vertical toolbar containing various drawing and editing tools. The toolbar includes multiple icons arranged vertically, with painting and color-related tools visible in the middle-lower section. Near the bottom portion of the visible toolbar, there are several tool icons including what appears to be fill and color manipulation tools. The layers panel on the right shows "Paint Layer 1" is active with "Background" layer below it. The document title shows "[Not Saved] (43.3 MiB)" indicating unsaved work. The canvas shows only the green stroke outline of the circle with an unfilled white interior.

\vspace{0.3em}

\noindent\textbf{Thought:} The previous attempt to select the fill bucket tool failed because the click position was too far at the left edge of the toolbar, essentially at the border of the interface where no actual tool button exists. The coordinates were at the extreme edge, missing the clickable area of the tool icons themselves. I need to click more centrally within the actual tool button area to properly select it.

The task progress so far: I've successfully drawn a circle with a green outline. Now I need to fill this circle with green color before adding the inner shadow effect with Multiply blending mode. The green color appears to already be set based on the stroke color. To fill the circle, I need to first select the fill bucket tool from the toolbar.

Looking at the toolbar structure, the fill bucket tool should be located in the middle-lower section among the painting and color tools. I need to click on this tool with better positioning - more toward the center of the tool icon rather than at the extreme left edge. This will ensure I actually activate the fill bucket tool. Once selected, I can then click inside the circle to fill it with green, and subsequently add the inner shadow effect with the required Multiply blending mode.

\vspace{0.3em}

\noindent\textbf{Reflection:} Comparing the two full screenshots, there is no visible change in the interface. The circle with green outline remains on the canvas, and the toolbar on the left appears unchanged. The fill bucket tool does not appear to be selected - there's no visual indication that the tool has been activated. Looking at the action coordinates (x=0.0109, y=0.4416), this position is extremely close to the left edge of the screen, essentially at the border of the interface. The image patch shows the toolbar area, and the red circle indicator appears to be positioned at the very edge, not on any actual tool button. This explains why the click failed - the coordinates are too far to the left, missing the clickable area of the fill bucket tool icon. The agent needs to click more centrally within the actual tool button area, further to the right, to properly select the fill bucket tool.

\vspace{0.8em}
\noindent\rule{\textwidth}{0.4pt}

\noindent\textbf{Step 8} \hfill {\small\texttt{pyautogui.moveTo(x=0.3693, y=0.5158)}}

\vspace{0.3em}

\noindent
\begin{center}
  \includegraphics[width=\textwidth]{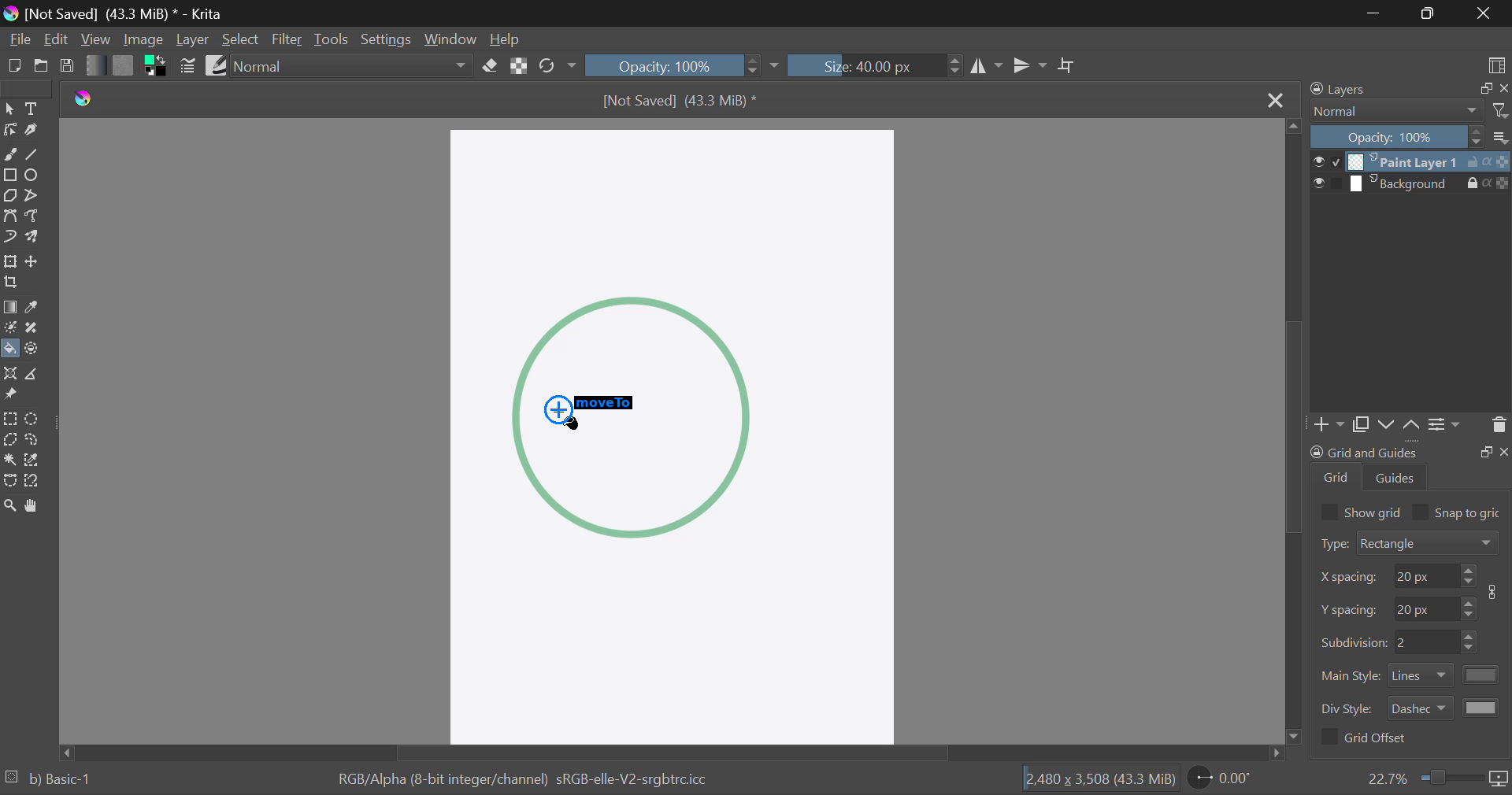}
\end{center}

\noindent\textbf{Observation:} The Krita application is open with a white canvas displaying a green circle outline in the center. The circle has a green stroke but no fill. On the left side, there is a vertical toolbar containing various drawing and editing tools arranged vertically, including selection tools, brush tools, shape tools, and color tools. The Layers panel on the right shows "Paint Layer 1" is active. The current blending mode is set to "Normal" with 100\% opacity. The top toolbar shows brush settings with "Normal" mode, 100\% opacity, and a size of 40.00 px. The title bar indicates the file is not saved with "[Not Saved] (43.3 MiB)". The canvas area shows the green circle positioned centrally on the white background.

\vspace{0.3em}

\noindent\textbf{Thought:} The previous click attempt failed because the coordinates were at the extreme left edge of the screen at x=0.0109, which is essentially at the border of the interface rather than on any actual tool button. The toolbar icons are positioned more centrally within the toolbar area, not at the very edge. This explains why the fill bucket tool wasn't activated.

I've successfully drawn a circle with a green outline. The next step in completing the task is to fill this circle with green color before adding the inner shadow effect with Multiply blending mode. To accomplish this, I need to select the fill bucket tool from the left toolbar. However, before clicking on the fill bucket tool, I should move the mouse to a better position. Looking at the task workflow, after selecting the fill bucket tool, I'll need to click inside the circle to fill it. The current action appears to be positioning the mouse inside the circle area, which would be the location where I'll click to apply the fill once the fill bucket tool is selected. This is a preparatory movement that positions the cursor in the interior of the circle, ready for the fill operation.

\vspace{0.3em}

\noindent\textbf{Reflection:} The visual changes between the two screenshots show that the mouse cursor has moved from the upper-left area inside the circle (where there was a red circle indicator with a crosshair) to a new position slightly lower and to the right, still within the interior of the green circle outline. The red circle indicator is no longer visible in the second screenshot, indicating the mouse has moved to the new position. This step is necessary and effective because after selecting the fill bucket tool in the previous step, the agent needs to position the cursor inside the circle where the fill will be applied. Moving the mouse to the interior of the circle prepares for the next action, which would be clicking to fill the circle with green color. This is a logical preparatory step in the workflow before applying the fill.

\vspace{0.8em}
\noindent\rule{\textwidth}{0.4pt}

\noindent\textbf{Step 9} \hfill {\small\texttt{pyautogui.click(x=0.3703, y=0.5178)}}

\vspace{0.3em}

\noindent
\begin{center}
  \includegraphics[width=\textwidth]{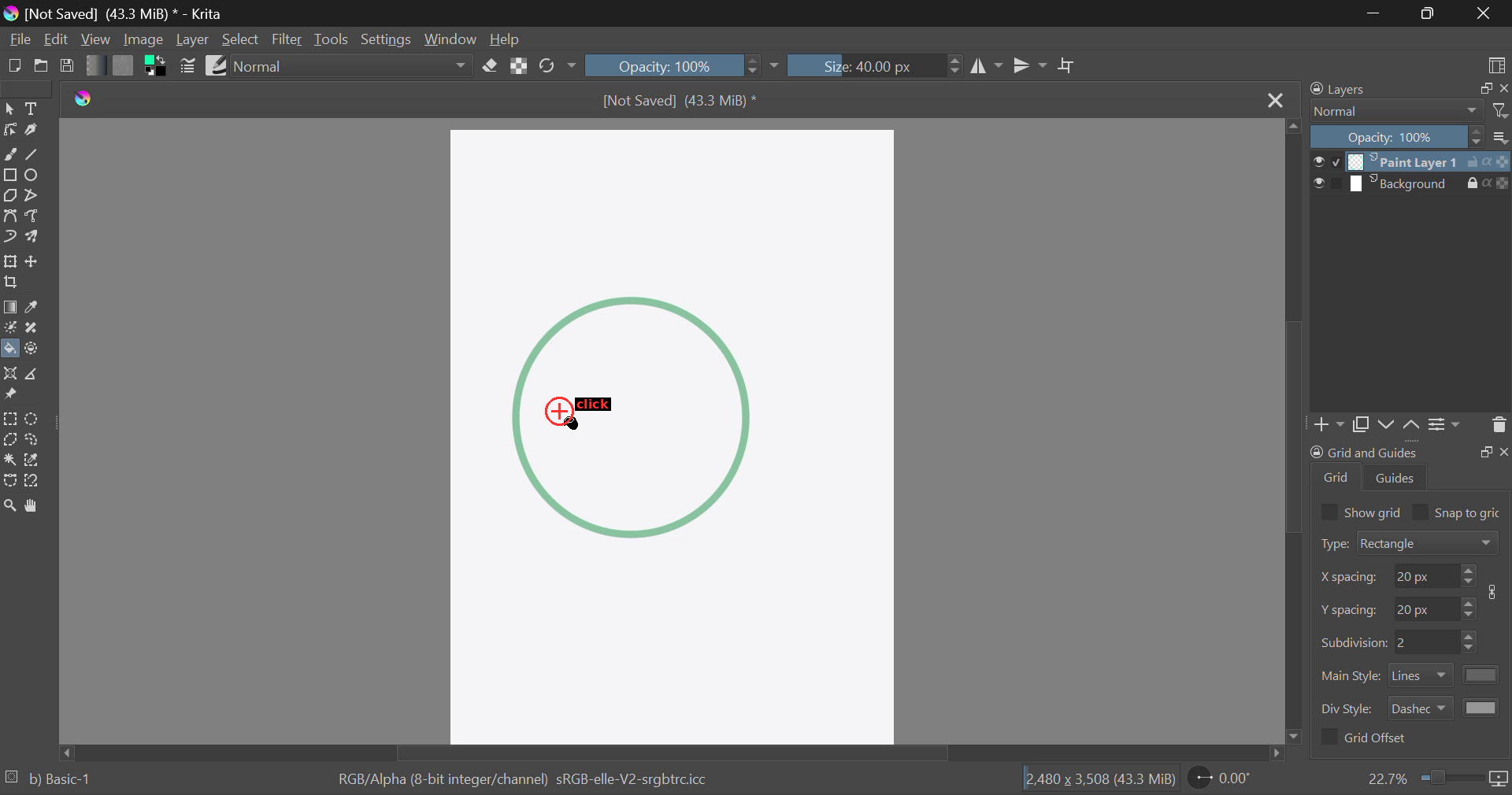}
\end{center}

\noindent\textbf{Observation:} The Krita application is open with a white canvas displayed in the center. On the canvas, there is a green circle outline positioned slightly left of center and above the vertical midpoint. The circle has a visible green stroke but no fill color yet. 

In the left vertical toolbar, various tools are visible including shape tools, selection tools, and painting tools. One of the tools in the lower section of the toolbar appears to be the fill bucket tool, which is used for filling enclosed areas with color.

The top toolbar shows "Normal" blending mode selected, with opacity at 100\% and a brush size of 40.00 px.

On the right side panel, the Layers panel shows two layers: "Paint Layer 1" (currently active/selected) and "Background" below it. The layer blending mode is set to "Normal" with 100\% opacity.

Below the Layers panel is a "Grid and Guides" section showing grid settings with Rectangle type, 20 px spacing for both X and Y, subdivision of 2, and Lines as the main style.

The interior of the green circle is currently empty/white, waiting to be filled with color.

\vspace{0.3em}

\noindent\textbf{Thought:} The cursor has been positioned inside the green circle outline in the upper-left interior area. The fill bucket tool should be active based on the previous action of selecting it from the toolbar. 

The task requires me to: 1) draw a circle (completed), 2) fill it with green color (in progress), and 3) add an inner shadow with Multiply blending mode (not yet started).

Progress so far:
- Successfully drew a circle with a green outline on the canvas
- Selected the fill bucket tool from the toolbar
- Positioned the cursor inside the circle's interior

The next logical step is to click inside the circle to apply the fill. When using the fill bucket tool, clicking inside a bounded area fills that area with the current foreground color. Since the circle outline is green, the foreground color should be green, and clicking inside will fill the circle's interior with green.

This click action is necessary because:
1. The fill bucket tool requires a click action to apply the fill
2. The cursor is positioned within the bounded area of the circle
3. This directly accomplishes the task requirement to "fill the circle with the color green"
4. After filling, I can proceed to the final step of adding an inner shadow with Multiply blending mode

Once I click, the circle's interior should become filled with green color, completing the filling portion of the task.

\vspace{0.3em}

\noindent\textbf{Reflection:} The visual change between the two screenshots is significant: the circle that previously had only a green outline is now completely filled with a solid green color. The fill bucket tool successfully applied the green fill to the interior of the circle. This step was necessary and effective because it accomplished the second requirement of the task - filling the circle with green color. The action correctly clicked inside the bounded area of the circle with the fill bucket tool active, which resulted in the expected fill operation. The task can now proceed to the final step of adding an inner shadow with Multiply blending mode.

\vspace{0.8em}
\noindent\rule{\textwidth}{0.4pt}

\noindent\textbf{Step 10} \hfill {\small\texttt{pyautogui.moveTo(x=0.9135, y=0.2079)}}

\vspace{0.3em}

\noindent
\begin{center}
  \includegraphics[width=\textwidth]{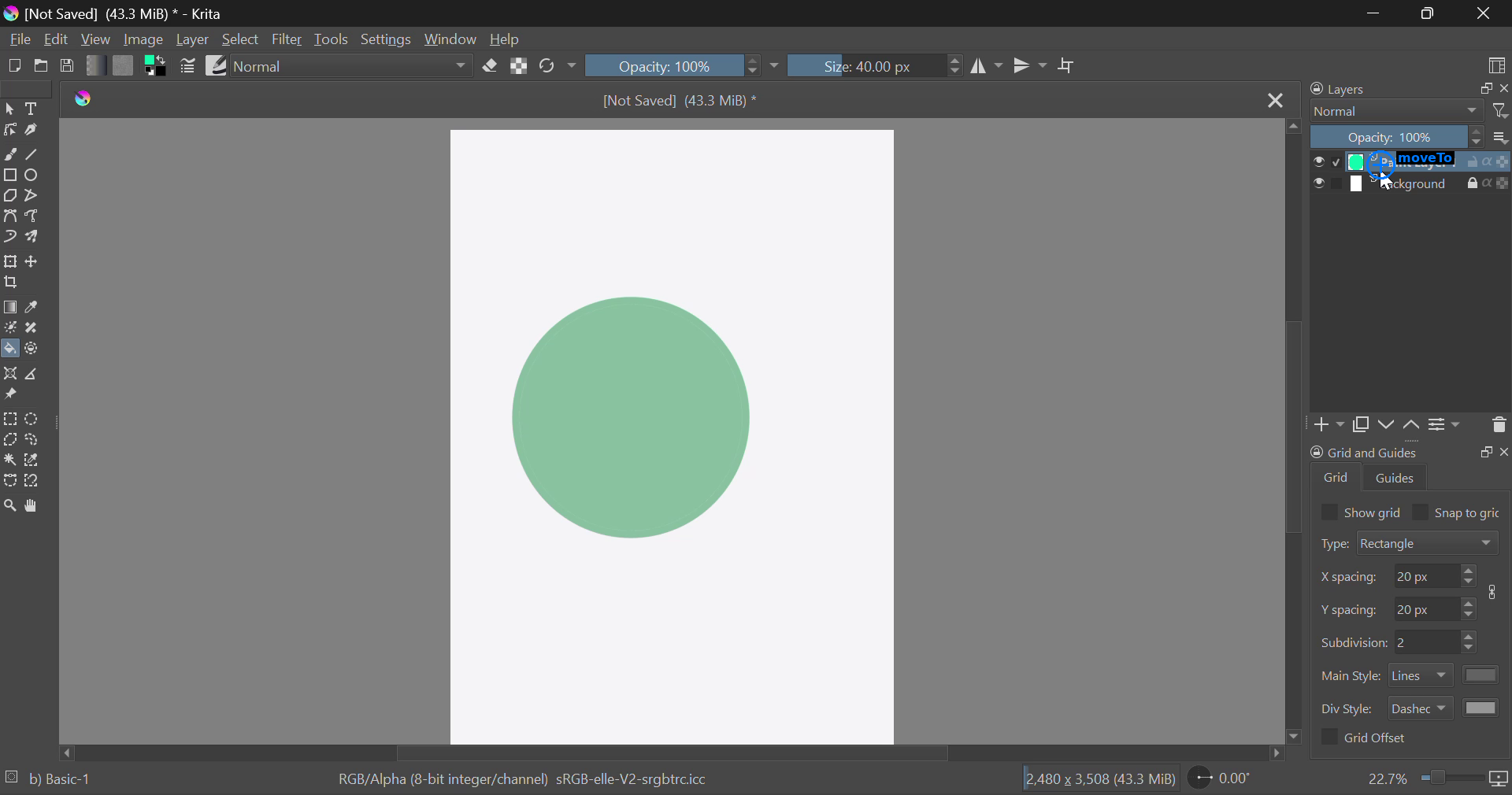}
\end{center}

\noindent\textbf{Observation:} The Krita application is open with a canvas displaying a green filled circle in the center. The right side panel shows the Layers docker, which contains two layers: "Paint Layer 1" (currently selected, indicated by blue highlighting) and "Background" below it. The "Paint Layer 1" has a small thumbnail showing the green circle. Above the layers list, there is a blending mode dropdown currently set to "Normal" and an opacity slider set to "100\%". The layer panel shows visibility icons (eye icons) next to each layer, indicating both layers are visible. At the top of the Layers docker, there is a close button (X) in the upper left corner. The layers panel header displays "Layers" with various control icons to the right.

\vspace{0.3em}

\noindent\textbf{Thought:} The circle has been successfully drawn and filled with green color. The green circle is on "Paint Layer 1" which is currently selected in the Layers panel. To add an inner shadow with Multiply blending mode, I need to access layer effects or layer styles. In Krita, layer effects are typically accessed through the Layer menu or by right-clicking on the layer.

The progress so far:
1. Circle drawn - completed
2. Filled with green color - completed
3. Add inner shadow with Multiply blending mode - not yet started

To add an inner shadow effect, I should right-click on the current layer to access layer properties and effects. Looking at the Layers panel on the right, "Paint Layer 1" is selected and visible. The most logical next step would be to right-click on this layer to access a context menu that should contain options for layer effects or layer styles, where I can add an inner shadow and set its blending mode to Multiply. This is the standard workflow in Krita for applying layer effects. After right-clicking, I expect a context menu to appear with various layer options including layer styles or effects.

\vspace{0.3em}

\noindent\textbf{Reflection:} Comparing the two full screenshots, there is no visible change in the interface. The mouse cursor has been moved to hover over 'Paint Layer 1' in the Layers panel, as indicated by the red circle marker in the cropped image showing the layer entry. This is a preparatory step for the next action, which would typically be a right-click to access layer properties or effects. The step is correct because it positions the mouse on the correct layer ('Paint Layer 1' which contains the green circle) in preparation for accessing layer effects to add an inner shadow with Multiply blending mode. This is a necessary intermediate step in the workflow to add layer effects in Krita, as you need to select or interact with the specific layer before applying effects to it.

\vspace{0.8em}
\noindent\rule{\textwidth}{0.4pt}

\noindent\textbf{Step 11} \hfill {\small\texttt{pyautogui.click(x=0.9146, y=0.2079)}}

\vspace{0.3em}

\noindent
\begin{center}
  \includegraphics[width=\textwidth]{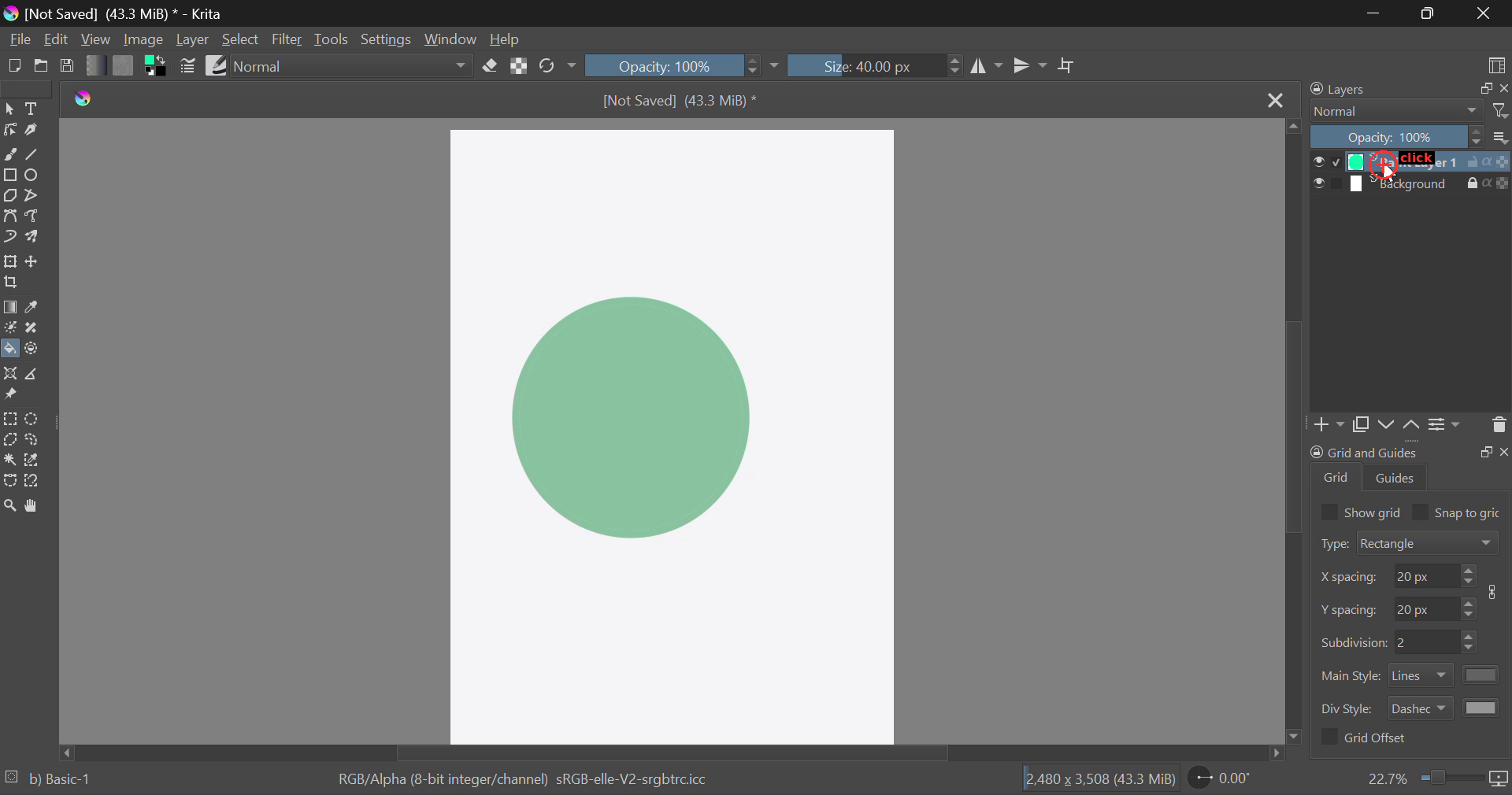}
\end{center}

\noindent\textbf{Observation:} The Krita application is open with a white canvas displaying a green filled circle in the center. The interface shows:

- **Top menu bar**: File, Edit, View, Image, Layer, Select, Filter, Tools, Settings, Window, Help
- **Left toolbar**: Various drawing and editing tools arranged vertically
- **Top toolbar**: Brush settings including "Normal" blending mode, "Opacity: 100\%", and "Size: 40.00 px"
- **Canvas area**: White background with a green circle drawn and filled in the center
- **Right panel - Layers section**: 
  - "Layers" tab is active at the top
  - Blending mode dropdown showing "Normal"
  - Opacity slider set to "100\%"
  - Two layers visible:
    - "Paint Layer 1" (currently selected/highlighted in blue) - contains a small green circle thumbnail icon
    - "Background" layer below it
  - Layer visibility icons (eye icons) are visible next to both layers
  - Various layer management icons at the bottom of the panel
- **Right panel - Grid and Guides section**: Below the Layers panel, showing grid configuration options including Rectangle type, 20 px spacing settings
- **Bottom status bar**: Shows "Basic-1", color mode "RGB/Alpha (8-bit integer/channel)", file format "sRGB-elle-V2-srgbtrc.icc", canvas dimensions "2,480 x 3,508 (43.3 MiB)", and zoom level "22.7\%"

The "Paint Layer 1" entry is highlighted in blue in the Layers panel, indicating it is the currently active layer.

\vspace{0.3em}

\noindent\textbf{Thought:} The previous action positioned the cursor over "Paint Layer 1" in the Layers panel. The green circle has been successfully drawn and filled with green color on this layer. Now I need to add an inner shadow effect with Multiply blending mode to complete the task.

Progress assessment:
1. Circle drawn - completed
2. Filled with green color - completed
3. Add inner shadow with Multiply blending mode - needs to be done

To add an inner shadow effect in Krita, I need to access layer effects or layer styles for "Paint Layer 1". The standard approach is to right-click on the layer entry in the Layers panel to open a context menu. This context menu typically contains options such as "Layer Properties", "Layer Style", or similar entries that provide access to layer effects.

Since "Paint Layer 1" is already selected and contains the green circle, I should right-click on this layer entry to access the context menu. From there, I can navigate to layer effects or layer styles where I can add an inner shadow and configure its blending mode to Multiply. This is the standard workflow in Krita for applying effects to layers.

After performing this right-click action, I expect a context menu to appear with various layer-related options, including access to layer effects where I can configure the inner shadow with the Multiply blending mode as required by the task.

\vspace{0.3em}

\noindent\textbf{Reflection:} The visual change between the two screenshots shows that a context menu has appeared after right-clicking on 'Paint Layer 1'. The menu displays various layer-related options including 'Properties...', 'Layer Style...', 'Copy Layer Style', 'Cut Layer', 'Copy Layer', 'Paste Layer', 'Remove Layer', 'Duplicate Layer or Mask', and other layer management options. The 'Layer Style...' option is visible in the menu, which is exactly what is needed to add an inner shadow effect with Multiply blending mode. This step is necessary and effective as it successfully opened the context menu that provides access to layer effects, which is the correct path to accomplish the task of adding an inner shadow to the green circle.

\vspace{0.8em}
\noindent\rule{\textwidth}{0.4pt}

\noindent\textbf{Step 12} \hfill {\small\texttt{pyautogui.moveTo(x=0.8484, y=0.2446)}}

\vspace{0.3em}

\noindent
\begin{center}
  \includegraphics[width=\textwidth]{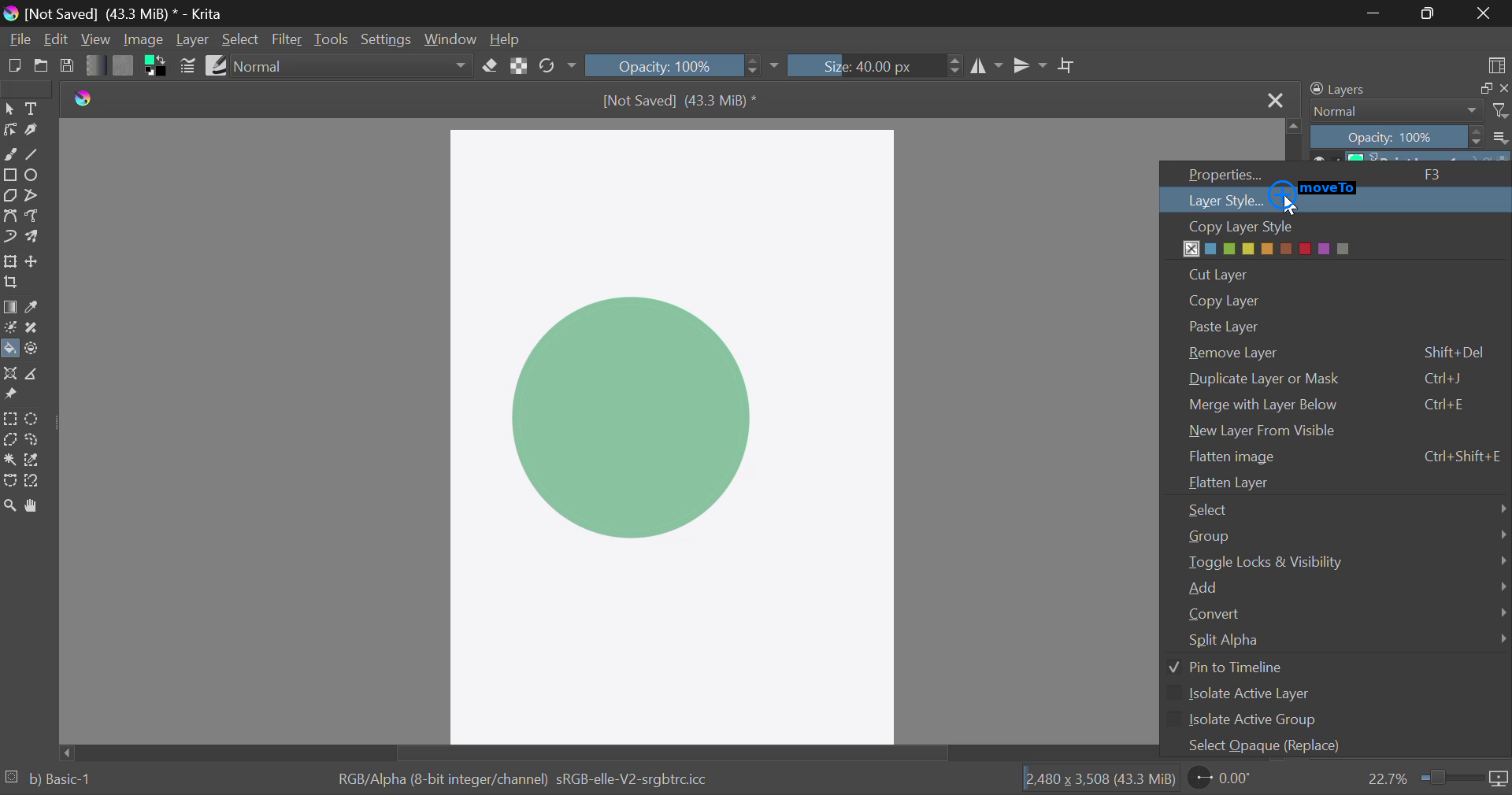}
\end{center}

\noindent\textbf{Observation:} The Krita application is currently open with a white canvas displaying a green-filled circle in the center. On the right side of the screen, a context menu has appeared overlaying the Layers panel. This context menu contains multiple options including "Properties..." at the top, followed by "Layer Style..." which is highlighted in blue. Below that are additional options such as "Copy Layer Style" with a row of colored squares, "Cut Layer", "Copy Layer", "Paste Layer", "Remove Layer", "Duplicate Layer or Mask", "Merge with Layer Below", "New Layer From Visible", "Flatten image", "Flatten Layer", "Select", "Group", "Toggle Locks \& Visibility", "Add", "Convert", "Split Alpha", "Pin to Timeline", "Isolate Active Layer", "Isolate Active Group", and "Select Opaque (Replace)". The Layers panel shows "Normal" blending mode and "Opacity: 100\%" settings at the top.

\vspace{0.3em}

\noindent\textbf{Thought:} The context menu has successfully appeared after the right-click action on "Paint Layer 1". I can see the "Layer Style..." option is available and highlighted in the menu. This is the correct pathway to access layer effects in Krita.

Assessing the task progress:
1. Circle drawn - completed
2. Filled with green color - completed  
3. Add inner shadow with Multiply blending mode - needs to be done

To add an inner shadow effect with Multiply blending mode, I need to access the layer style settings. The "Layer Style..." option in the context menu is specifically designed for this purpose. Clicking on this option will open a dialog where I can configure various layer effects including inner shadow, and set the blending mode to Multiply.

The next logical step is to move the mouse to the "Layer Style..." option in the context menu. This will position the cursor over this menu item in preparation for clicking it. After moving to this option, I expect to be able to click it in the subsequent action, which will open the Layer Style dialog where I can add and configure the inner shadow effect with the required Multiply blending mode.

\vspace{0.3em}

\noindent\textbf{Reflection:} The visual comparison between the two screenshots shows no significant change, which is expected for a mouse movement action. The context menu remains open with the same options visible, and the mouse cursor has been repositioned over the 'Layer Style...' option as indicated by the action. This is a necessary preparatory step before clicking on the 'Layer Style...' option to open the layer style dialog where the inner shadow effect with Multiply blending mode can be configured. The action correctly moves the mouse to the intended menu item, positioning it for the next click action that will open the layer style settings dialog.

\vspace{0.8em}
\noindent\rule{\textwidth}{0.4pt}

\noindent\textbf{Step 13} \hfill {\small\texttt{pyautogui.click(x=0.8484, y=0.2465)}}

\vspace{0.3em}

\noindent
\begin{center}
  \includegraphics[width=\textwidth]{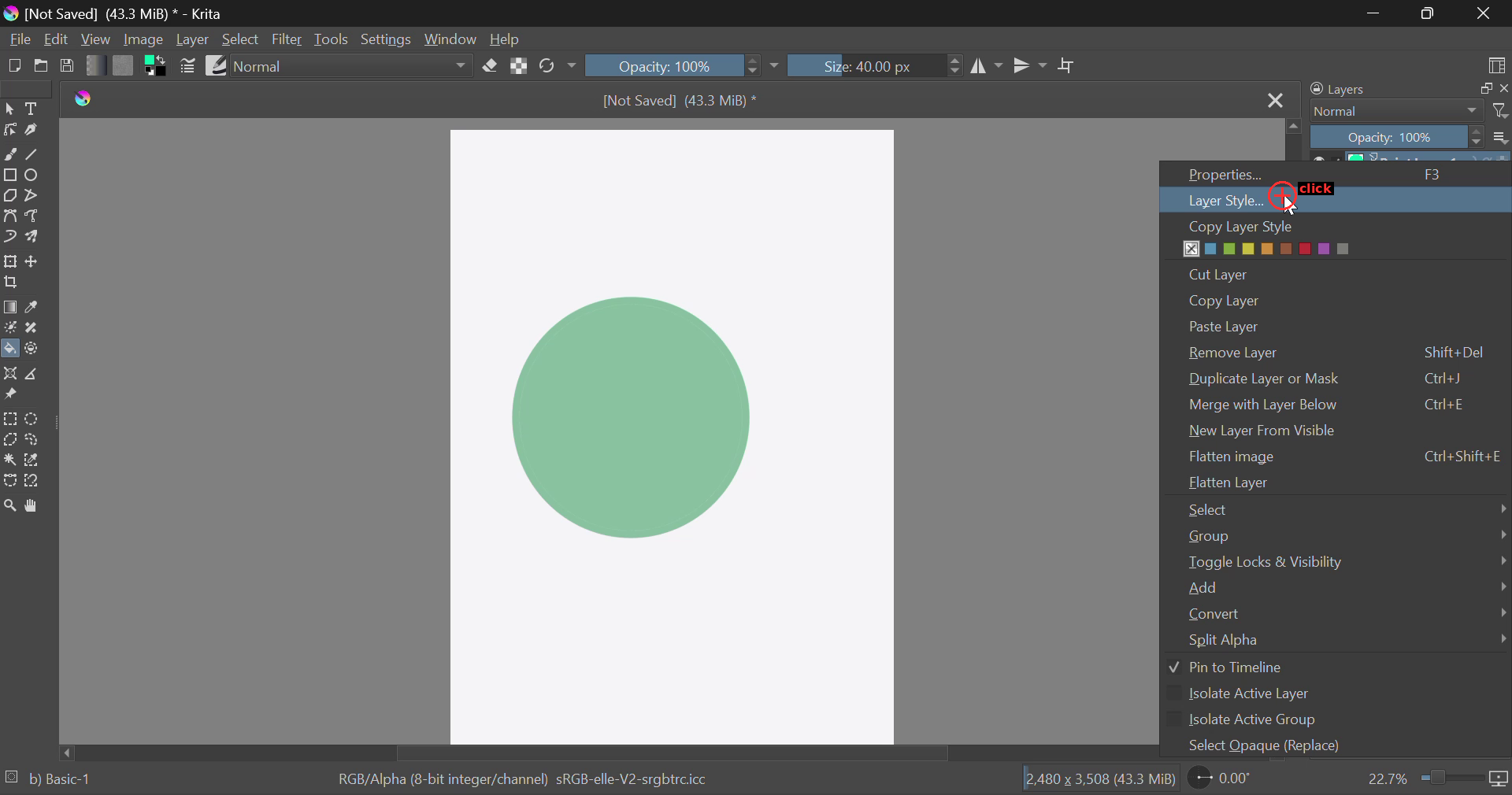}
\end{center}

\noindent\textbf{Observation:} The Krita application is currently open with a white canvas displaying a green filled circle in the center. On the right side of the screen, a context menu has appeared with various layer-related options. The menu includes options such as "Properties...", "Layer Style...", "Copy Layer Style", "Cut Layer", "Copy Layer", "Paste Layer", "Remove Layer", "Duplicate Layer or Mask", "Merge with Layer Below", "New Layer From Visible", "Flatten image", "Flatten Layer", "Select", "Group", "Toggle Locks \& Visibility", "Add", "Convert", "Split Alpha", "Pin to Timeline", "Isolate Active Layer", "Isolate Active Group", and "Select Opaque (Replace)". The "Layer Style..." option appears in blue highlighting near the top of this context menu. The Layers panel is visible on the right side showing "Paint Layer 1" with Normal blending mode and 100\% opacity settings.

\vspace{0.3em}

\noindent\textbf{Thought:} The context menu has successfully appeared and the "Layer Style..." option is visible and accessible. Looking at the progress so far, I have completed drawing the circle and filling it with green color. Now I need to add an inner shadow effect with Multiply blending mode to complete the task. The "Layer Style..." option in the context menu is the correct path to access layer effects in Krita. By clicking on this option, a Layer Style dialog will open where I can enable the inner shadow effect and configure its blending mode to Multiply. This is the necessary next step to achieve the goal. Once I click on "Layer Style...", the dialog should appear with various effect options including inner shadow, and I'll be able to configure the blending mode and other parameters for the inner shadow effect.

\vspace{0.3em}

\noindent\textbf{Reflection:} The visual changes between the screenshots show that clicking on 'Layer Style...' successfully opened the Layer Styles dialog window. The dialog displays various layer effect options on the left side including 'Drop Shadow', 'Inner Shadow', 'Outer Glow', 'Inner Glow', 'Bevel and Emboss', 'Contour', 'Texture', 'Satin', 'Color Overlay', 'Gradient Overlay', 'Pattern Overlay', and 'Stroke'. The right side shows blending options and other settings. This is exactly the expected outcome - the Layer Styles dialog is now open and ready for configuring the inner shadow effect. The step was necessary and effective as it provides access to the inner shadow effect that needs to be applied to complete the task. The next logical step would be to enable the 'Inner Shadow' option and configure its blending mode to 'Multiply'.

\vspace{0.8em}
\noindent\rule{\textwidth}{0.4pt}

\noindent\textbf{Step 14} \hfill {\small\texttt{pyautogui.moveTo(x=0.2969, y=0.2099)}}

\vspace{0.3em}

\noindent
\begin{center}
  \includegraphics[width=\textwidth]{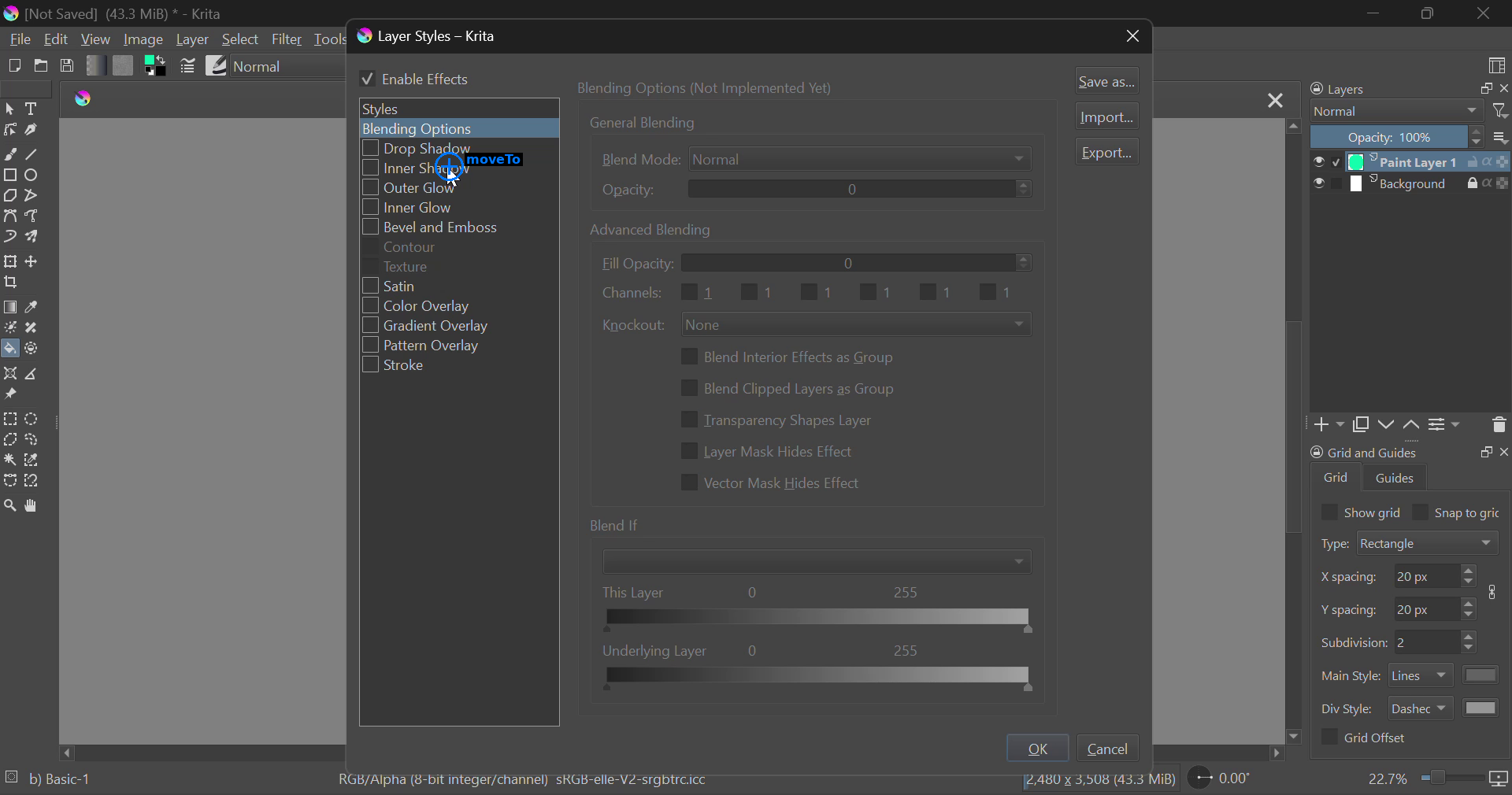}
\end{center}

\noindent\textbf{Observation:} The Layer Styles dialog window is currently open in Krita. The dialog displays a list of layer effects on the left side, with "Enable Effects" checkbox checked at the top. The effects list includes "Styles" and "Blending Options" at the top, followed by various effect options: "Drop Shadow", "Inner Shadow", "Outer Glow", "Inner Glow", "Bevel and Emboss", "Contour", "Texture", "Satin", "Color Overlay", "Gradient Overlay", "Pattern Overlay", and "Stroke". Each effect has an unchecked checkbox next to it.

The "Blending Options" item is currently highlighted in blue. The right side of the dialog shows blending settings including "General Blending" with a "Blend Mode" dropdown set to "Normal" and an "Opacity" slider set to 0. Below that is an "Advanced Blending" section with "Fill Opacity" slider, "Channels" checkboxes, "Knockout" dropdown set to "None", and several checkboxes for options like "Blend Interior Effects as Group", "Blend Clipped Layers as Group", "Transparency Shapes Layer", "Layer Mask Hides Effect", and "Vector Mask Hides Effect". At the bottom are "Blend If" sliders and "OK" and "Cancel" buttons.

\vspace{0.3em}

\noindent\textbf{Thought:} The Layer Styles dialog has opened successfully, showing all available layer effects. I can see the complete list of effects with "Inner Shadow" visible in the list. Currently, "Blending Options" is selected and displayed on the right side, but none of the specific effects are enabled as all checkboxes are unchecked.

To complete the task of adding an inner shadow with Multiply blending mode to the green circle, I need to enable the "Inner Shadow" effect and configure its blending mode. The "Inner Shadow" option is located in the effects list, positioned as the second effect below "Drop Shadow" and above "Outer Glow".

The next logical step is to click on the "Inner Shadow" text in the effects list. This will allow me to access the Inner Shadow settings panel where I can enable the effect and configure its blending mode to Multiply. Clicking on the effect name typically selects it and displays its specific settings on the right side of the dialog, where I'll be able to find the blending mode option and change it from the default to Multiply.

\vspace{0.3em}

\noindent\textbf{Reflection:} Comparing the two screenshots, there is no visible change in the interface. The Layer Styles dialog remains open with the same view - 'Blending Options' is still selected and displayed on the right side, and all effect checkboxes remain unchecked. The mouse cursor has moved to hover over the 'Inner Shadow' text in the effects list, as indicated by the red circle position in the image patch. This is a necessary step because moving the mouse to 'Inner Shadow' is the prerequisite action before clicking on it to enable and configure the inner shadow effect. The lack of visual change is expected since this is just a mouse movement action, not a click. The step is correct as it positions the cursor exactly where it needs to be for the next action (clicking on 'Inner Shadow'), which will then allow access to the Inner Shadow settings where the Multiply blending mode can be configured.

\vspace{0.8em}
\noindent\rule{\textwidth}{0.4pt}

\noindent\textbf{Step 15} \hfill {\small\texttt{pyautogui.click(x=0.2969, y=0.2139)}}

\vspace{0.3em}

\noindent
\begin{center}
  \includegraphics[width=\textwidth]{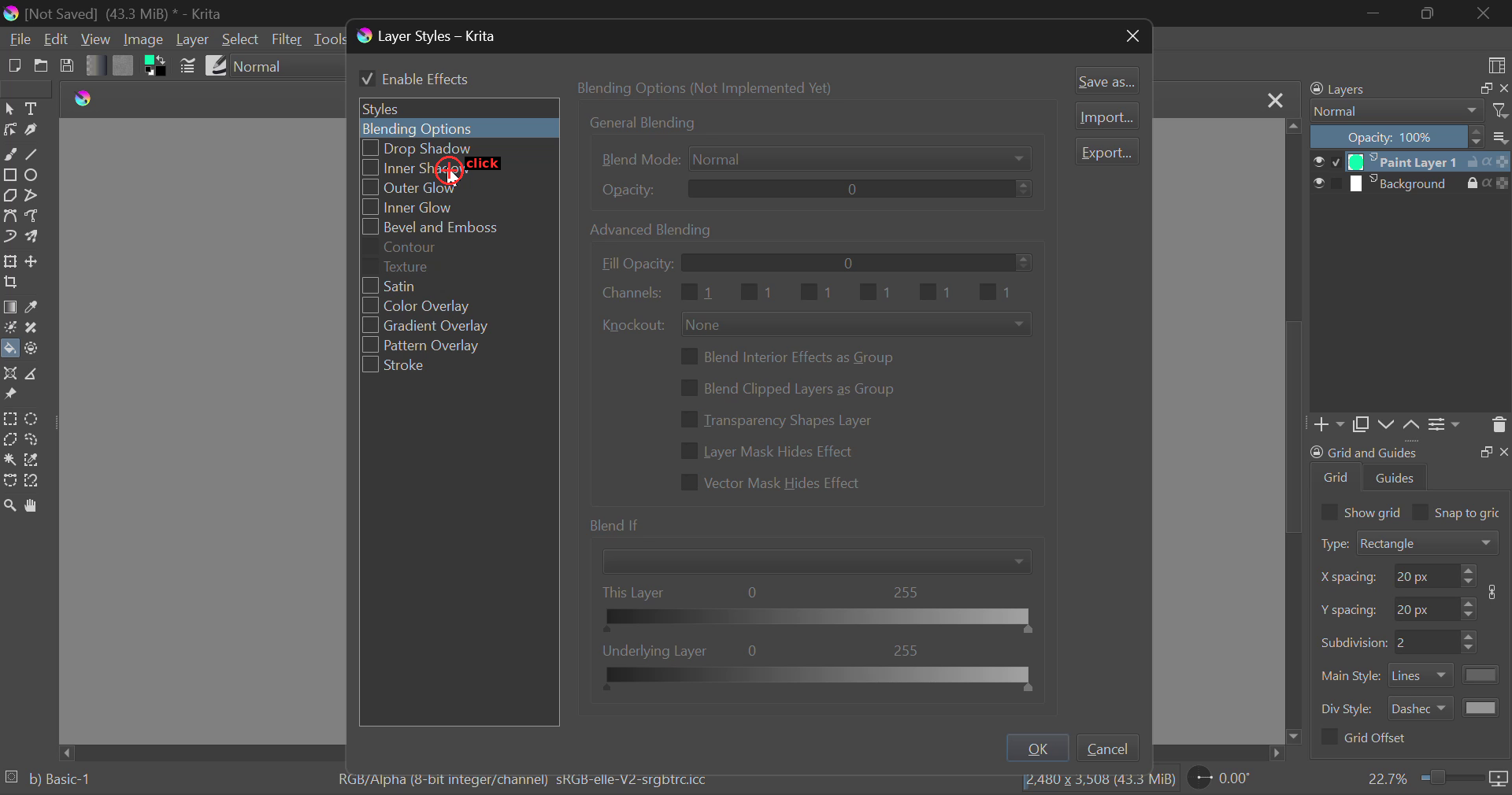}
\end{center}

\noindent\textbf{Observation:} The Layer Styles dialog window is currently open in Krita, displaying the effects configuration panel. The dialog shows a list of available layer effects on the left side, including "Blending Options" (currently selected and highlighted in blue), "Drop Shadow", "Inner Shadow", "Outer Glow", "Inner Glow", "Bevel and Emboss", "Contour", "Texture", "Satin", "Color Overlay", "Gradient Overlay", "Pattern Overlay", and "Stroke". All effect checkboxes are currently unchecked.

On the right side of the dialog, the "Blending Options (Not Implemented Yet)" panel is displayed, showing various blending settings including General Blending section with Blend Mode set to "Normal" and Opacity at 0, Advanced Blending section with Fill Opacity at 0, Channels settings, Knockout set to "None", and several checkboxes for blending options. The Blend If section shows "This Layer" and "Underlying Layer" sliders.

At the top of the dialog, there's a checked "Enable Effects" checkbox. The dialog has "Save as...", "Import...", and "Export..." buttons in the upper right, along with a close button. At the bottom right are "OK" and "Cancel" buttons.

In the main Krita interface behind the dialog, the Layers panel on the right shows "Paint Layer 1" is selected, and the canvas area shows a green filled circle that was previously created.

\vspace{0.3em}

\noindent\textbf{Thought:} The Layer Styles dialog is open and displaying the Blending Options panel. I have successfully navigated to this dialog after creating and filling a green circle. Now I need to enable and configure the Inner Shadow effect with Multiply blending mode to complete the task.

Looking at the current state, the task progress includes:
1. Circle shape drawn successfully
2. Circle filled with green color
3. Layer Styles dialog opened and visible
4. Currently viewing Blending Options

To add an inner shadow with Multiply blending mode, I need to access the Inner Shadow effect settings. The "Inner Shadow" option is visible in the effects list on the left side of the dialog, positioned below "Drop Shadow" and above "Outer Glow". 

The most logical next action is to click on "Inner Shadow" in the effects list. This will select the Inner Shadow effect and display its configuration panel on the right side of the dialog, replacing the current Blending Options view. Once the Inner Shadow settings panel appears, I will be able to access the blending mode dropdown menu and change it to "Multiply" as required.

This action is necessary because I cannot configure the Inner Shadow's blending mode without first selecting the effect and accessing its settings panel. After clicking on "Inner Shadow", I expect the right panel to update and show Inner Shadow-specific controls including structure, quality, and blending mode options.

\vspace{0.3em}

\noindent\textbf{Reflection:} The visual changes between the two screenshots show that clicking on 'Inner Shadow' in the effects list successfully changed the right panel from 'Blending Options' to 'Inner Shadow' settings. The right panel now displays Inner Shadow-specific controls including Structure settings (Blend Mode: Multiply, Opacity: 75\%, Angle: 120$^{\circ}$, Distance: 21 px, Choke: 0\%, Size: 21 px) and Quality settings (Contour, Anti-aliased, Noise: 0\%). The 'Inner Shadow' item in the left effects list is now highlighted in blue, indicating it is selected. This step is necessary and effective because it provides access to the Inner Shadow configuration panel where the blending mode can be verified or adjusted. Notably, the Blend Mode is already set to 'Multiply' as required by the task, which means the effect is properly configured.

\vspace{0.8em}
\noindent\rule{\textwidth}{0.4pt}

\noindent\textbf{Step 16} \hfill {\small\texttt{pyautogui.moveTo(x=0.2448, y=0.2099)}}

\vspace{0.3em}

\noindent
\begin{center}
  \includegraphics[width=\textwidth]{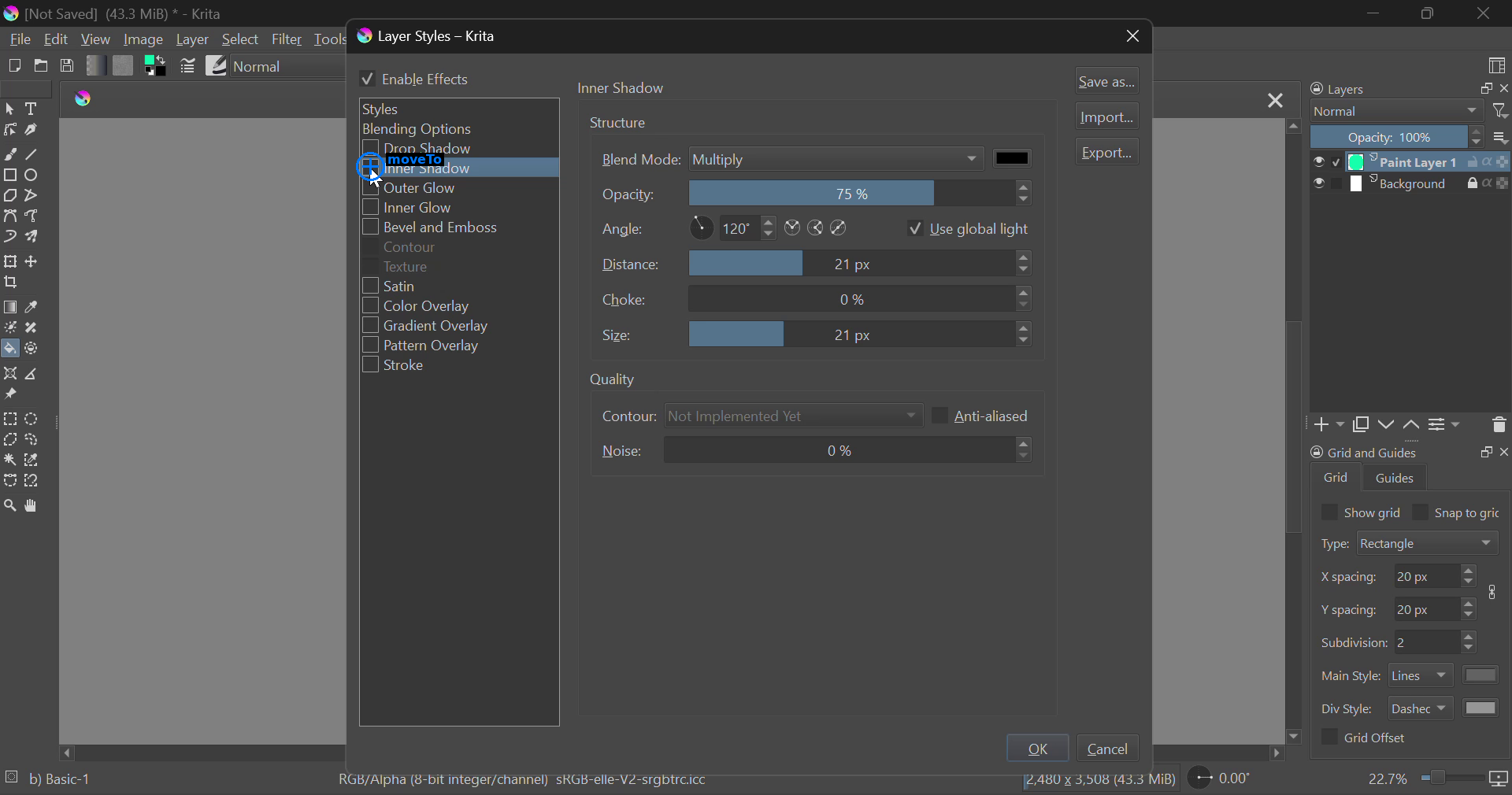}
\end{center}

\noindent\textbf{Observation:} The Krita application is open with the Layer Styles dialog window displayed in the foreground. The dialog shows a list of layer effects on the left side, including "Blending Options," "Drop Shadow," "Inner Shadow" (currently highlighted in blue), "Outer Glow," "Inner Glow," "Bevel and Emboss," "Contour," "Texture," "Satin," "Color Overlay," "Gradient Overlay," "Pattern Overlay," and "Stroke."

On the right side of the dialog, the Inner Shadow settings panel is visible, displaying:
- Structure section with Blend Mode set to "Multiply," Opacity at 75\%, Angle at 120$^{\circ}$, Distance at 21 px, Choke at 0\%, and Size at 21 px
- Quality section with Contour set to "Not Implemented Yet," Anti-aliased checkbox, and Noise at 0\%

There is a checkbox next to "Inner Shadow" in the effects list on the left. The "Enable Effects" checkbox at the top of the dialog is checked. At the bottom of the dialog are "OK" and "Cancel" buttons. In the top-right corner are "Save as...," "Import...," and "Export..." buttons, along with a close button (X).

The main Krita window behind the dialog shows the canvas area and the Layers panel on the right, which displays "Paint Layer 1" and "Background" layers.

\vspace{0.3em}

\noindent\textbf{Thought:} The Inner Shadow effect has been successfully selected and its settings panel is now displayed. The Blend Mode is already set to "Multiply" as required by the task. I have completed drawing a circle shape, filling it with green color, opening the Layer Styles dialog, and selecting the Inner Shadow effect.

Looking at the current state, the Inner Shadow settings show the correct blend mode configuration. However, to ensure the effect is properly applied to the layer, I need to enable the Inner Shadow effect by checking its checkbox in the effects list. The checkbox is located to the left of the "Inner Shadow" text in the effects list panel.

The task progress so far:
1. Circle shape drawn successfully
2. Circle filled with green color
3. Layer Styles dialog opened
4. Inner Shadow effect selected with Multiply blend mode configured

To complete the task, I need to enable the Inner Shadow effect by clicking on its checkbox. This will activate the inner shadow with the Multiply blending mode on the green circle. After enabling the checkbox, I can then click "OK" to apply all changes and close the Layer Styles dialog.

The most logical next action is to click on the checkbox next to "Inner Shadow" in the effects list to enable the effect and ensure it is applied to the layer.

\vspace{0.3em}

\noindent\textbf{Reflection:} The visual changes between the two screenshots show that the mouse cursor has moved from the checkbox area to a position over the 'Inner Shadow' text in the effects list. The checkbox next to 'Inner Shadow' remains unchecked in both screenshots, indicating that the mouse movement action was executed but no click occurred yet. This step is necessary and effective as it positions the cursor correctly over the checkbox area in preparation for clicking to enable the Inner Shadow effect. The action aligns with the task requirement to add an inner shadow with Multiply blending mode to the green circle. The movement is a preparatory step before the actual click action that will enable the effect.

\vspace{0.8em}
\noindent\rule{\textwidth}{0.4pt}

\noindent\textbf{Step 17} \hfill {\small\texttt{pyautogui.click(x=0.2458, y=0.2119)}}

\vspace{0.3em}

\noindent
\begin{center}
  \includegraphics[width=\textwidth]{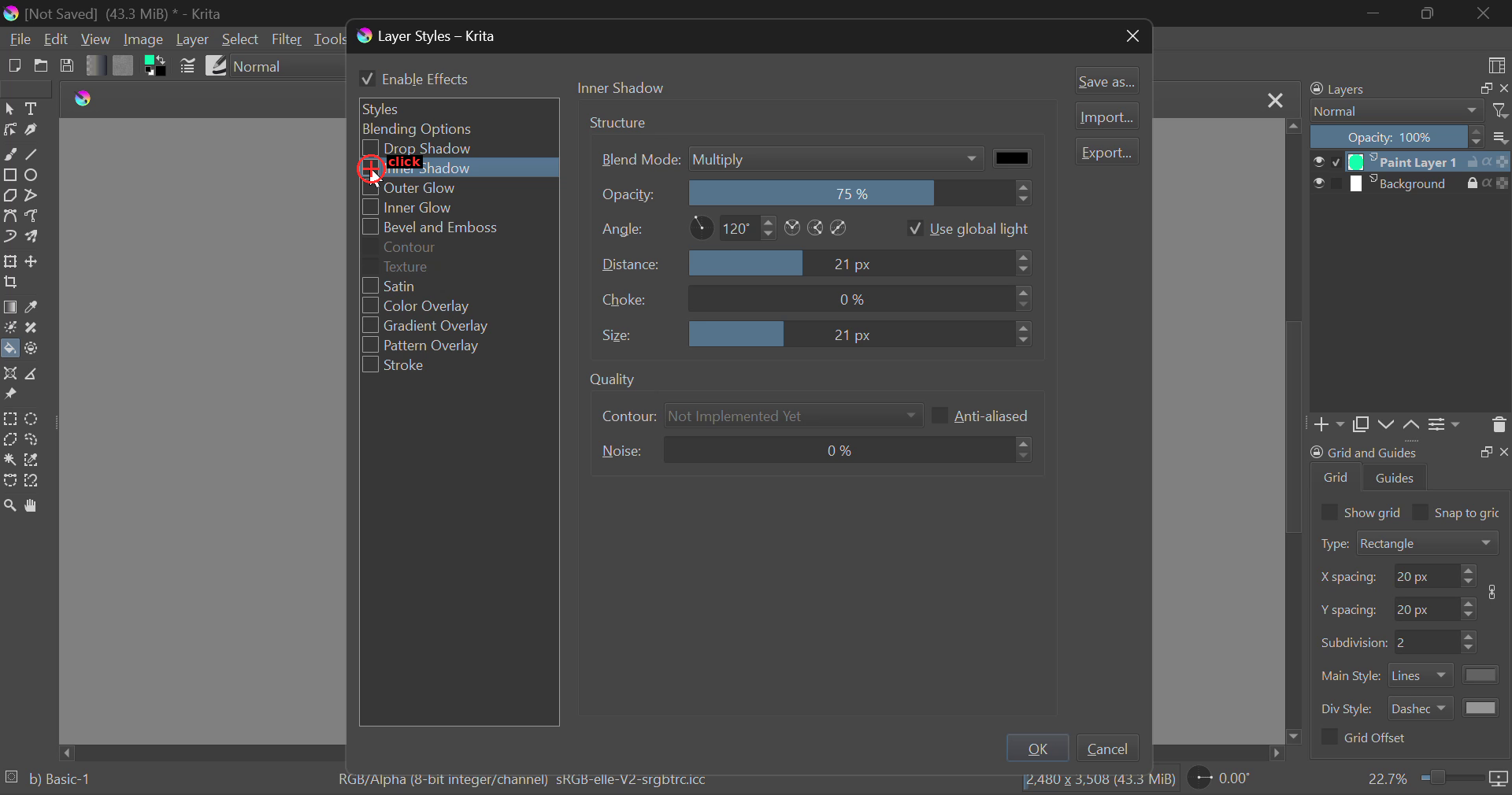}
\end{center}

\noindent\textbf{Observation:} The Krita application is open with the Layer Styles dialog window displayed in the foreground. The dialog shows a list of layer effects on the left side, including "Drop Shadow," "Inner Shadow," "Outer Glow," "Inner Glow," "Bevel and Emboss," and other options under categories like "Styles," "Blending Options," "Contour," and "Texture." The "Inner Shadow" effect is currently selected and highlighted in blue in the effects list. To the left of "Inner Shadow" is an unchecked checkbox, indicating that the effect is not yet enabled.

On the right side of the dialog, the Inner Shadow settings panel displays:
- Blend Mode: set to "Multiply"
- Opacity: 75\%
- Angle: 120$^{\circ}$
- Distance: 21 px
- Choke: 0\%
- Size: 21 px
- Quality settings including Contour dropdown showing "Not Implemented Yet" and an "Anti-aliased" checkbox
- Noise: 0\%

At the bottom of the dialog are "OK" and "Cancel" buttons. In the main Krita window behind the dialog, the Layers panel on the right shows "Paint Layer 1" is selected with 100\% opacity, and below it is the "Background" layer. The canvas area contains a green circle shape that was previously drawn and filled.

\vspace{0.3em}

\noindent\textbf{Thought:} The previous mouse movement positioned the cursor near the checkbox area for the Inner Shadow effect. The checkbox remains unchecked, which means the Inner Shadow effect is selected but not yet enabled on the layer.

Reviewing the task progress:
1. Circle shape has been successfully drawn
2. Circle has been filled with green color
3. Layer Styles dialog has been opened
4. Inner Shadow effect has been selected, and the Blend Mode is already configured to "Multiply" as required

The current state shows that the Inner Shadow effect is selected and its settings panel is visible with the correct Multiply blending mode. However, the checkbox next to "Inner Shadow" in the effects list is still unchecked, meaning the effect is not yet active on the layer.

To complete the task, I need to enable the Inner Shadow effect by clicking on the checkbox. This will activate the inner shadow with the Multiply blending mode on the green circle. The checkbox is positioned to the left of the "Inner Shadow" text in the effects list.

The logical next action is to click on the checkbox next to "Inner Shadow" to enable the effect. This will check the box and apply the inner shadow effect to the green circle layer. After this action, the checkbox should become checked, and the inner shadow effect will be active. Following this, I would need to click "OK" to apply all changes and close the Layer Styles dialog to complete the task.

\vspace{0.3em}

\noindent\textbf{Reflection:} The visual change between the two screenshots shows that the checkbox next to 'Inner Shadow' in the effects list has been successfully checked. In the first screenshot, the checkbox was unchecked (empty), and in the second screenshot, it now displays a checkmark. This indicates that the Inner Shadow effect has been enabled for the layer. The action was necessary and effective because it activated the Inner Shadow effect with the Multiply blending mode (which was already configured) on the green circle layer. This step directly contributes to completing the task requirement of adding an inner shadow using the Multiply blending mode to the green circle.

\vspace{0.8em}
\noindent\rule{\textwidth}{0.4pt}

\noindent\textbf{Step 18} \hfill {\small\texttt{pyautogui.moveTo(x=0.6776, y=0.9356)}}

\vspace{0.3em}

\noindent
\begin{center}
  \includegraphics[width=\textwidth]{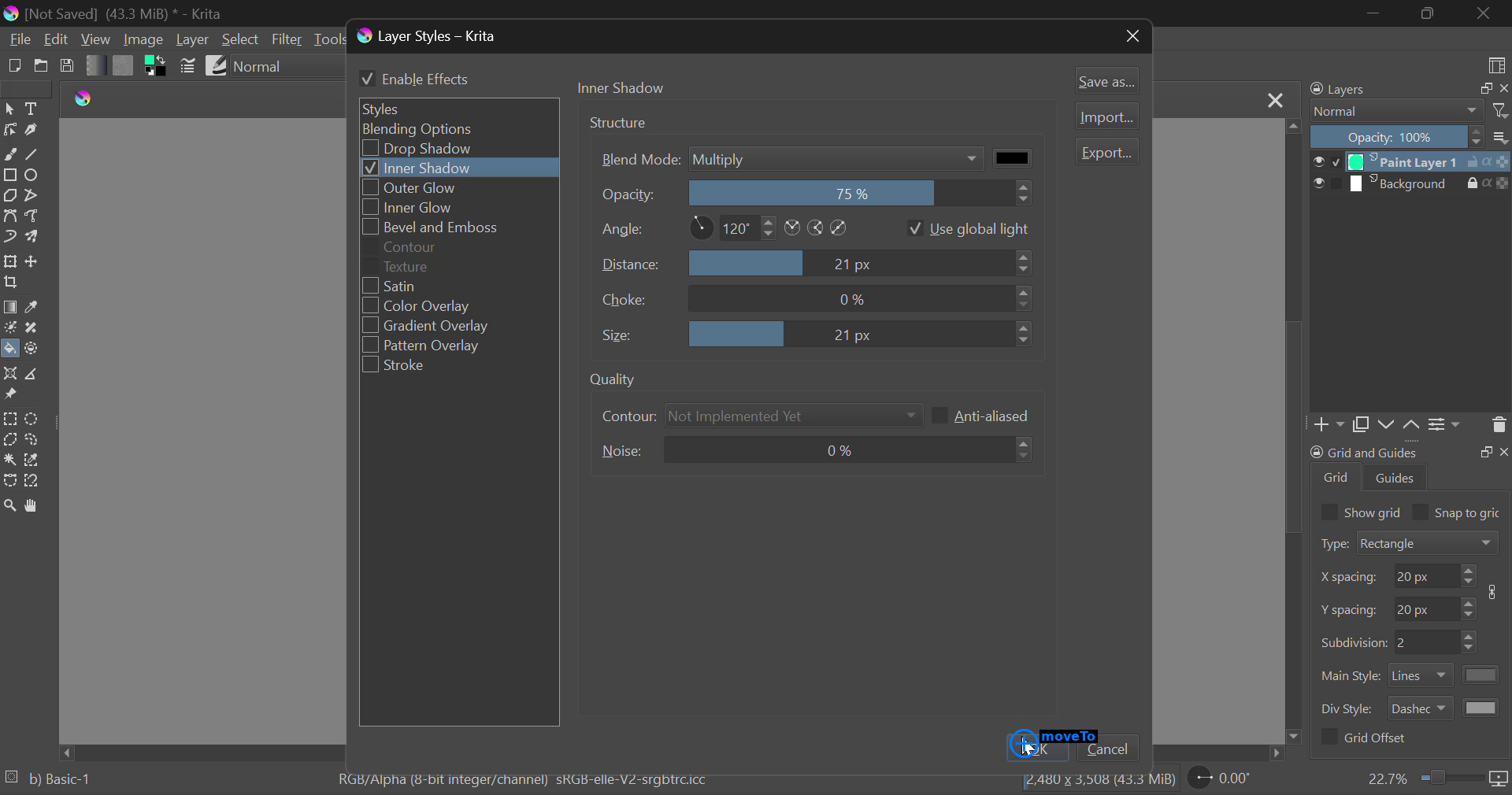}
\end{center}

\noindent\textbf{Observation:} The current state shows the Krita application with the Layer Styles dialog window open. The dialog is titled "Layer Styles --- Krita" and displays various layer effect options on the left side, including "Drop Shadow," "Inner Shadow" (which is currently checked and highlighted), "Outer Glow," "Inner Glow," "Bevel and Emboss," "Contour," "Texture," "Satin," "Color Overlay," "Gradient Overlay," "Pattern Overlay," and "Stroke."

The right side of the dialog shows the Inner Shadow settings panel with the following configuration:
- Structure section with Blend Mode set to "Multiply"
- Opacity at 75\%
- Angle at 120$^{\circ}$
- Distance at 21 px
- Choke at 0\%
- Size at 21 px
- Quality section with Contour set to "Not Implemented Yet" and Anti-aliased checkbox
- Noise at 0\%

At the bottom right of the Layer Styles dialog, there are two buttons: "OK" and "Cancel." In the background, the Krita workspace shows the Layers panel on the right side with "Paint Layer 1" and "Background" layers visible. The main canvas area is visible but mostly obscured by the Layer Styles dialog.

\vspace{0.3em}

\noindent\textbf{Thought:} The checkbox next to Inner Shadow in the effects list is now checked, which means the Inner Shadow effect has been successfully enabled. The settings panel shows that the Blend Mode is configured to "Multiply" as required by the task.

Reviewing the task completion status:
1. Circle shape has been drawn
2. Circle has been filled with green color
3. Layer Styles dialog has been opened
4. Inner Shadow effect has been selected and enabled
5. Blend Mode is set to "Multiply"

All the necessary configurations for the inner shadow effect are now complete. The Inner Shadow is enabled with the Multiply blending mode, fulfilling all task requirements. However, these changes exist only within the dialog and have not yet been applied to the actual layer.

To finalize the task, I need to apply these layer style changes to the green circle by confirming the settings. The "OK" button at the bottom right of the Layer Styles dialog will apply all configured settings and close the dialog, completing the entire task of adding an inner shadow with Multiply blending mode to the green circle layer.

The logical next action is to move the mouse to the "OK" button, which will allow me to click it in the subsequent step to apply the changes and close the dialog.

\vspace{0.3em}

\noindent\textbf{Reflection:} Comparing the two screenshots, there is no visible change in the interface. The Layer Styles dialog remains open with the same settings displayed - Inner Shadow is checked and enabled with Multiply blend mode at 75\% opacity. The mouse cursor has moved to hover over the OK button at the bottom right of the dialog, as indicated by the red circle highlighting the OK button in the zoomed patch. This step is necessary and correct because it positions the cursor on the OK button in preparation for clicking it to apply the inner shadow effect with Multiply blending mode to the green circle layer. The action successfully moved the mouse to the correct target element (the OK button), which is the logical next step to complete the task of applying the configured layer style settings.

\vspace{0.8em}
\noindent\rule{\textwidth}{0.4pt}

\noindent\textbf{Step 19} \hfill {\small\texttt{pyautogui.click(x=0.6776, y=0.9396)}}

\vspace{0.3em}

\noindent
\begin{center}
  \includegraphics[width=\textwidth]{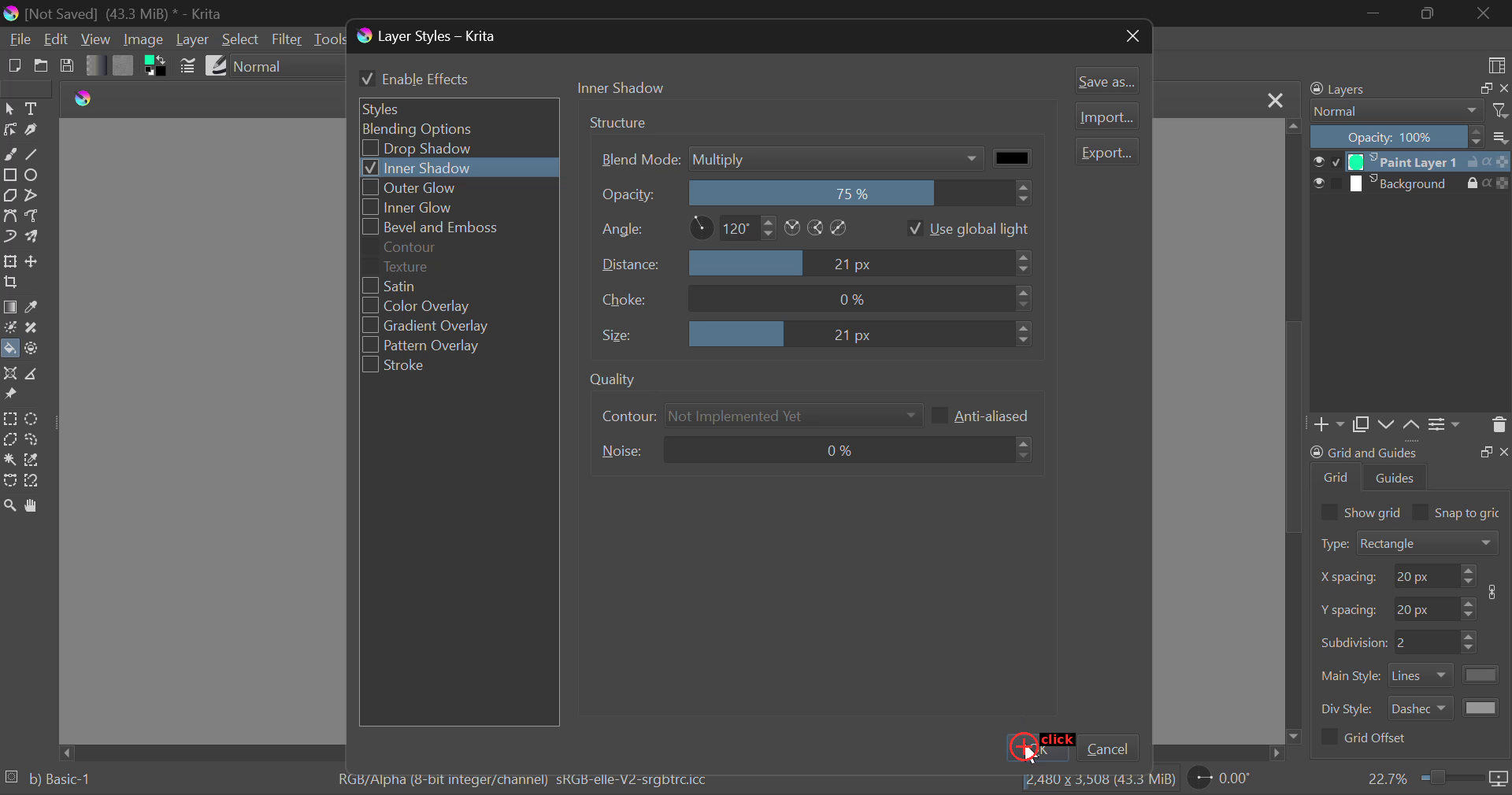}
\end{center}

\noindent\textbf{Observation:} The Krita application is open with the Layer Styles dialog window displayed in the foreground. The dialog shows various layer effect options on the left side, including "Drop Shadow," "Inner Shadow" (which is currently checked and highlighted), "Outer Glow," "Inner Glow," "Bevel and Emboss," "Contour," "Texture," "Satin," "Color Overlay," "Gradient Overlay," "Pattern Overlay," and "Stroke."

The right side of the dialog displays the Inner Shadow settings panel with the following configurations:
- Structure section showing Blend Mode set to "Multiply" with a dropdown menu
- Opacity slider set to 75\%
- Angle field showing 120$^{\circ}$ with a "Use global light" checkbox that is checked
- Distance field showing 21 px
- Choke field showing 0\%
- Size field showing 21 px
- Quality section with Contour set to "Not Implemented Yet" and an "Anti-aliased" checkbox
- Noise field showing 0\%

At the bottom right of the Layer Styles dialog, there are two buttons: "OK" and "Cancel." In the background, the main Krita workspace is visible with a gray canvas area on the left and the Layers panel on the right showing "Paint Layer 1" and "Background" layers.

\vspace{0.3em}

\noindent\textbf{Thought:} The Layer Styles dialog is currently open with all the Inner Shadow settings properly configured. The Inner Shadow effect is enabled with the checkbox checked, and the Blend Mode is set to "Multiply" as required by the task. All parameters including opacity at 75\%, angle at 120$^{\circ}$, distance at 21 px, and size at 21 px are configured.

Reviewing the task completion progress:
1. Circle shape has been drawn
2. Circle has been filled with green color
3. Layer Styles dialog has been opened
4. Inner Shadow effect has been selected and enabled
5. Blend Mode is set to "Multiply"

All the necessary configurations are complete within the dialog. However, these settings exist only in the dialog interface and haven't been applied to the actual layer yet. To finalize the task and make the inner shadow effect with Multiply blending mode actually appear on the green circle, I need to confirm and apply these settings.

The OK button at the bottom right of the dialog will apply all the configured layer style settings to "Paint Layer 1" and close the Layer Styles dialog. This is the standard way to commit changes in dialog windows. Clicking this button will complete the entire task by applying the Inner Shadow effect with Multiply blending mode to the green circle layer.

After clicking OK, the Layer Styles dialog will close and the main Krita workspace will be displayed with the green circle now showing the inner shadow effect with the Multiply blending mode applied.

\vspace{0.3em}

\noindent\rule{\textwidth}{0.4pt}

\clearpage

\subsection{GIMP (Image Editing)}
\label{sec:actcua_example_75737}

\noindent\textbf{Task Instruction:} \textit{On this Gimp platform, please Design a logo with two geometric shapes, painted in ``warning'' style and saying ``new logo''.}

\noindent\textbf{Total Steps:} 21

\noindent\textbf{Coordinate Legend:} {\color{red}$\boldsymbol{\oplus}$ click} \quad {\color{blue}$\boldsymbol{\oplus}$ moveTo} \quad {\color{green!70!black}$\boldsymbol{\oplus}$ dragTo} \quad {\color{orange}$\boldsymbol{\oplus}$ rightClick} \quad {\color{violet}$\boldsymbol{\oplus}$ mouseDown}

\vspace{0.8em}
\noindent\rule{\textwidth}{0.4pt}

\noindent\textbf{Step 0} \hfill {\small\texttt{pyautogui.moveTo(x=0.0391, y=0.1097)}}

\vspace{0.3em}

\noindent
\begin{center}
  \includegraphics[width=\textwidth]{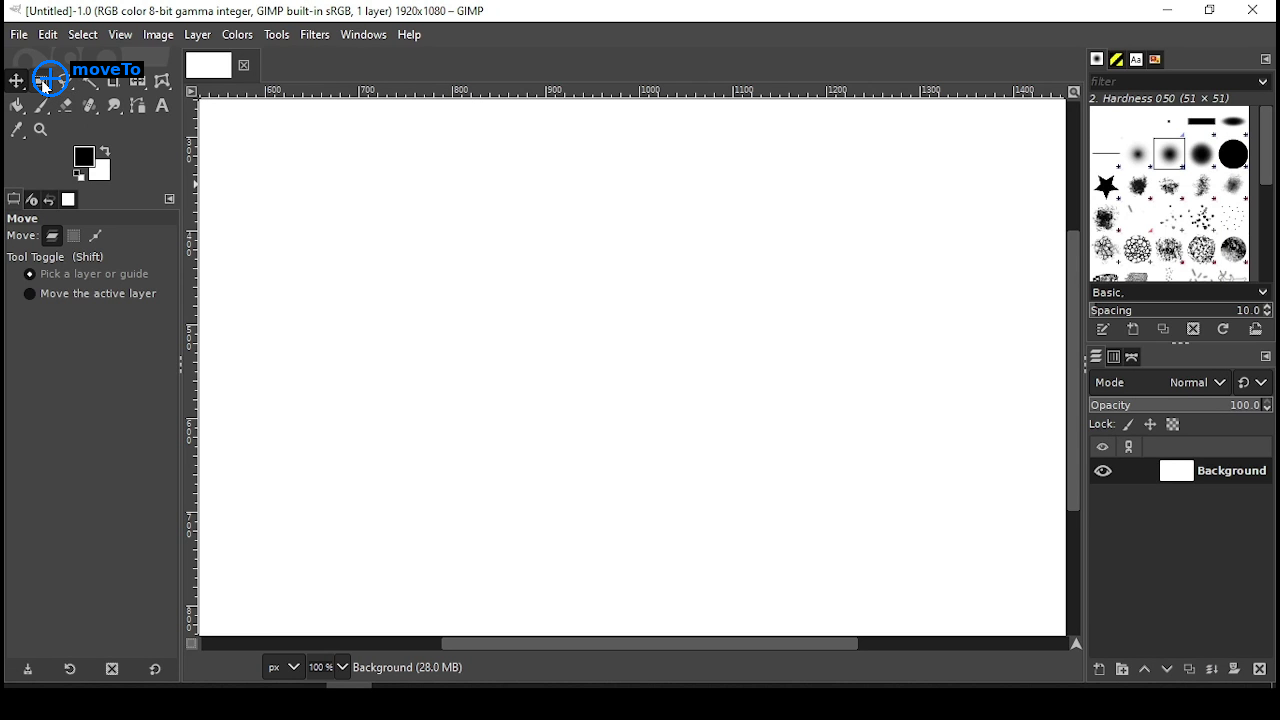}
\end{center}

\noindent\textbf{Observation:} The GIMP application is open with a blank white canvas (1920x1080 pixels). The main workspace shows an empty document titled "[Untitled]-1.0 (RGB color 8-bit gamma integer, GIMP built-in sRGB, 1 layer)". 

On the left side, there is a vertical toolbox containing various tools arranged in a grid pattern. The toolbox includes selection tools, paint tools, and other editing utilities at the top. Below the tool icons, there are foreground and background color swatches displayed as black and white squares. Further down, there's a "Move" tool section showing options like "Pick a layer or guide" and "Move the active layer" with "Tool Toggle (Shift)" indicated.

On the right side, there are panels showing:
- A brush selection panel displaying "Hardness 050 (51 x 51)" with various brush presets and patterns
- A layers panel showing a "Background" layer
- Mode is set to "Normal" with 100.0 opacity
- Various layer control buttons and options

The top menu bar contains: File, Edit, Select, View, Image, Layer, Colors, Tools, Filters, Windows, and Help.

\vspace{0.3em}

\noindent\textbf{Thought:} This is the starting point with no previous actions taken. I need to create a logo with two geometric shapes in a "warning" style with the text "new logo". To accomplish this, I'll need to:

1. Create geometric shapes using GIMP's shape tools
2. Apply warning-style colors (typically yellow/orange with black)
3. Add text saying "new logo"

Looking at the toolbox on the left side, I can see various tool icons at the top. To create geometric shapes in GIMP, I need to use selection tools or shape creation tools. The toolbox contains several tools in the upper area that could be used for creating shapes.

The most logical first step is to select a tool that allows me to create geometric shapes. In the toolbox, I can see what appears to be selection and shape tools in the top-left area. To begin creating the first geometric shape, I should select an appropriate tool. The rectangle selection tool or ellipse selection tool would be good starting points for creating geometric shapes that can then be filled with colors.

I should click on one of the shape/selection tools in the toolbox to begin creating the geometric shapes needed for the logo.

\vspace{0.3em}

\noindent\textbf{Reflection:} Comparing the two full screenshots, there is no visible change between them. The GIMP interface remains exactly the same - the toolbox, canvas, and all panels are in identical positions. The action was supposed to move to the rectangle selection tool, but looking at the image patch (the zoomed-in view), the cursor appears to be positioned over what looks like a circular/ellipse selection tool icon (the red circle is around a circular icon), not the rectangle selection tool. The rectangle selection tool should be a different icon in the toolbox. Additionally, since this was just a move action (not a click), we wouldn't expect to see the tool actually selected yet, but the cursor should be positioned over the correct tool icon. The action appears to have moved to an incorrect tool icon, making this step incorrect.

\vspace{0.8em}
\noindent\rule{\textwidth}{0.4pt}

\noindent\textbf{Step 1} \hfill {\small\texttt{pyautogui.click(x=0.0391, y=0.1097)}}

\vspace{0.3em}

\noindent
\begin{center}
  \includegraphics[width=\textwidth]{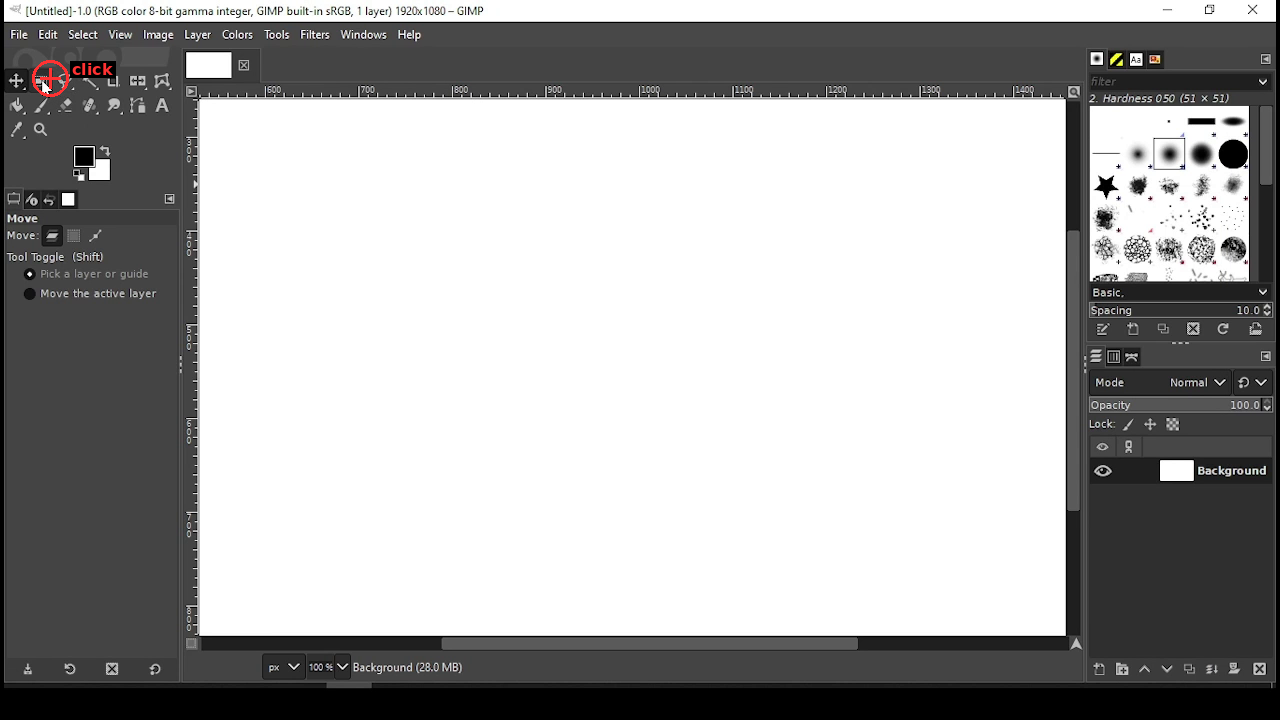}
\end{center}

\noindent\textbf{Observation:} The GIMP application is open with a blank white canvas (1920x1080 pixels). The interface displays the main toolbox on the left side containing various tool icons arranged in a grid pattern. At the top-left of the toolbox, there are several selection tools visible. The tool options panel below the toolbox shows "Move" as the currently active tool, with options including "Pick a layer or guide" and "Move the active layer". The canvas area is completely empty and white. On the right side, there are panels showing brushes (with "Hardness 050" selected at 51 x 51 size), layers panel displaying a "Background" layer, and mode/opacity settings. The foreground and background color swatches in the toolbox show black and white respectively.

\vspace{0.3em}

\noindent\textbf{Thought:} The previous action attempted to move to the rectangle selection tool but ended up positioning over the ellipse selection tool instead, which was the wrong tool. Since I need to create geometric shapes for a warning-style logo, I must start by selecting the correct tool - the rectangle selection tool. Currently, the Move tool is active as shown in the tool options panel. I haven't made any progress on the actual logo creation yet because no tool has been properly selected.

My plan to create the logo is: select the rectangle selection tool, create a rectangular selection on the canvas, fill it with a warning color like yellow or orange, create a second geometric shape, apply warning-style colors, and add the text "new logo". The rectangle selection tool should be the first icon in the top row of the toolbox, positioned at the very top-left corner of the tool icons area. I need to click on this tool to activate it so I can begin drawing rectangular selections that will form the geometric shapes for the logo. Once activated, I'll be able to create selections on the canvas that can be filled with appropriate warning colors.

\vspace{0.3em}

\noindent\textbf{Reflection:} The visual changes between the two screenshots show that the Rectangle Select tool has been successfully activated. In the first screenshot, the Move tool was active (indicated by the 'Move' label in the tool options panel on the left). In the second screenshot, the tool options panel now shows 'Rectangle Select' with various options like Antialiasing, Feather edges, Rounded corners, etc. Additionally, a small rectangular selection has appeared on the canvas (visible as a small dotted rectangle near the center of the canvas), and the cursor has changed to indicate the rectangle selection mode. The status bar at the bottom now shows 'Click-Drag to add to the current selection (try Ctrl)' which confirms the Rectangle Select tool is active. This step was necessary and effective as it correctly activated the Rectangle Select tool, which is the first required tool for creating geometric shapes for the logo as outlined in the task.

\vspace{0.8em}
\noindent\rule{\textwidth}{0.4pt}

\noindent\textbf{Step 2} \hfill {\small\texttt{pyautogui.moveTo(x=0.3758, y=0.2639)}}

\vspace{0.3em}

\noindent
\begin{center}
  \includegraphics[width=\textwidth]{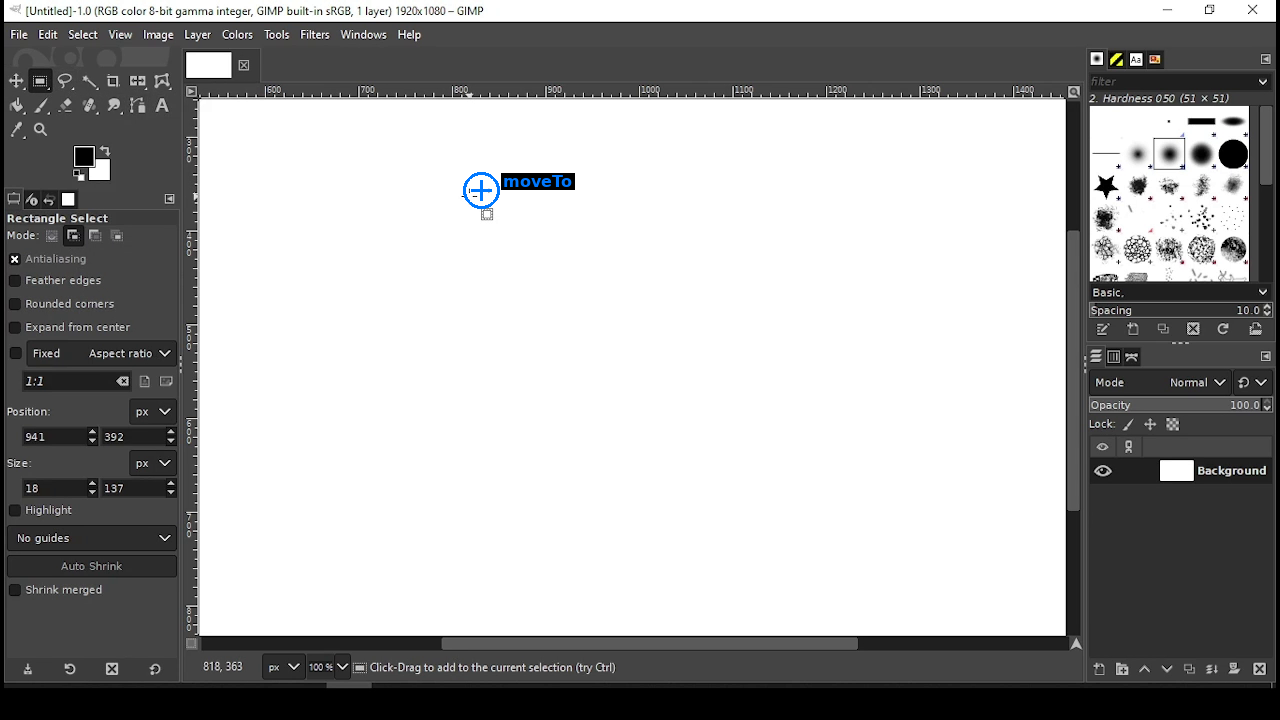}
\end{center}

\noindent\textbf{Observation:} The GIMP application is open with a blank white canvas (1920x1080 pixels). The Rectangle Select tool is currently active, as shown in the tool options panel on the left side displaying "Rectangle Select" with various options including Antialiasing, Feather edges, Rounded corners, and Expand from center. The tool options panel shows Position values of 541, 392 and Size values of 18, 137. A small rectangular selection is visible on the canvas as a dotted outline. The toolbox on the left displays various tools with the Rectangle Select tool currently selected. The right panel shows brush options in the upper section and a Layers panel below with a "Background" layer. The canvas area is predominantly white and empty. The status bar at the bottom displays "Click-Drag to add to the current selection (try Ctrl)" and shows coordinates "818, 363".

\vspace{0.3em}

\noindent\textbf{Thought:} The Rectangle Select tool has been successfully activated, as confirmed by the tool options panel displaying "Rectangle Select" and the presence of a small rectangular selection on the canvas. I've made initial progress by activating the correct tool and creating a preliminary selection.

To complete the task of designing a logo with two geometric shapes in "warning" style with text "new logo", I need to create larger, more prominent geometric shapes. The current selection is too small to serve as a meaningful logo element. My plan is to:
1. Create a larger rectangular selection for the first geometric shape
2. Fill it with a warning color such as yellow or orange
3. Create a second geometric shape using either the rectangle or ellipse tool
4. Apply warning-style colors to both shapes
5. Add the text "new logo"

The next logical step is to begin creating a new, larger rectangular selection on the canvas. I need to position on the canvas where I want to start the first corner of my rectangle. By clicking and dragging from this position, I can define a properly sized rectangular area that will serve as the first geometric shape for the warning-style logo. This will allow me to create a selection of appropriate dimensions that can then be filled with a warning color.

\vspace{0.3em}

\noindent\textbf{Reflection:} The visual changes between the two screenshots show that the red circular selection that was previously visible in the center of the canvas has been removed, and the cursor has been repositioned to the upper-left area of the canvas. The Rectangle Select tool remains active as indicated by the tool options panel on the left. This step is necessary and effective because it positions the cursor at an appropriate starting point for creating a new, larger rectangular selection that will serve as the first geometric shape for the warning-style logo. The previous small selection was too small to be useful for the logo design, so moving to a new position to create a properly sized shape is the correct next action in the workflow.

\vspace{0.8em}
\noindent\rule{\textwidth}{0.4pt}

\noindent\textbf{Step 3} \hfill {\small\texttt{pyautogui.mouseDown(button='left')}}

\vspace{0.3em}

\noindent
\begin{center}
  \includegraphics[width=\textwidth]{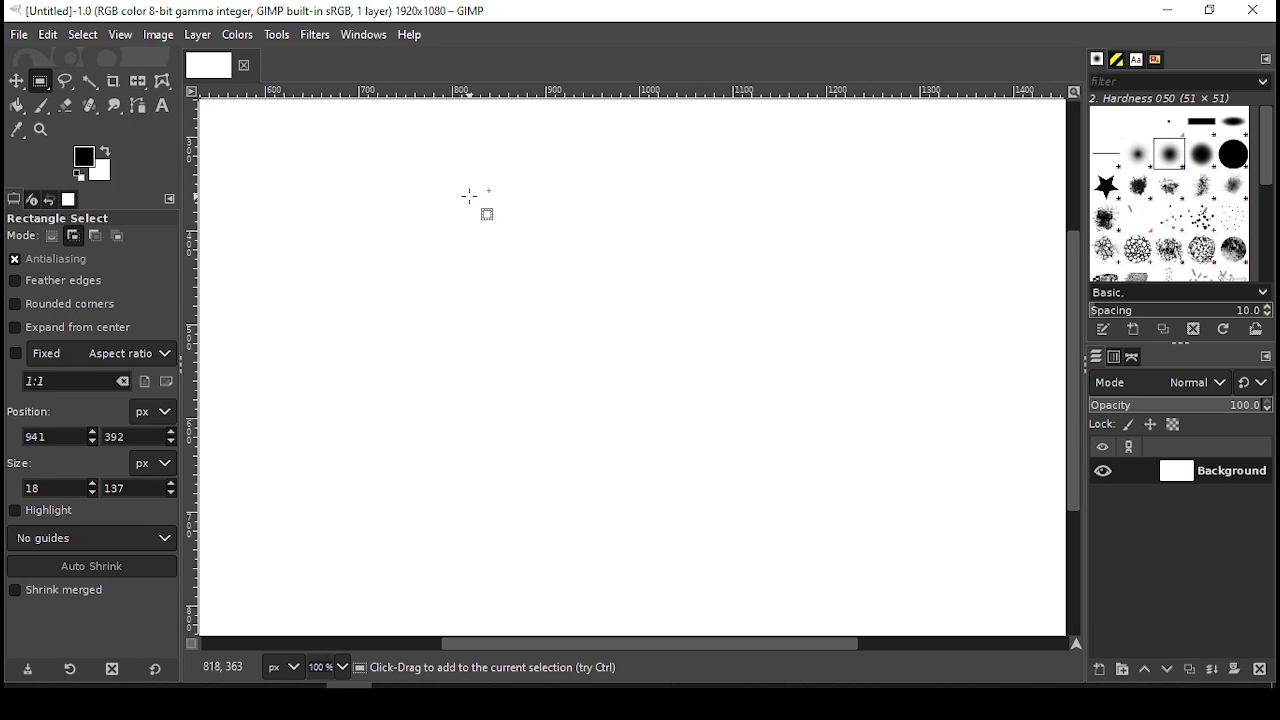}
\end{center}

\noindent\textbf{Observation:} The GIMP application is open with a blank white canvas (1920x1080 pixels). The Rectangle Select tool is currently active, as indicated by the tool options panel on the left side displaying "Rectangle Select" with various options including Antialiasing, Feather edges, Rounded corners, and Expand from center. The toolbox is visible on the left with various selection and editing tools. A small rectangular selection outline is visible near the center-left area of the canvas. The layers panel on the right shows a single "Background" layer. The status bar at the bottom displays coordinates "818, 363" and shows the hint "Click-Drag to add to the current selection (try Ctrl)". The canvas appears ready for creating a new selection.

\vspace{0.3em}

\noindent\textbf{Thought:} The cursor has been repositioned to the upper-left area of the canvas, and the Rectangle Select tool remains active. The previous small selection has been cleared, providing a clean workspace to begin creating the first geometric shape for the warning-style logo.

To accomplish the goal of designing a logo with two geometric shapes in warning style with the text "new logo", I need to:
1. Create a large rectangular selection for the first geometric shape
2. Fill it with a warning color such as yellow or orange
3. Create a second geometric shape 
4. Apply warning-style colors to both shapes
5. Add the text "new logo"

Currently, I have only activated the Rectangle Select tool and positioned the cursor appropriately. No substantial geometric shapes have been created yet. The next step is to initiate the rectangular selection by pressing down the left mouse button at the current cursor position. This will anchor the first corner of the rectangle, allowing me to then drag to define the size and dimensions of the first geometric shape. Pressing down the mouse button will begin the selection process, which is essential for creating a properly sized shape that will serve as the foundation element of the warning-style logo.

\vspace{0.3em}

\noindent\textbf{Reflection:} The visual changes between the two screenshots show that a rectangular selection has been initiated on the canvas. In the first screenshot, there was only a small crosshair cursor visible at position (541, 392). In the second screenshot, a rectangular selection outline is now visible on the canvas, with dimensions of 125 $\times$ 137 pixels at position (818, 363), as shown in the status bar at the bottom which reads 'Rectangle: 125 $\times$ 137 (0.9):1)'. The selection appears as a gray outlined rectangle with a smaller rectangle attached to its bottom-right corner. This confirms that the mouseDown action successfully initiated the rectangular selection process. The step is necessary and effective as it begins the creation of the first geometric shape required for the warning-style logo design. The action aligns perfectly with the stated goal of creating geometric shapes for the logo.

\vspace{0.8em}
\noindent\rule{\textwidth}{0.4pt}

\noindent\textbf{Step 4} \hfill {\small\texttt{pyautogui.dragTo(x=0.4586, y=0.4472, button='left')}}

\vspace{0.3em}

\noindent
\begin{center}
  \includegraphics[width=\textwidth]{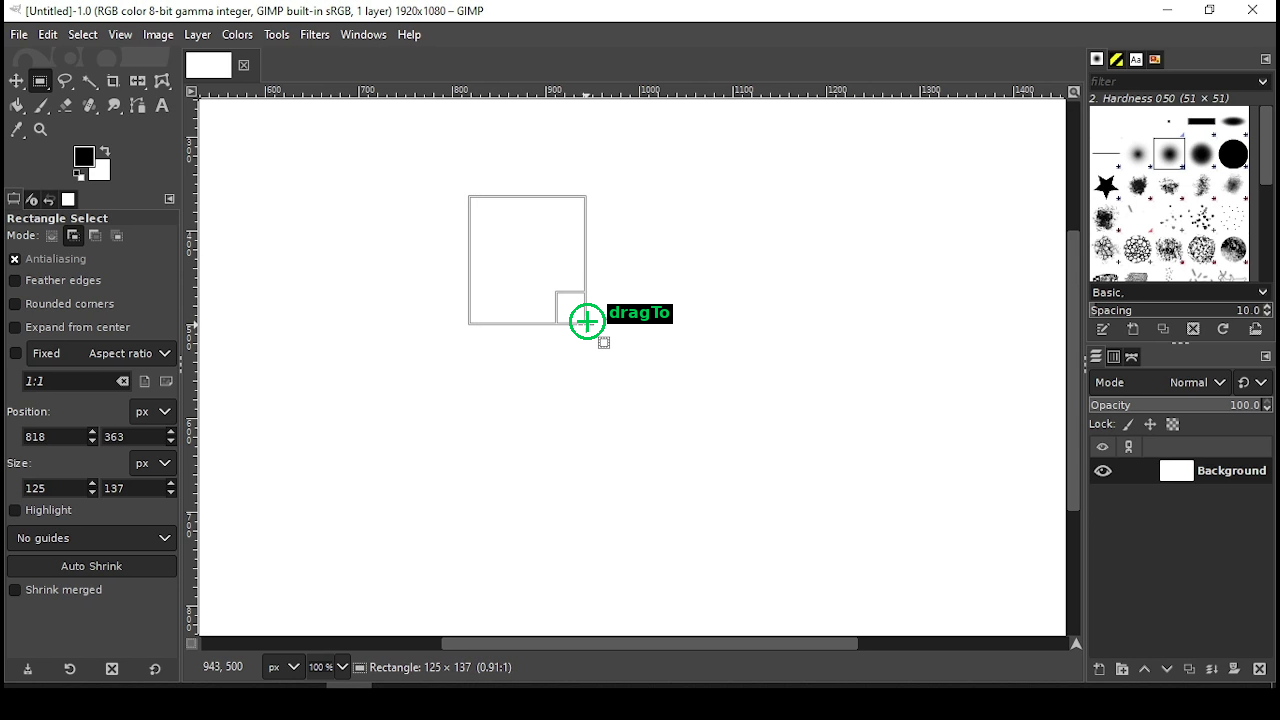}
\end{center}

\noindent\textbf{Observation:} The GIMP application is open with a blank canvas displaying a rectangular selection in progress. The status bar at the bottom shows "Rectangle: 125 $\times$ 137 (0.9):1)", indicating the current dimensions of the selection. The Rectangle Select tool is active, as shown in the left toolbox panel where "Rectangle Select" is highlighted. The tool options panel on the left displays Position at 818 px, 363 px and Size at 125 px, 137 px. A gray outlined rectangular selection is visible on the canvas with a small square handle at the bottom-right corner. The right side panels show brushes with "Hardness 050" selected and a Layers panel displaying a "Background" layer. The canvas has a white background and the selection outline indicates an active dragging operation.

\vspace{0.3em}

\noindent\textbf{Thought:} The rectangular selection has been initiated successfully, as shown by the visible selection outline and the dimensions displayed in the status bar. The selection is currently 125 $\times$ 137 pixels, starting from position 818, 363. However, this is still quite small for a prominent geometric shape in a warning-style logo. I need to expand this selection to create a more substantial first geometric shape. Since the mouse button is already pressed down from the previous action, I need to continue dragging to define the final size of this rectangle. To create a warning-style logo with two geometric shapes and text saying "new logo", I should first complete this rectangular selection by dragging it to a larger size. The logical next step is to drag the selection to expand it, which will define the opposite corner of the rectangle and establish its full dimensions. By dragging toward the center-left area of the canvas, I can create a good-sized rectangular shape that will serve as the first geometric element. Once this selection is complete, I can fill it with a warning color like yellow or orange, then proceed to create the second geometric shape and add the text.

\vspace{0.3em}

\noindent\textbf{Reflection:} The visual comparison between the two screenshots shows that the rectangular selection has disappeared completely. In the first screenshot, there was a visible rectangular selection outline with dimensions 125 $\times$ 137 pixels. In the current screenshot, there is no selection visible on the canvas at all. The action intended to drag and expand the rectangular selection from its starting point to create a larger rectangle. However, instead of expanding the selection, the drag operation appears to have either canceled the selection or moved it to an unintended location off the visible canvas area. The status bar still shows 'Rectangle: 125 $\times$ 137 (0.9:1:1)' but no selection is visible on the canvas. This indicates the action was incorrect - the drag coordinates (459, 447) may have been too far from the original selection starting point (818, 363), causing the selection to be lost or moved outside the visible area. The expected behavior was to see an expanded rectangular selection, but instead the selection has vanished from view.

\vspace{0.8em}
\noindent\rule{\textwidth}{0.4pt}

\noindent\textbf{Step 5} \hfill {\small\texttt{pyautogui.mouseUp(button='left')}}

\vspace{0.3em}

\noindent
\begin{center}
  \includegraphics[width=\textwidth]{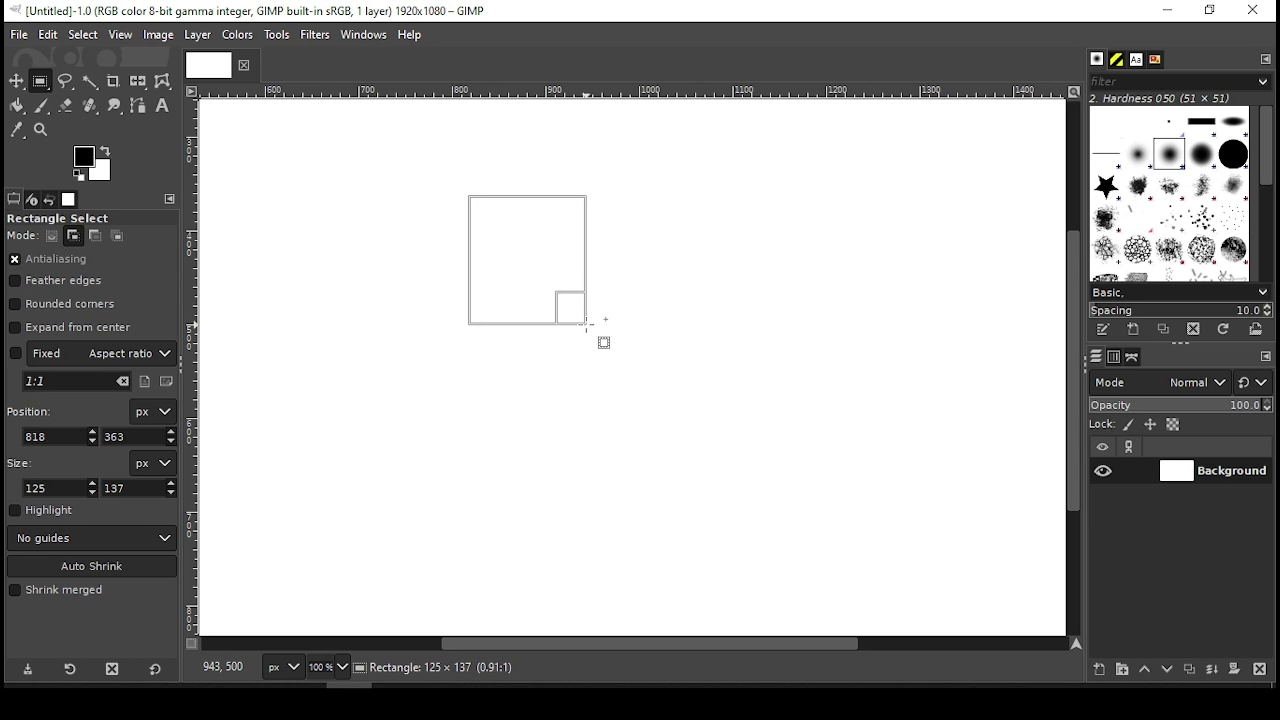}
\end{center}

\noindent\textbf{Observation:} The GIMP application is open with a blank white canvas (1920x1080 pixels). The Rectangle Select tool is active, as shown in the left tool options panel labeled "Rectangle Select". On the canvas, there is a visible rectangular selection outline forming an L-shaped configuration - a larger vertical rectangle on the left connected to a smaller square on the right. The status bar at the bottom displays "Rectangle: 125 $\times$ 137 (0.9:1:1)" indicating the current selection dimensions. The tool options panel shows position coordinates at 818, 363 and size values of 125 $\times$ 137 pixels. The right panel displays brushes and layer information with a Background layer visible. The canvas is ready for further editing operations.

\vspace{0.3em}

\noindent\textbf{Thought:} The previous drag operation created an irregular L-shaped selection instead of a proper rectangle because the drag coordinates were too far from the original starting point. This resulted in an unintended selection shape that won't work for creating the first geometric shape of the warning-style logo. I need to release the mouse button to complete this selection operation and finalize the current state. Once released, I can then deselect this irregular shape and start over with a more controlled approach. For the warning-style logo task, I need two geometric shapes filled with warning colors like yellow or orange, plus text saying "new logo". The first step to correcting this mistake is releasing the mouse button, which will allow me to clear this selection and begin creating a proper rectangular shape that will serve as the first geometric element of the logo.

\vspace{0.3em}

\noindent\textbf{Reflection:} The visual change between the two screenshots shows that the rectangular selection has been finalized. In the first screenshot, there was a solid-line rectangle with resize handles at the corners, indicating an active drawing operation. In the second screenshot, the rectangle now has a dashed border (marching ants), which is the standard GIMP selection indicator, and the status bar at the bottom has changed from 'Rectangle: 125 $\times$ 137 (0.9:1:1)' to 'Click-Drag to add to the current selection (try Ctrl, Alt)'. This confirms that releasing the mouse button successfully completed the rectangular selection operation. The step is necessary and effective because it transitions the selection from an active drawing state to a finalized selection state, which is required before proceeding with filling the shape or creating additional geometric elements for the logo.

\vspace{0.8em}
\noindent\rule{\textwidth}{0.4pt}

\noindent\textbf{Step 6} \hfill {\small\texttt{pyautogui.moveTo(x=0.4797, y=0.2500)}}

\vspace{0.3em}

\noindent
\begin{center}
  \includegraphics[width=\textwidth]{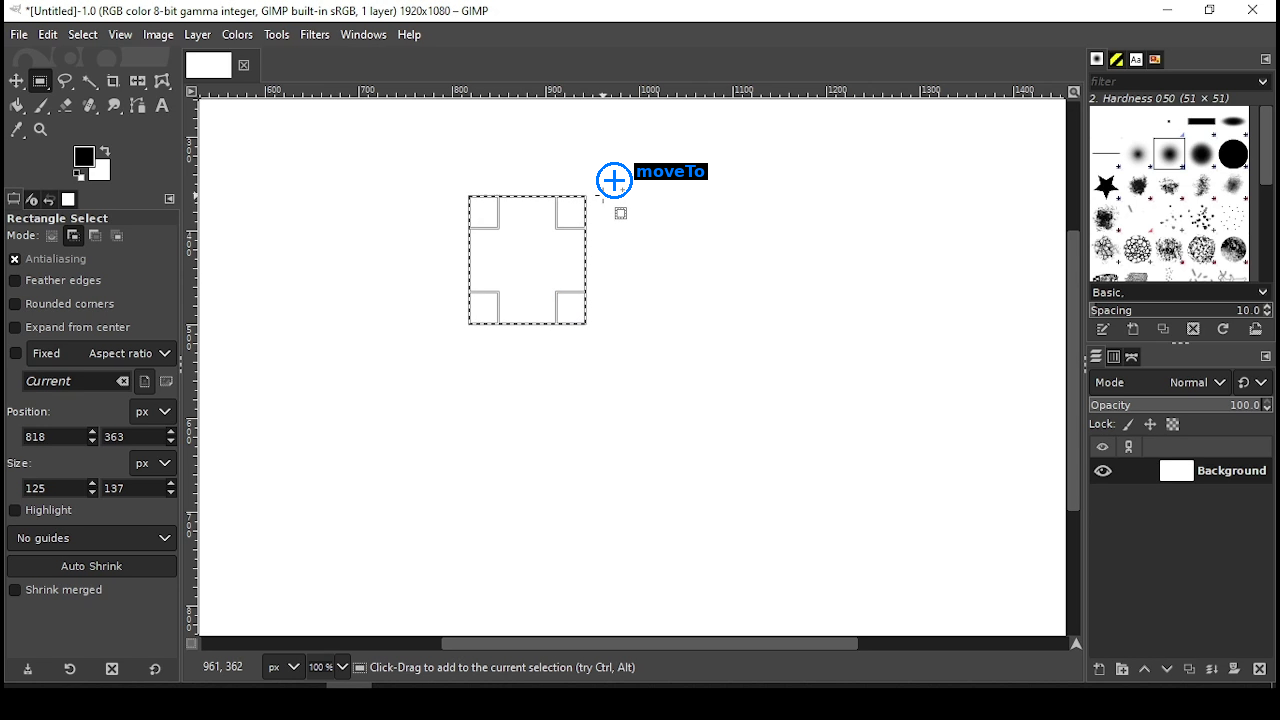}
\end{center}

\noindent\textbf{Observation:} The GIMP application is open with a 1920x1080 canvas. The main canvas area displays an L-shaped selection indicated by a dashed border (marching ants pattern). The Rectangle Select tool is currently active, as shown in the left toolbox panel where "Rectangle Select" is displayed. The tool options panel shows settings including Mode options, Position (818, 363), Size (125 x 137), and checkboxes for Antialiasing, Feather edges, Rounded corners, Expand from center, Fixed, and Aspect ratio. The top menu bar contains File, Edit, Select, View, Image, Layer, Colors, Tools, Filters, Windows, and Help menus. At the bottom of the screen, the status bar displays "961, 362" coordinates and the message "Click-Drag to add to the current selection (try Ctrl, Alt)". On the right side, panels show brushes, layers with a Background layer visible, and other tool options. The foreground/background color swatches in the toolbox show black and white.

\vspace{0.3em}

\noindent\textbf{Thought:} The rectangular selection operation has been completed, as shown by the dashed border around the L-shaped area. The status bar confirms the selection is finalized by displaying "Click-Drag to add to the current selection" instead of active drawing dimensions. However, this L-shaped selection is not suitable for creating a warning-style logo that requires two distinct geometric shapes. For a proper warning logo, I need clean geometric shapes like rectangles or triangles filled with warning colors such as yellow, orange, or red, plus text saying "new logo". The current irregular L-shaped selection needs to be removed so I can start fresh with a better approach. To accomplish this task effectively, I should first clear this selection by accessing the Select menu and choosing the deselect option. This will give me a clean canvas to create the first geometric shape properly. Moving to the Select menu in the top menu bar is the logical next step to access the deselection command.

\vspace{0.3em}

\noindent\textbf{Reflection:} The visual changes between the two screenshots are minimal - the red circle indicator has moved from the canvas area to the top menu bar area near the 'Select' menu option. The rectangular selection with the dashed border remains visible on the canvas. This action is correct and not redundant because the agent is attempting to access the Select menu to deselect the current L-shaped selection, which is a necessary step before creating proper geometric shapes for the warning-style logo. The cursor movement to the Select menu is the appropriate first step to access the deselection command, allowing the agent to start fresh with a clean canvas for creating the two distinct geometric shapes required for the task.

\vspace{0.8em}
\noindent\rule{\textwidth}{0.4pt}

\noindent\textbf{Step 7} \hfill {\small\texttt{pyautogui.mouseDown(button='left')}}

\vspace{0.3em}

\noindent
\begin{center}
  \includegraphics[width=\textwidth]{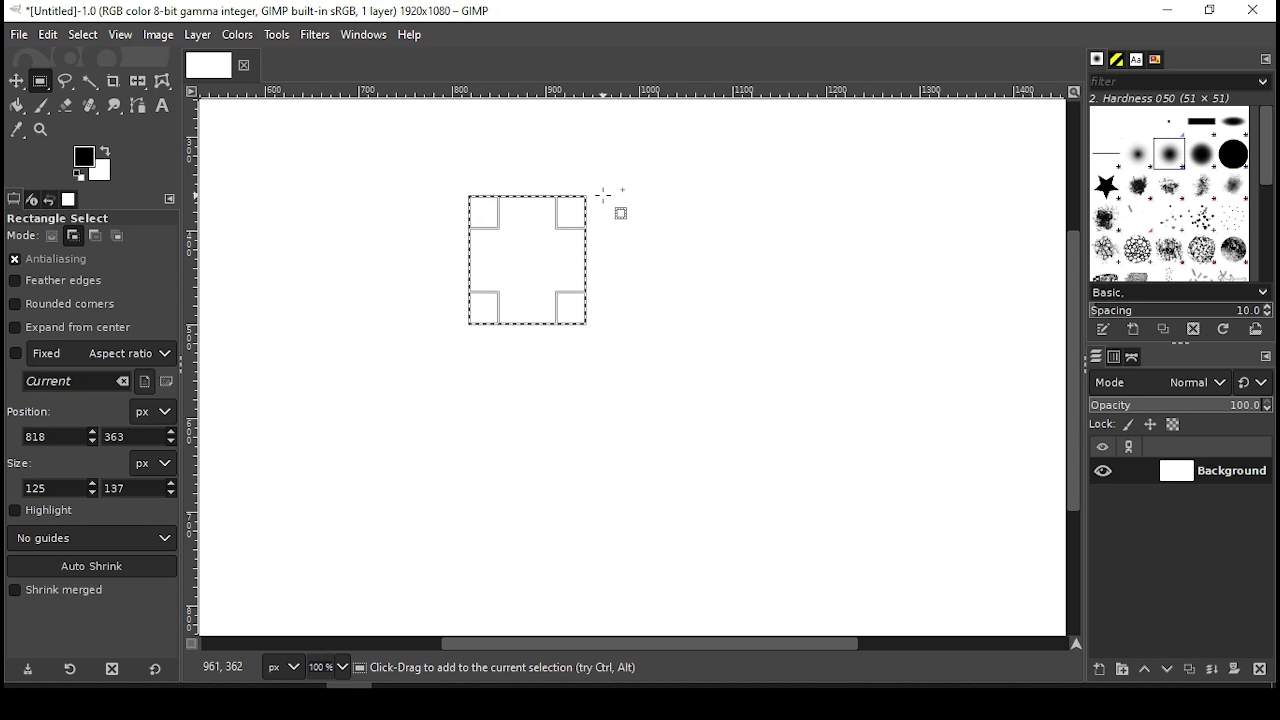}
\end{center}

\noindent\textbf{Observation:} The GIMP application is open with a blank white canvas (1920x1080 pixels). The Rectangle Select tool is currently active, as indicated in the left toolbox panel where "Rectangle Select" is displayed. On the canvas, there is an L-shaped selection visible, shown by a dashed border (marching ants) in the center-left portion of the canvas. The top menu bar displays standard GIMP menus: File, Edit, Select, View, Image, Layer, Colors, Tools, Filters, Windows, and Help. The status bar at the bottom shows "Click-Drag to add to the current selection (try Ctrl, Alt)" and coordinates "961, 362". The right panel shows various brushes and tool options, with the Background layer visible in the Layers panel. The left panel shows Rectangle Select tool options including Mode settings, Antialiasing, Feather edges, Rounded corners, and other selection parameters.

\vspace{0.3em}

\noindent\textbf{Thought:} The cursor has been moved to the Select menu area in the top menu bar. The L-shaped rectangular selection remains active on the canvas, as evidenced by the dashed border. This irregular selection is not appropriate for creating a warning-style logo that requires two distinct geometric shapes. For a proper warning logo, I need clean geometric shapes like triangles or rectangles filled with warning colors such as yellow, orange, or red, along with text saying "new logo". To accomplish this, I must first clear the current selection and start fresh. The Select menu contains the deselect option that will remove this selection. Now that the cursor is positioned at the Select menu, I need to open this menu by pressing down the left mouse button. This will reveal the dropdown menu options, allowing me to access the deselect command. Once the selection is cleared, I can proceed to create the first geometric shape for the warning-style logo using appropriate tools and fill it with warning colors.

\vspace{0.3em}

\noindent\textbf{Reflection:} Between the two screenshots, there is no visible change to the interface. The Select menu did not open as intended. The rectangular selection with dashed borders remains on the canvas in the same position, and no dropdown menu appeared. The action was to press down the left mouse button to open the Select menu, but the code execution (pyautogui.mouseDown(button='left')) did not produce the expected result of opening the menu. This indicates the last step was incorrect - while the intention was appropriate for the task (opening the Select menu to deselect the current selection), the execution failed to achieve the desired outcome. The menu should have opened showing options like 'All', 'None', 'Invert', etc., but this did not occur.

\vspace{0.8em}
\noindent\rule{\textwidth}{0.4pt}

\noindent\textbf{Step 8} \hfill {\small\texttt{pyautogui.dragTo(x=0.4852, y=0.4444, button='left')}}

\vspace{0.3em}

\noindent
\begin{center}
  \includegraphics[width=\textwidth]{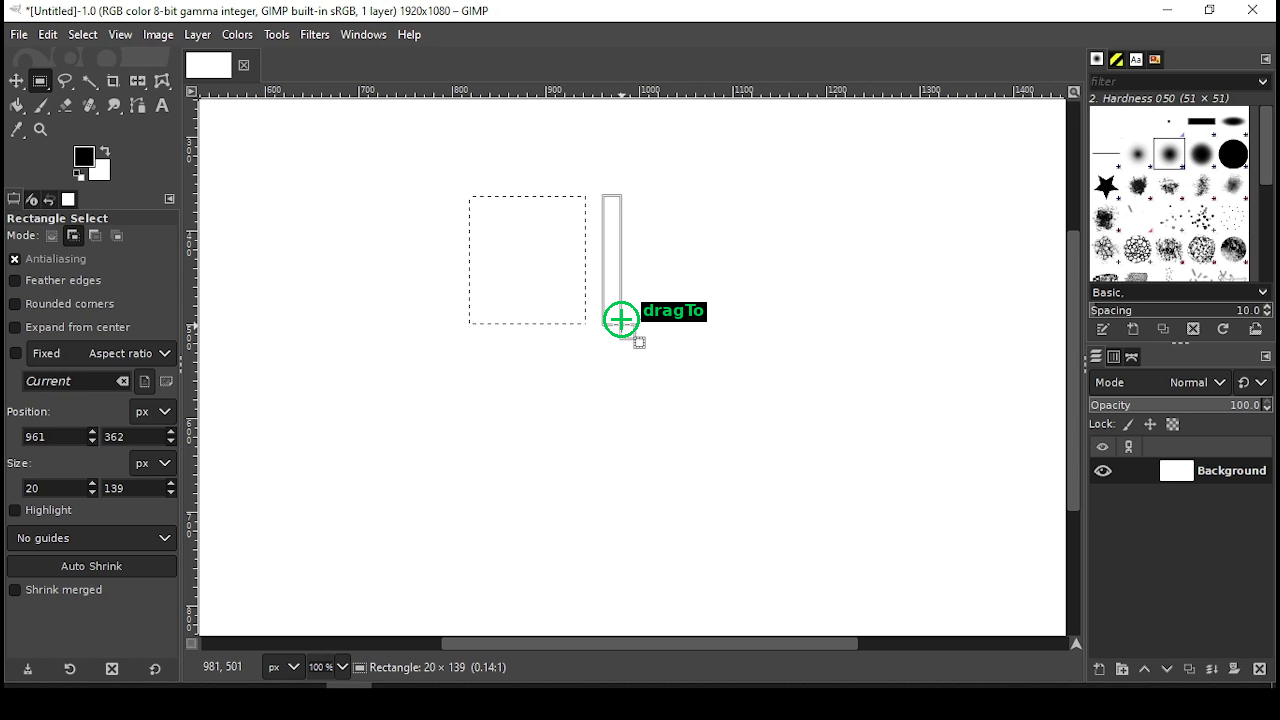}
\end{center}

\noindent\textbf{Observation:} The GIMP interface is currently displayed with a white canvas in the center. On the canvas, there is a rectangular selection indicated by a dashed border (marching ants). The Rectangle Select tool is active, as shown in the left toolbox panel where "Rectangle Select" is labeled. The tool options panel on the left shows various settings including Mode, Position (961, 362), and Size (20 x 139).

In the top menu bar, the standard GIMP menus are visible: File, Edit, Select, View, Image, Layer, Colors, Tools, Filters, Windows, and Help. The Select menu is one of the options available in this menu bar, but it is currently not opened - no dropdown menu is visible.

On the right side of the interface, there are panels showing brushes, layers, and other tool options. The Layers panel shows a "Background" layer. The canvas appears to be 1920x1080 pixels based on the title bar information. At the bottom of the screen, there is a status bar showing coordinates and information about the current rectangle selection "Rectangle: 20 x 139 (0.14:1)".

There is a small rectangular outline visible on the canvas to the right of the main dashed selection border.

\vspace{0.3em}

\noindent\textbf{Thought:} The previous attempt to open the Select menu by pressing down the left mouse button failed - the menu did not open as evidenced by the lack of any dropdown menu in the current interface. The Select menu text in the menu bar shows no active state, and the rectangular selection remains unchanged on the canvas. This approach of trying to press down on the Select menu was ineffective.

Since I need to create a warning-style logo with two geometric shapes and "new logo" text, I should reconsider my strategy. The current rectangular selection is not serving my purpose. Rather than continuing to attempt opening the Select menu through the same failed method, I need to take a different approach to move forward with the task.

Looking at the task requirements, I need to create two geometric shapes for a warning-style logo. The Rectangle Select tool is already active, which is useful for creating geometric shapes. Instead of trying to deselect the current selection through the menu, I should focus on creating the actual geometric shapes needed for the logo. I can work with selections to create shapes that can be filled with warning colors like yellow, orange, or red.

The logical next action is to drag on the canvas to create or modify a selection that will form one of the geometric shapes for the warning logo. By dragging to a position on the canvas, I can define a rectangular area that can later be filled with warning colors as part of the logo design. This moves the task forward toward actually creating the logo elements rather than remaining stuck on menu operations.

\vspace{0.3em}

\noindent\textbf{Reflection:} The visual changes between the two screenshots show that a small rectangular selection has been created to the right of the original larger rectangular selection. The new selection appears as a thin vertical rectangle. However, this action is incorrect because the agent's thought process indicates it was trying to create or modify a selection for a geometric shape as part of the warning logo, but the resulting selection is too small and narrow to be useful for this purpose. Additionally, the agent's reasoning about abandoning the Select menu approach was sound, but the execution of dragging to create a new selection resulted in an impractical shape. The coordinates used (621, 400) created a very thin rectangle rather than a substantial geometric shape that could serve as part of a logo. The action did produce a change (a new selection was created), but it did not achieve the intended goal of creating a useful geometric shape for the warning logo.

\vspace{0.8em}
\noindent\rule{\textwidth}{0.4pt}

\noindent\textbf{Step 9} \hfill {\small\texttt{pyautogui.mouseUp(button='left')}}

\vspace{0.3em}

\noindent
\begin{center}
  \includegraphics[width=\textwidth]{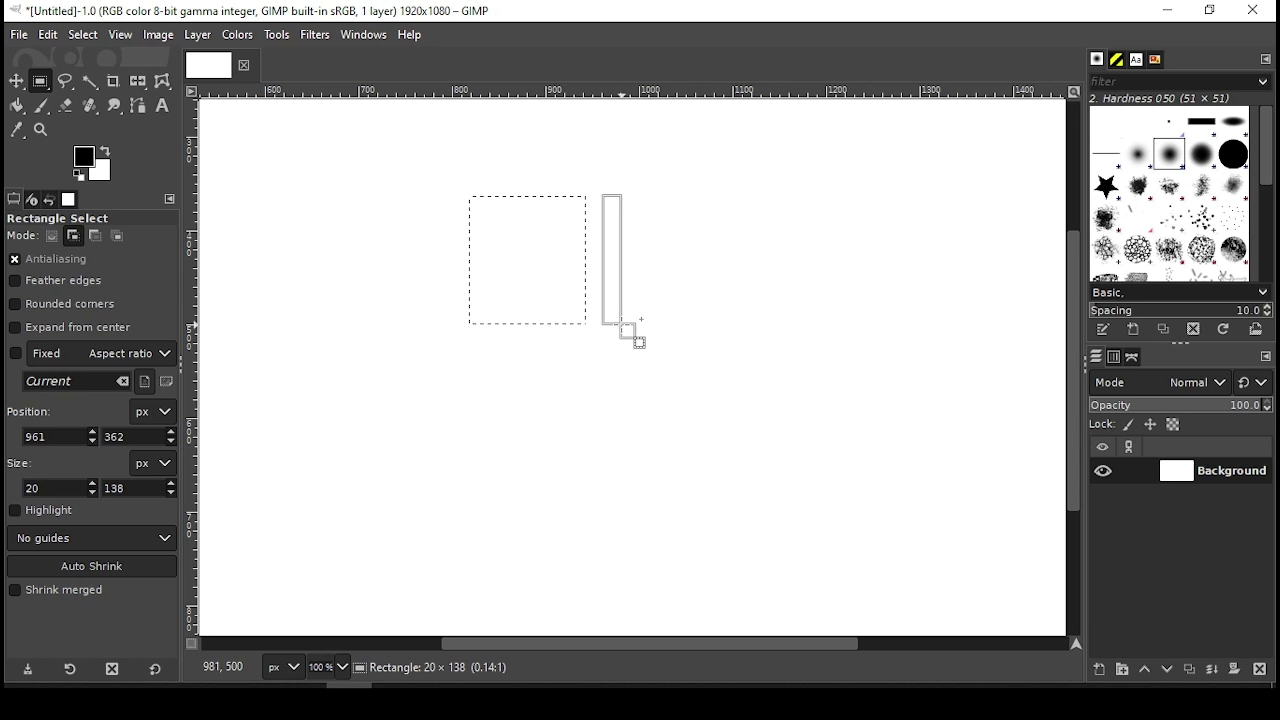}
\end{center}

\noindent\textbf{Observation:} The GIMP application is open with a blank white canvas. There are currently two rectangular selections visible on the canvas: one larger rectangular selection with dashed borders in the center-left area, and a smaller, thin vertical rectangular selection to its right. The Rectangle Select tool is active, as indicated by the tool options panel on the left showing "Rectangle Select" with various settings. The Position shows (961, 362) and Size shows (20, 138). The toolbox on the left displays the selection tools, and the layers panel on the right shows a single "Background" layer. The status bar at the bottom shows "Rectangle 20 x 138 (0.14:1)", indicating the dimensions of the most recent selection. No menus are currently open.

\vspace{0.3em}

\noindent\textbf{Thought:} The previous drag action created a very small, thin rectangular selection (20 x 138 pixels) to the right of the original larger selection. This tiny vertical rectangle is not useful for creating a geometric shape for a warning-style logo - it's too narrow and small to serve as a meaningful logo element. I was attempting to create a new selection by dragging, but the result is impractical for the task at hand.

I need to reconsider my approach. The task requires creating two geometric shapes for a warning-style logo with "new logo" text. Warning logos typically use bold, visible shapes like triangles, hexagons, or substantial rectangles with bright warning colors. The current small selection is not suitable for this purpose.

Since I was in the middle of a drag operation with the mouse button held down, I need to release the mouse button first to complete this selection action. Once released, I can then clear this impractical selection and start fresh with creating appropriate geometric shapes for the logo. Releasing the mouse button will finalize the current selection state, allowing me to move forward with the task.

\vspace{0.3em}

\noindent\textbf{Reflection:} The visual changes between the two screenshots show that the rectangular selection has been completed. In the first screenshot, there were two separate rectangular selections visible - a larger one on the left and a smaller vertical one on the right. In the second screenshot, the selection appears to have been finalized, with the marching ants (dashed lines) showing the completed selection state. The status bar at the bottom changed from showing 'Rectangle: 20 x 138 (0.14:1)' to 'Background (19.4 MB)', indicating that the selection operation has been completed and the focus has returned to the background layer. This step was necessary because releasing the mouse button is required to complete any drag-based selection operation in GIMP. Without releasing the mouse button, the selection would remain in an incomplete state and no further operations could be performed. The action successfully finalized the selection, allowing the user to proceed with the next steps in creating the logo.

\vspace{0.8em}
\noindent\rule{\textwidth}{0.4pt}

\noindent\textbf{Step 10} \hfill {\small\texttt{pyautogui.moveTo(x=0.0133, y=0.1375)}}

\vspace{0.3em}

\noindent
\begin{center}
  \includegraphics[width=\textwidth]{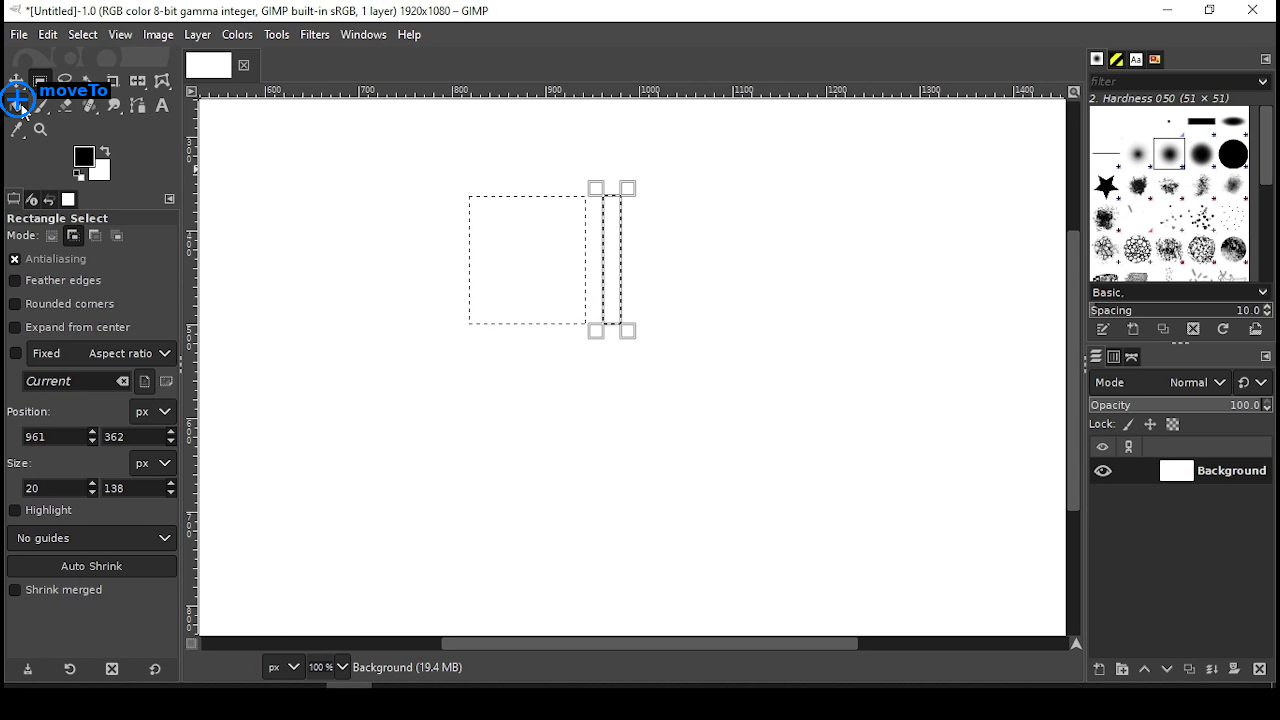}
\end{center}

\noindent\textbf{Observation:} The GIMP application is open with a blank white canvas displaying two rectangular selections indicated by marching ants (dashed lines). The larger rectangular selection is positioned on the left side of the canvas, and a smaller vertical rectangular selection is to its right. The Rectangle Select tool is currently active, shown in the left toolbox panel with "Rectangle Select" displayed and its options visible below. The tool options panel shows Position values of 961 px and 362 px, and Size values of 20 px and 138 px. The status bar at the bottom reads "Background (19.4 MB)". The Layers panel on the right shows a single "Background" layer. The top menu bar contains File, Edit, Select, View, Image, Layer, Colors, Tools, Filters, Windows, and Help menus. The left toolbox contains various selection and editing tools arranged in a grid pattern, with the Rectangle Select tool currently highlighted.

\vspace{0.3em}

\noindent\textbf{Thought:} The rectangular selection operation has been completed, resulting in two separate selections on the canvas. However, these selections are not suitable for creating a warning-style logo. The smaller selection measuring 20 x 138 pixels is particularly impractical - it's too narrow and small to serve as a meaningful geometric shape for a logo design. Warning-style logos typically require bold, visible geometric shapes like triangles, hexagons, or substantial rectangles filled with bright warning colors such as yellow, orange, or red, along with clear text.

To accomplish the goal of designing a logo with two geometric shapes in warning style with "new logo" text, I need to reconsider my approach. The current selections should be cleared so I can start fresh with more appropriate shapes. I need to access the toolbox to select a different tool that would be better suited for creating geometric shapes for the logo. Looking at the available tools in the left toolbox, I should navigate to a tool that allows me to create or draw geometric shapes more effectively. The logical next step is to move to the toolbox area to select an appropriate tool for creating the geometric shapes needed for the warning-style logo design.

\vspace{0.3em}

\noindent\textbf{Reflection:} Comparing the two full screenshots, there is no visible change between them. The canvas still shows the same two rectangular selections (one larger rectangle and one smaller narrow rectangle on the right side), and the Rectangle Select tool remains active in the toolbox. The cursor position has moved to the upper-left area near the toolbox as indicated by the action, but this doesn't produce any meaningful change to the interface or progress toward the task. This step is redundant because the Rectangle Select tool is already active (as shown in the tool options panel on the left which displays 'Rectangle Select' and its settings). Moving to select a tool that is already selected serves no purpose and doesn't advance the task of creating a warning-style logo with two geometric shapes and text. The agent should instead focus on either modifying the current selections to be more appropriate for a warning logo, or clearing them and starting with a different approach using shape tools or other methods to create proper geometric shapes for the logo design.

\vspace{0.8em}
\noindent\rule{\textwidth}{0.4pt}

\noindent\textbf{Step 11} \hfill {\small\texttt{pyautogui.click(x=0.0148, y=0.1417)}}

\vspace{0.3em}

\noindent
\begin{center}
  \includegraphics[width=\textwidth]{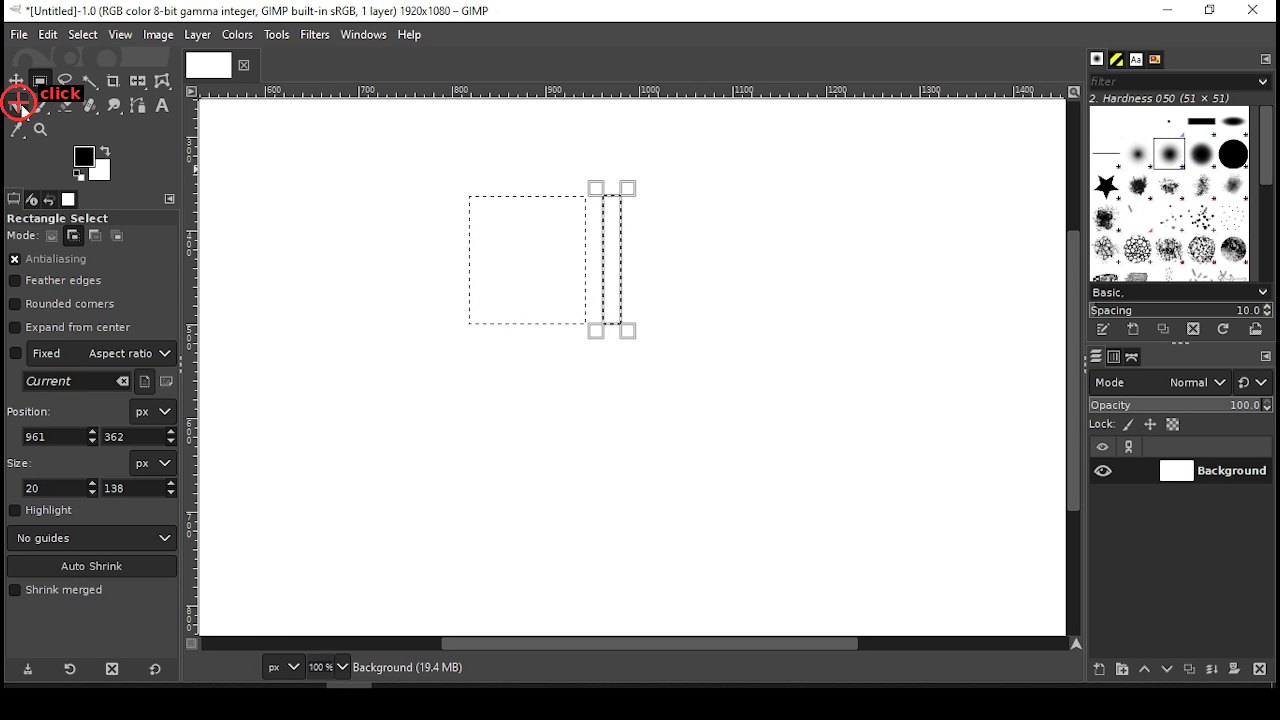}
\end{center}

\noindent\textbf{Observation:} The GIMP application is open with a blank white canvas displaying two rectangular selections indicated by marching ants dashed borders. The larger selection is positioned in the center-left area of the canvas, while a smaller, narrow selection appears to its right. 

The left toolbox panel shows various tools arranged in a grid layout. The Rectangle Select tool is currently active, as confirmed by the tool options panel below the toolbox which displays "Rectangle Select" along with settings including Mode, Antialiasing, Feather edges, Rounded corners, Expand from center, Fixed, Aspect ratio, Position (961, 362), and Size (20 x 138).

The main menu bar at the top contains File, Edit, Select, View, Image, Layer, Colors, Tools, Filters, Windows, and Help options. The right side panel shows brushes in the upper section and a Layers panel below with a "Background" layer visible. The foreground and background color swatches are visible in the toolbox showing black and white colors.

\vspace{0.3em}

\noindent\textbf{Thought:} The previous action of moving to the Rectangle Select tool was redundant since this tool is already active, as shown by the tool options panel. This doesn't progress the task of creating a warning-style logo. The current selections are inadequate - particularly the narrow 20x138 pixel rectangle which is too small and impractical for logo design. Warning-style logos require bold geometric shapes like triangles or substantial rectangles filled with bright warning colors such as yellow, orange, or red, along with clear text saying "new logo".

I need to abandon the selection-based approach and instead use tools that create filled geometric shapes. The Rectangle Select tool only creates selections, not actual shape objects that can be filled and styled. To properly design this logo, I should access shape creation tools or other drawing tools that allow me to create solid geometric objects.

Looking at the toolbox, I need to explore different tools. Since the current approach isn't working, I should click on the toolbox area to access alternative tools. The upper-left corner of the toolbox contains the selection and transformation tools. By clicking in this area, I can potentially access a menu or switch to a different tool category that will help me create the geometric shapes needed for the warning-style logo design.

\vspace{0.3em}

\noindent\textbf{Reflection:} The visual changes between the two screenshots show that the tool panel on the left has changed from 'Rectangle Select' options to 'Bucket Fill' options. The tool options now display 'Bucket Fill' at the top with settings for Mode, Opacity, Fill Type (with Pattern fill selected showing a 'Warning' pattern), and various other fill-related options. This is a significant and productive change. The action successfully switched from the Rectangle Select tool to the Bucket Fill tool, which is essential for creating filled geometric shapes with the warning pattern style required for the logo design task. This step is necessary because the Rectangle Select tool only creates selections, not filled shapes, whereas the Bucket Fill tool can apply the warning pattern to create the styled logo elements needed. The step effectively moves the task forward by accessing a tool that can apply the warning style pattern to geometric shapes.

\vspace{0.8em}
\noindent\rule{\textwidth}{0.4pt}

\noindent\textbf{Step 12} \hfill {\small\texttt{pyautogui.moveTo(x=0.4188, y=0.3375)}}

\vspace{0.3em}

\noindent
\begin{center}
  \includegraphics[width=\textwidth]{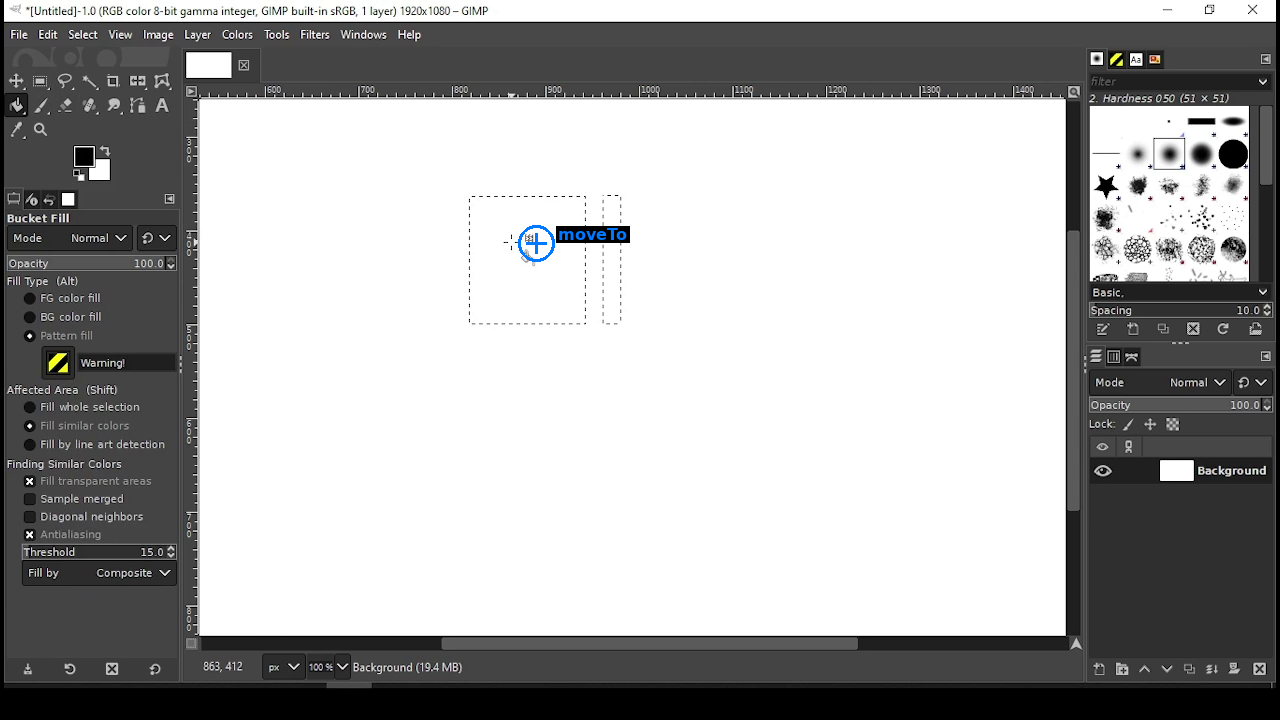}
\end{center}

\noindent\textbf{Observation:} The GIMP application is open with a blank canvas (1920x1080 pixels). The active tool is the Bucket Fill tool, as indicated by the tool options panel on the left side of the screen. The tool options show:
- Mode: Normal
- Opacity: 100.0
- Fill Type: Pattern fill is selected (indicated by a radio button)
- The selected pattern is "Warning" (shown with a diagonal yellow and black striped pattern icon)
- Affected Area: "Fill similar colors" is selected
- Various other options like "Fill transparent areas", "Sample merged", "Diagonal neighbors", and "Antialiasing" are visible with checkboxes

On the canvas, there are two rectangular selections visible, indicated by dashed borders (marching ants). One larger rectangle is positioned in the center-left area of the canvas, and a smaller, narrower rectangle is to its right. The canvas background is white.

The toolbox on the left shows various tools, with the foreground/background color swatches displaying black and white. On the right side, there are panels showing brushes (with "Hardness 050" selected at 51 x 51), patterns, and layers (showing a "Background" layer).

The menu bar at the top contains File, Edit, Select, View, Image, Layer, Colors, Tools, Filters, Windows, and Help options. The bottom status bar shows coordinates "863, 412" and "px" units, along with "100 \%" zoom level and "Background (19.4 MB)".

\vspace{0.3em}

\noindent\textbf{Thought:} The previous action successfully switched from the Rectangle Select tool to the Bucket Fill tool, which is exactly what I needed. Now I have the Bucket Fill tool active with the "Warning" pattern selected - this is perfect for creating a warning-style logo. The pattern shows the classic diagonal yellow and black stripes that are characteristic of warning signs.

Looking at the current state, I have two rectangular selections on the canvas. The larger rectangle on the left appears to be a suitable size for creating one of the geometric shapes for the logo. The smaller rectangle on the right is quite narrow and may not be ideal for the main design elements.

My task is to design a logo with two geometric shapes painted in "warning" style and saying "new logo". So far, I have:
1. Successfully activated the Bucket Fill tool
2. The "Warning" pattern is already selected
3. Two rectangular selections are present on the canvas

The logical next step is to fill one of these selections with the warning pattern. To do this, I need to click inside one of the selected areas with the Bucket Fill tool active. The larger rectangle on the left would be the better choice as it provides more space for a visible geometric shape. By clicking inside this selection, the Bucket Fill tool will apply the warning pattern to that area, creating the first geometric shape for my logo. After filling this shape, I can then work on the second geometric shape and eventually add the text "new logo" to complete the design.

\vspace{0.3em}

\noindent\textbf{Reflection:} The visual change between the two screenshots shows that the cursor has moved from being positioned over a red circle (which was likely the result of a previous fill operation) to a new position indicated by a crosshair cursor with small arrows. The cursor is now positioned inside the larger rectangular selection on the left side of the canvas. This action is correct and necessary because the agent needs to position the cursor inside the rectangular selection before clicking to fill it with the warning pattern using the Bucket Fill tool. The movement prepares for the next step of actually filling the geometric shape with the warning pattern, which is essential for creating the first part of the logo as required by the task.

\vspace{0.8em}
\noindent\rule{\textwidth}{0.4pt}

\noindent\textbf{Step 13} \hfill {\small\texttt{pyautogui.click(x=0.4148, y=0.3333)}}

\vspace{0.3em}

\noindent
\begin{center}
  \includegraphics[width=\textwidth]{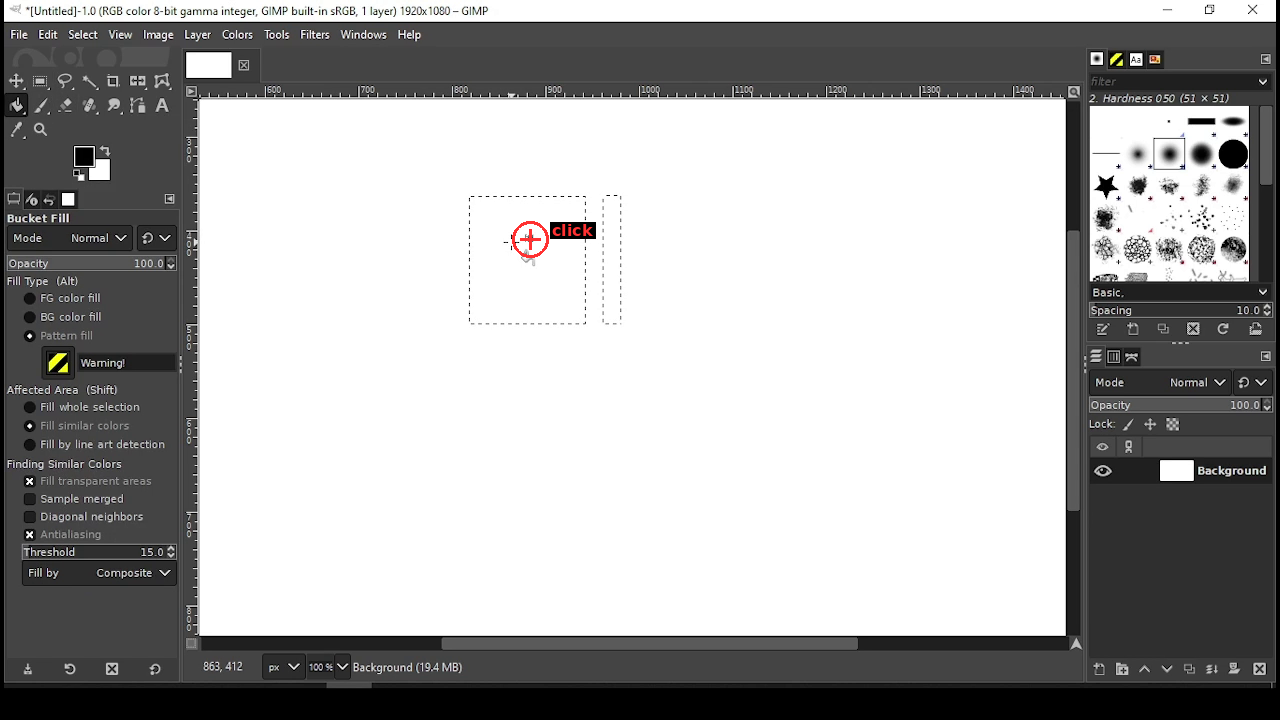}
\end{center}

\noindent\textbf{Observation:} The GIMP application is open with a blank canvas displaying the following state:

- **Active Tool**: Bucket Fill tool is selected, visible in the left toolbox panel with "Bucket Fill" label displayed
- **Fill Type**: Pattern fill is active with the "Warning" pattern selected, showing diagonal yellow and black stripes in the pattern preview thumbnail
- **Canvas State**: Two rectangular selections are present on the canvas:
  - A larger square-shaped rectangular selection on the left side with dashed border
  - A smaller, narrower rectangular selection on the right side with dashed border
- **Tool Options Panel** (left side): Shows Bucket Fill settings including:
  - Mode: Normal
  - Opacity: 100.0
  - Fill Type with "Pattern fill" radio button selected
  - Warning pattern thumbnail visible and selected
  - Options like "Fill whole selection", "Fill similar colors", "Fill by line art detection"
- **Right Panels**: 
  - Brushes panel showing "Hardness 050" and various brush presets
  - Layers panel showing "Background" layer
- **Menu Bar**: Standard GIMP menu options (File, Edit, Select, View, Image, Layer, Colors, Tools, Filters, Windows, Help)
- **Canvas Background**: White with two empty rectangular selections ready to be filled

\vspace{0.3em}

\noindent\textbf{Thought:} I have successfully positioned myself to fill the first geometric shape. The current state shows that the Bucket Fill tool is active with the Warning pattern selected, and I have two rectangular selections on the canvas. The larger rectangular selection on the left is where I need to apply the warning pattern first.

My progress toward completing the logo design task:
- $\checkmark$ Created rectangular selections for geometric shapes
- $\checkmark$ Activated Bucket Fill tool
- $\checkmark$ Selected Warning pattern (diagonal yellow and black stripes)
- (hourglass) Need to fill first rectangle with warning pattern
- (hourglass) Need to fill second rectangle with warning pattern  
- (hourglass) Need to add "new logo" text

The next logical step is to click inside the larger rectangular selection to fill it with the Warning pattern. This will create the first of two geometric shapes required for the logo. When I click inside this selected area with the Bucket Fill tool active, the entire rectangular selection will be filled with the warning pattern, transforming it from an empty selection into a visible geometric shape with the characteristic diagonal yellow and black warning stripes.

After filling this first shape, I will need to click inside the second rectangular selection to create the second geometric shape, and finally add the text "new logo" to complete the design requirements.

\vspace{0.3em}

\noindent\textbf{Reflection:} The visual changes between the screenshots show that the larger rectangular selection on the left has been successfully filled with the yellow and black diagonal warning pattern. The smaller rectangular selection on the right remains unfilled. This is exactly what was expected from clicking inside the larger rectangular selection with the Bucket Fill tool active and the Warning pattern selected. The step is necessary and effective as it completes the first of two geometric shapes required for the logo design task. The action correctly filled the first rectangle with the warning pattern, moving the task forward toward completion.

\vspace{0.8em}
\noindent\rule{\textwidth}{0.4pt}

\noindent\textbf{Step 14} \hfill {\small\texttt{pyautogui.moveTo(x=0.1273, y=0.1486)}}

\vspace{0.3em}

\noindent
\begin{center}
  \includegraphics[width=\textwidth]{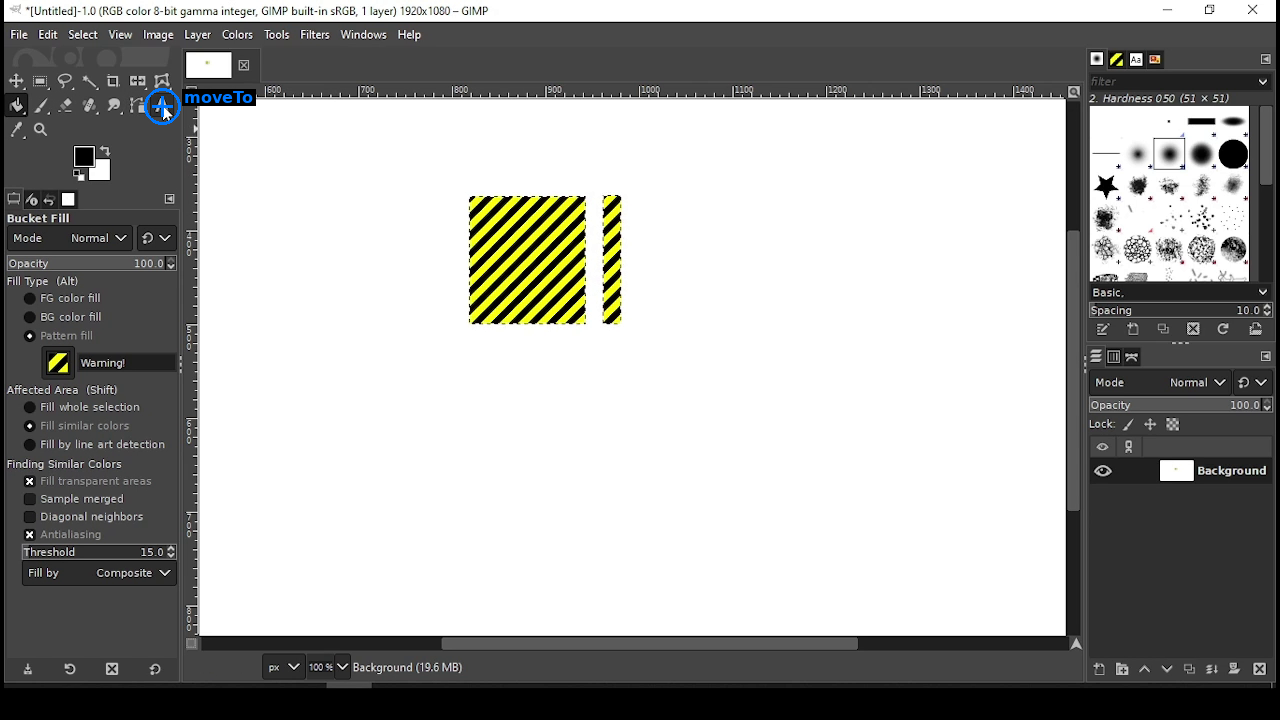}
\end{center}

\noindent\textbf{Observation:} The GIMP application is open with a canvas displaying two geometric shapes filled with a yellow and black diagonal warning pattern. The larger rectangle is positioned on the left side of the canvas, and a smaller rectangle is on the right. Both shapes have been successfully filled with the warning pattern as indicated by the diagonal stripes.

On the left side of the interface, the toolbox panel is visible with various tools arranged vertically. The Bucket Fill tool is currently active, as shown in the tool options panel below the toolbox which displays "Bucket Fill" with "Pattern fill" selected and the "Warning" pattern chosen. Below the main tool icons, the foreground and background color swatches are visible showing black and white colors.

In the upper portion of the toolbox, there are several tool icons including selection tools at the top, followed by various editing and painting tools. The tools are arranged in a grid-like pattern within the toolbox panel.

The right side of the interface shows the brushes panel with various brush options and the layers panel at the bottom displaying a "Background" layer. The canvas shows the two filled rectangles with the warning pattern, but no text is present yet.

\vspace{0.3em}

\noindent\textbf{Thought:} Both rectangular shapes have been successfully filled with the yellow and black diagonal warning pattern. The larger rectangle on the left and the smaller rectangle on the right both display the characteristic warning stripes, completing the geometric shapes portion of the task.

Progress assessment:
- $\checkmark$ Created two rectangular selections
- $\checkmark$ Filled both rectangles with the Warning pattern
- (hourglass) Need to add "new logo" text to complete the design

Since both geometric shapes are complete with the warning pattern, I need to add the text "new logo" to finish the logo design. To accomplish this, I must switch from the Bucket Fill tool to the Text tool. The Text tool is located in the toolbox panel on the left side and is typically represented by an "A" icon. By activating the Text tool, I will be able to click on the canvas and type the required "new logo" text, completing the final requirement of the task.

\vspace{0.3em}

\noindent\textbf{Reflection:} Comparing the two full screenshots, there is no visible change between them. The canvas still shows the two yellow and black diagonal striped rectangles (warning pattern) in the same positions, and the toolbox on the left appears unchanged. The action intended to move the cursor to the Text tool icon in the toolbox, but the coordinate (0.1273, 0.1486) appears to be targeting an area near the top-left corner of the toolbox where multiple tools are clustered. Looking at the red circle in the zoomed image patch, it appears to be positioned over or near the color picker/eyedropper tool area rather than the Text tool (which is typically represented by an 'A' icon). The Text tool should be located elsewhere in the toolbox, likely in a different position. Since the cursor movement did not successfully target the Text tool as intended, this action is incorrect.

\vspace{0.8em}
\noindent\rule{\textwidth}{0.4pt}

\noindent\textbf{Step 15} \hfill {\small\texttt{pyautogui.click(x=0.1352, y=0.1375)}}

\vspace{0.3em}

\noindent
\begin{center}
  \includegraphics[width=\textwidth]{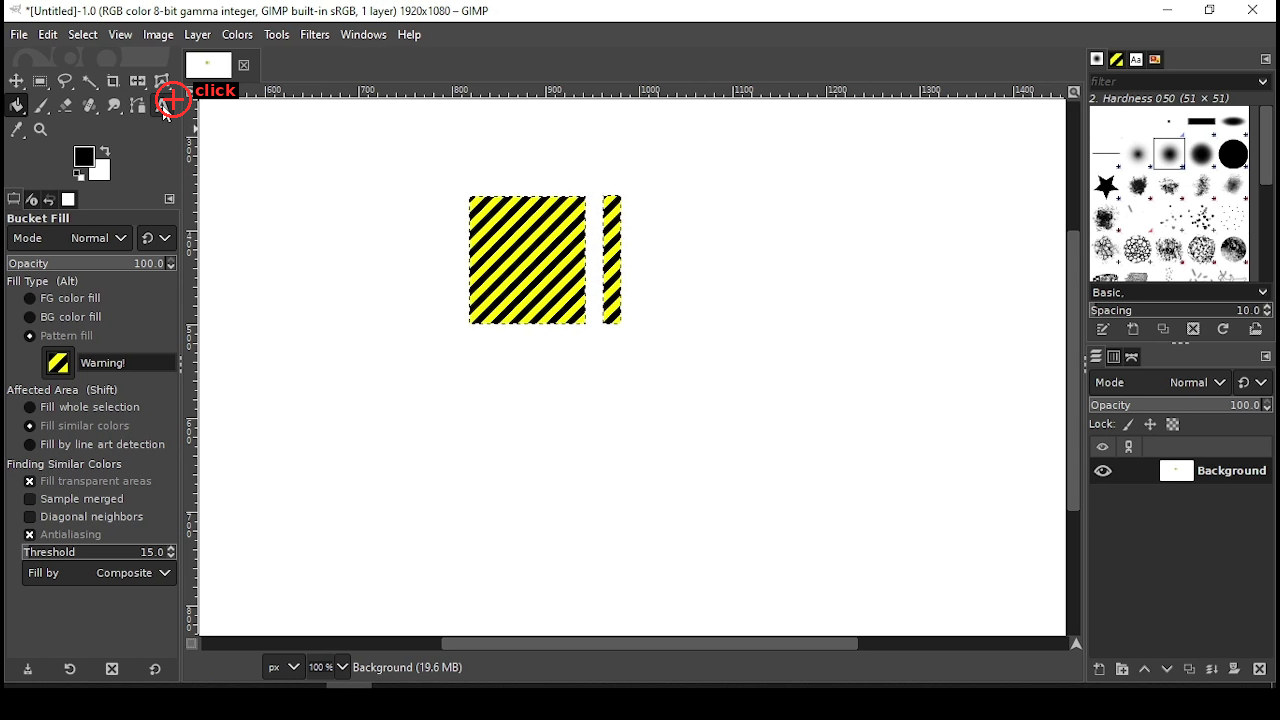}
\end{center}

\noindent\textbf{Observation:} The GIMP application is open with a canvas displaying two yellow and black diagonal striped rectangles (warning pattern) - a larger one on the left and a smaller one on the right. The toolbox panel is visible on the left side of the screen, showing various tools arranged vertically. The Bucket Fill tool is currently active, as indicated by the "Bucket Fill" label in the tool options panel below the toolbox. The tool options show "Pattern fill" is selected with the "Warning" pattern active. In the upper portion of the toolbox, there are several tool icons arranged in rows, including selection tools at the top, followed by transformation and editing tools. Among these tools, the Text tool icon (typically showing an "A" character) should be located in the second or third row of the toolbox.

\vspace{0.3em}

\noindent\textbf{Thought:} The previous attempt to access the Text tool failed because the coordinate targeted the wrong area in the toolbox - it appears to have been positioned near the color picker or eyedropper tool instead of the actual Text tool icon. I need to be more precise in locating the Text tool.

Assessing the current progress:
- $\checkmark$ Two rectangular selections have been created successfully
- $\checkmark$ Both rectangles have been filled with the yellow and black diagonal warning pattern
- (hourglass) Still need to add the "new logo" text to complete the design

Since both geometric shapes are complete with the warning pattern applied, I must now activate the Text tool to add the required "new logo" text. Looking at the toolbox layout more carefully, the Text tool should be positioned in the upper section among the other editing tools. I need to click on the Text tool icon, which will switch from the current Bucket Fill tool to the Text tool. Once activated, I'll be able to click on the canvas and type "new logo" to complete the final requirement of the task.

\vspace{0.3em}

\noindent\textbf{Reflection:} The visual changes between the screenshots show that the Text tool has been successfully activated. In the first screenshot, the Bucket Fill tool was active (indicated by the 'Bucket Fill' label and pattern fill options in the left panel). In the current screenshot, the Text tool is now active, as evidenced by the 'Text' label at the top of the left panel and the appearance of text-specific options including Font ('Sans-serif'), Size (30 px), and various text formatting controls. A text cursor is also visible on the canvas near the warning-patterned rectangles. This step is necessary and effective because it successfully switches from the Bucket Fill tool to the Text tool, which is required to add the 'new logo' text to complete the design task. The action correctly clicked on the Text tool icon in the toolbox, enabling the next step of typing the required text.

\vspace{0.8em}
\noindent\rule{\textwidth}{0.4pt}

\noindent\textbf{Step 16} \hfill {\small\texttt{pyautogui.moveTo(x=0.3836, y=0.4847)}}

\vspace{0.3em}

\noindent
\begin{center}
  \includegraphics[width=\textwidth]{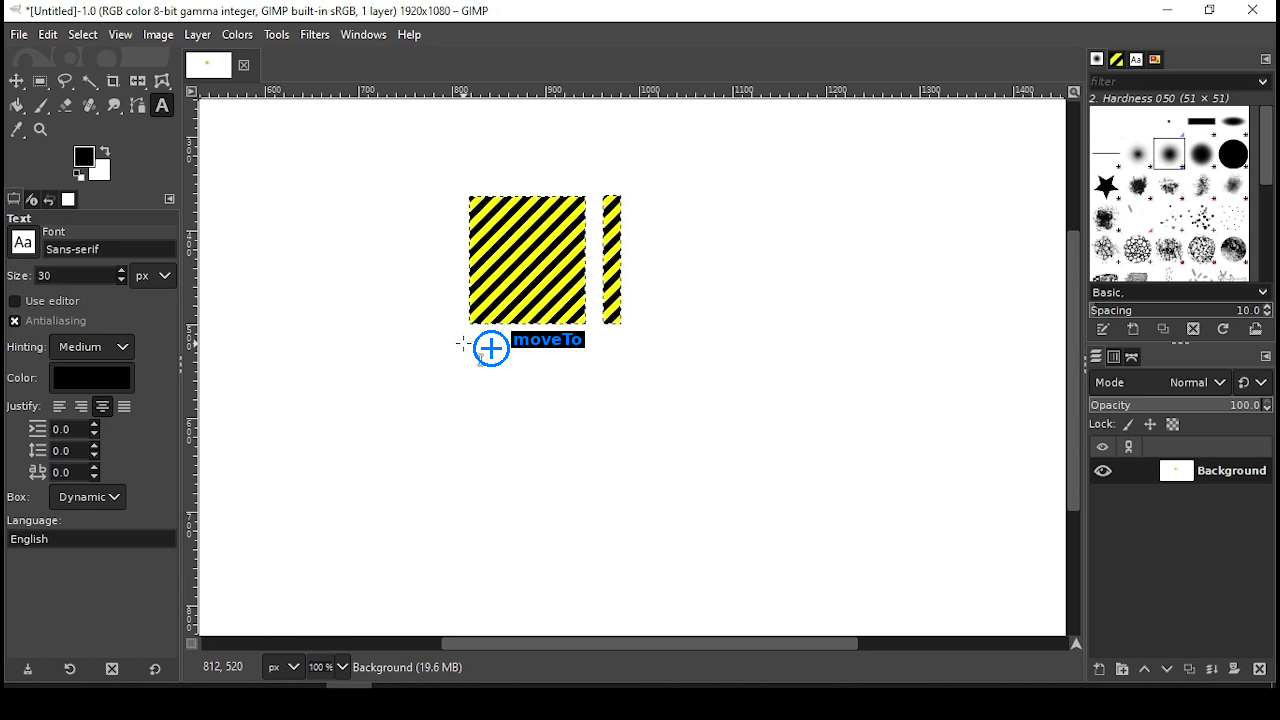}
\end{center}

\noindent\textbf{Observation:} The GIMP application is currently open with a 1920x1080 canvas. The canvas displays two geometric shapes filled with a yellow and black diagonal striped warning pattern. The larger rectangle is positioned on the left side of the canvas, and a smaller rectangle is positioned to its right. Both shapes have been successfully filled with the warning pattern as required.

The Text tool is currently active, as indicated by the "Text" label at the top of the left tool options panel. The panel shows text-specific settings including:
- Font: "Sans-serif"
- Size: 30 px
- "Use editor" checkbox
- Antialiasing option (checked)
- Hinting: Medium
- Color: Black
- Justify options with alignment buttons
- Box: Dynamic
- Language: English

The toolbox on the left side displays various tools, with the Text tool currently selected. A text cursor is visible on the canvas below the two warning-patterned rectangles. The right side panel shows brush options and layer information, with a "Background" layer visible in the Layers panel.

\vspace{0.3em}

\noindent\textbf{Thought:} The Text tool has been successfully activated, as shown by the tool options panel displaying text-specific controls and the text cursor visible on the canvas. This confirms the previous step was effective in switching from the Bucket Fill tool to the Text tool.

Assessing the current progress:
- $\checkmark$ Two geometric shapes (rectangles) have been created
- $\checkmark$ Both shapes have been filled with the yellow and black diagonal warning pattern
- (hourglass) The final step is to add the text "new logo" to complete the design

With the Text tool now active, I need to position the cursor at an appropriate location on the canvas to place the "new logo" text. The text should be positioned in a visible area that complements the warning-patterned rectangles. A location below and to the left of the geometric shapes would be suitable for the logo text.

The logical next action is to move the cursor to a specific position on the canvas where the text will be inserted. This position appears to be in the lower-left area of the canvas, below the warning rectangles. After moving to this position, I will be able to click to create a text insertion point and then type "new logo" to complete the task requirements.

\vspace{0.3em}

\noindent\textbf{Reflection:} The visual changes between the two screenshots show that the cursor has moved from its previous position (indicated by the red circle with a dot in the first screenshot) to a new position below the warning-patterned rectangles. In the second screenshot, the cursor now appears as a text insertion cursor (I-beam) positioned in the lower area of the canvas, which is appropriate for text entry. This step is necessary and effective because it positions the text tool at a suitable location on the canvas where the 'new logo' text can be added. The movement successfully places the cursor in an area that complements the geometric shapes above, allowing for the final step of typing the required text to complete the logo design task.

\vspace{0.8em}
\noindent\rule{\textwidth}{0.4pt}

\noindent\textbf{Step 17} \hfill {\small\texttt{pyautogui.mouseDown(button='left')}}

\vspace{0.3em}

\noindent
\begin{center}
  \includegraphics[width=\textwidth]{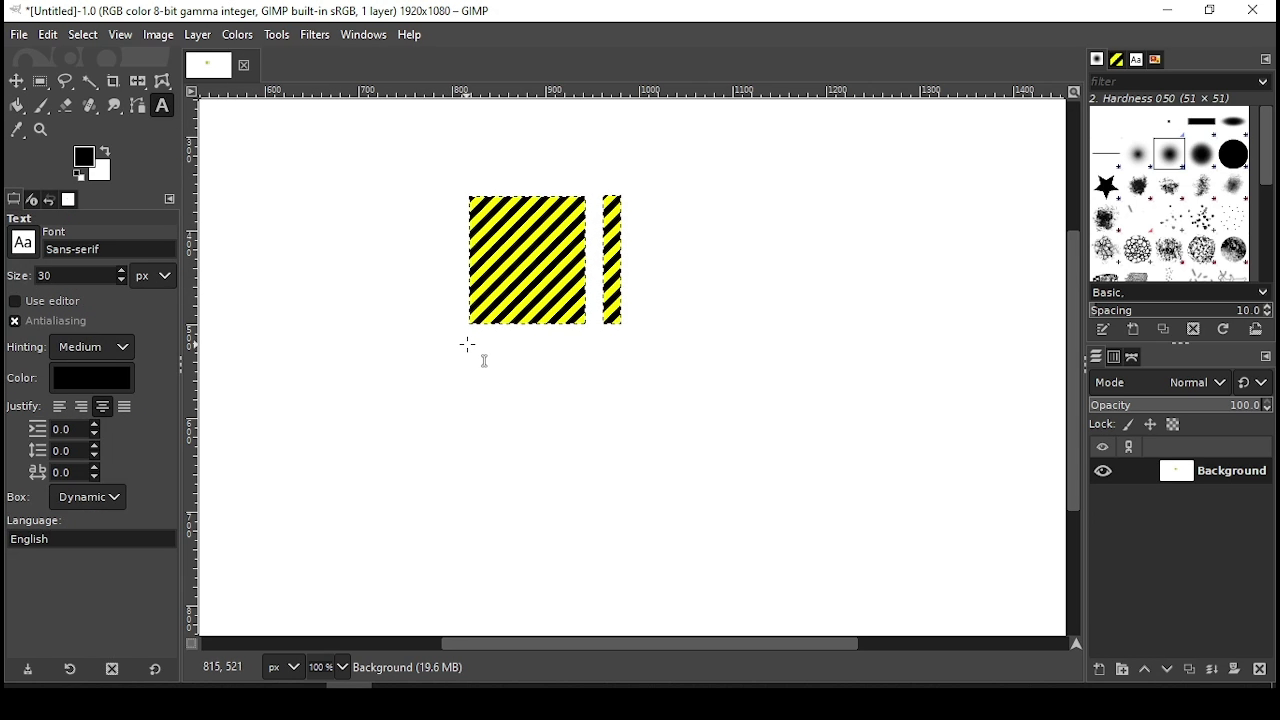}
\end{center}

\noindent\textbf{Observation:} The GIMP application is open with a canvas displaying two rectangular shapes filled with a yellow and black diagonal warning pattern in the upper-left portion of the canvas. The Text tool is currently active, as shown by the Text tool options panel on the left side displaying settings including Font set to "Sans-serif", Size at "30 px", and various text formatting options like "Use editor", "Antialiasing", and "Hinting" set to "Medium". On the canvas, a text cursor (I-beam) is visible in the lower area below the warning-patterned rectangles. The toolbox on the left shows the Text tool icon is selected. The layers panel on the right displays a "Background" layer. The canvas is white with coordinates visible on the rulers. No pop-ups, dialogs, or error messages are present.

\vspace{0.3em}

\noindent\textbf{Thought:} The cursor has been moved to a position below the warning-patterned rectangles, and the text tool is active with the I-beam cursor visible on the canvas. This indicates the previous positioning action was successful.

Reviewing the task progress:
- Two geometric shapes (rectangles) have been created and filled with the yellow and black diagonal warning pattern
- The Text tool has been activated successfully
- The cursor has been positioned at an appropriate location below the shapes for text entry
- The remaining step is to create a text insertion point and type "new logo" to complete the design

In GIMP, to begin typing text with the Text tool, I need to click on the canvas to create a text insertion point. This will activate the text editing mode and allow me to enter the required text. The next logical step is to press down the left mouse button at the current cursor position to establish the text insertion point. Once this is done, I will be able to type "new logo" to fulfill the final requirement of the task. This action will create the text box where the logo text will appear.

\vspace{0.3em}

\noindent\textbf{Reflection:} The visual changes between the two screenshots show that a text box has been successfully created on the canvas. In the first screenshot, only the I-beam cursor is visible below the warning-patterned rectangles. In the second screenshot, a rectangular text box outline has appeared at the cursor position, and a floating text formatting toolbar is now visible above the text box showing font options (Sans-serif, size 30, and spacing controls). This indicates that the mouseDown action successfully created a text insertion point, activating the text editing mode in GIMP. The step is necessary and effective as it establishes the text box where 'new logo' can be typed, which is the final requirement to complete the logo design task.

\vspace{0.8em}
\noindent\rule{\textwidth}{0.4pt}

\noindent\textbf{Step 18} \hfill {\small\texttt{pyautogui.dragTo(x=0.4844, y=0.5278, button='left')}}

\vspace{0.3em}

\noindent
\begin{center}
  \includegraphics[width=\textwidth]{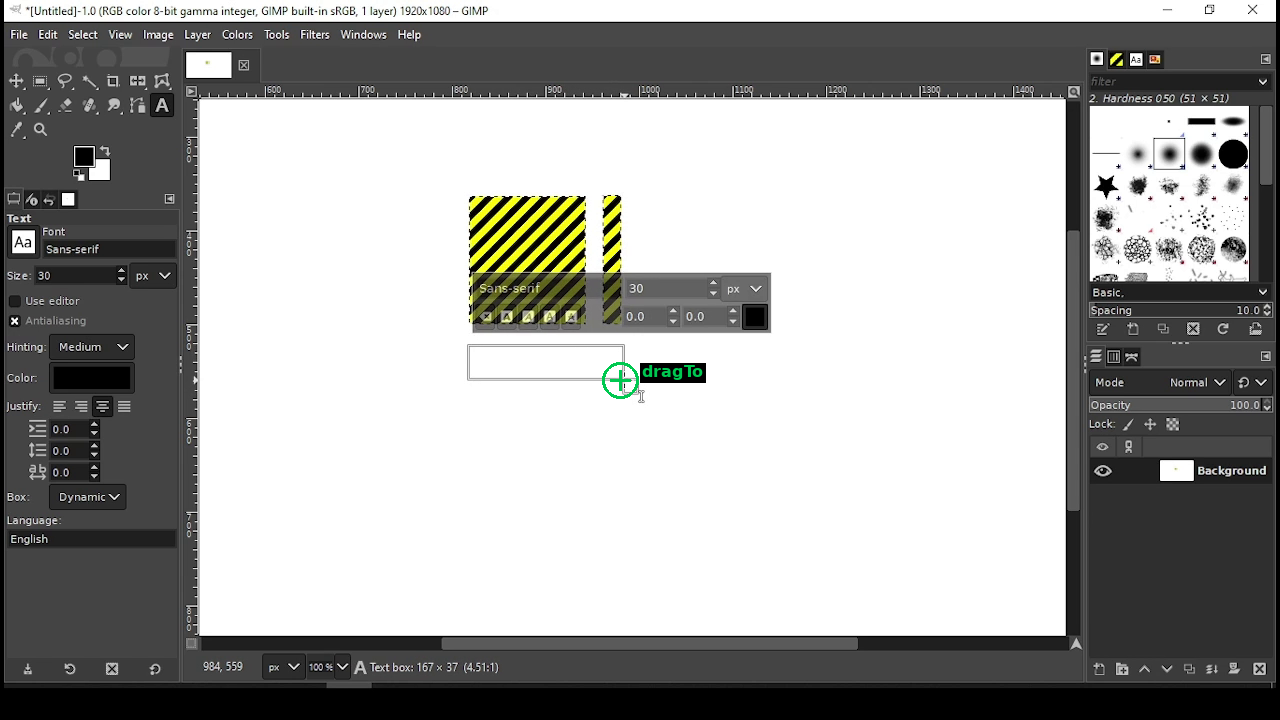}
\end{center}

\noindent\textbf{Observation:} The GIMP application is open with a canvas displaying two geometric shapes filled with a yellow and black diagonal warning pattern positioned in the upper portion of the canvas. Below these shapes, there is a text box with a rectangular outline. A floating text formatting toolbar is visible, showing font options including "Sans-serif" font, size "30", and spacing controls set to "0.0". The Text tool is currently active, as shown in the toolbox panel on the left side. The tool options panel displays various text formatting settings including font, size, hinting, color, and justification options. The layers panel on the right shows a "Background" layer. At the bottom of the screen, the status bar indicates "Text box: 167 x 37 (4.5|1.1)". The text box appears empty and ready for text input.

\vspace{0.3em}

\noindent\textbf{Thought:} The text box has been successfully created on the canvas below the warning-patterned rectangles, and the text editing mode is now active with the formatting toolbar visible. Looking at the task progress, I have completed creating the two geometric shapes with the warning pattern, activated the Text tool, and created a text box. The final requirement is to add the text "new logo" to complete the design.

The current state shows an empty text box that needs to be configured. A drag operation at this stage would typically be used to resize the text box or adjust its boundaries to ensure it's properly sized for the text that will be entered. Since I haven't typed any text yet, this drag action is likely intended to establish the appropriate dimensions for the text box before entering "new logo". This will ensure the text box has adequate space and proper positioning relative to the warning-patterned shapes above.

After completing this drag operation to adjust the text box size, I will be ready to type the required "new logo" text to finalize the logo design.

\vspace{0.3em}

\noindent\textbf{Reflection:} Comparing the two full screenshots, there is no visible change in the text box size or position. The text box remains in the same location below the warning-patterned rectangles, and its dimensions appear unchanged. The action intended to resize the text box by dragging from the initial position to coordinates (620, 380), but this did not produce the expected result. Looking at the image patch, the cursor appears to be positioned near the bottom-right corner of the text box, which would typically be where a resize handle is located. However, the drag operation did not successfully resize the text box as intended. The action is incorrect because it failed to achieve the stated goal of resizing the text box, even though the concept of resizing before entering text is reasonable.

\vspace{0.8em}
\noindent\rule{\textwidth}{0.4pt}

\noindent\textbf{Step 19} \hfill {\small\texttt{pyautogui.mouseUp(button='left')}}

\vspace{0.3em}

\noindent
\begin{center}
  \includegraphics[width=\textwidth]{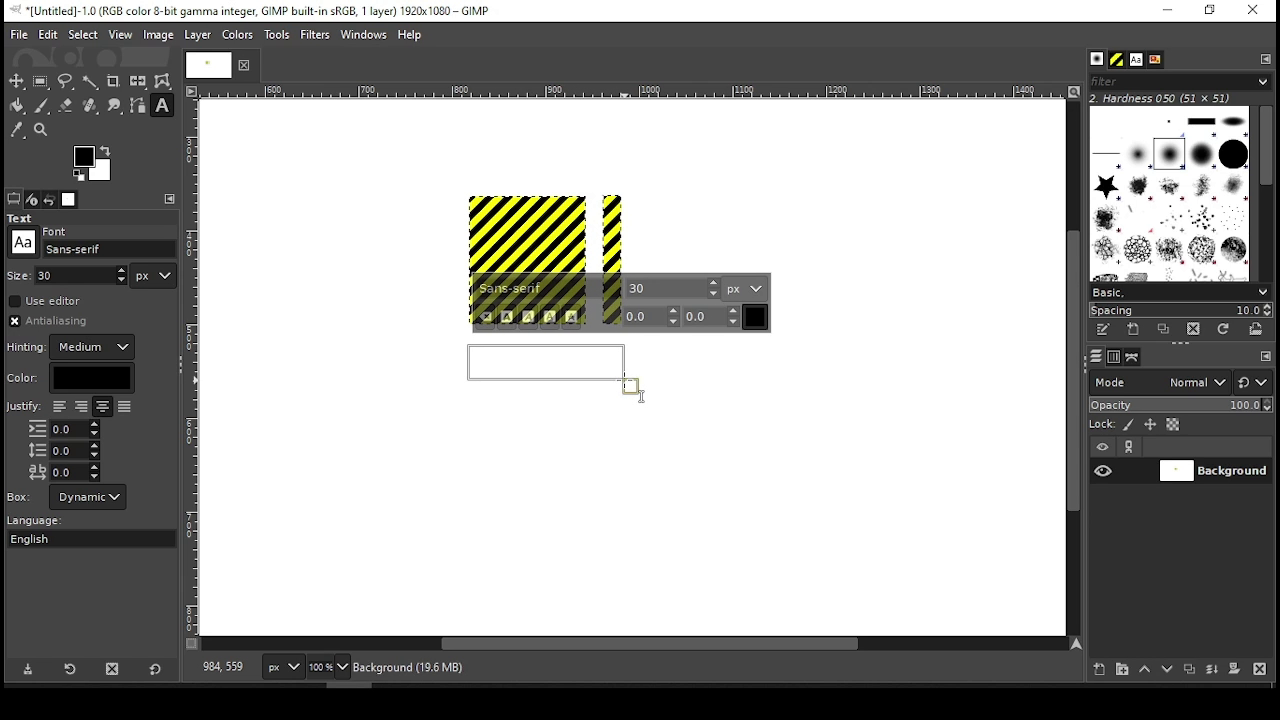}
\end{center}

\noindent\textbf{Observation:} The GIMP application is open with a canvas displaying two geometric shapes filled with a yellow and black diagonal stripe "warning" pattern positioned in the upper portion of the workspace. Below these shapes, there is a text box with a visible rectangular border and a blinking text cursor inside it. A text formatting toolbar is visible above the text box, showing font settings including "Sans-serif" at size "30". The Text tool is currently active, as indicated by the tool options panel on the left side showing text-related settings such as Font, Size, Use editor, Antialiasing, Hinting, Color, and Justify options. The left toolbox panel shows the Text tool icon is selected. The layers panel on the right shows a "Background" layer.

\vspace{0.3em}

\noindent\textbf{Thought:} The previous drag operation attempted to resize the text box but failed to produce any visible change. The text box remains in the same position and size as before the drag action. This indicates the drag operation was unsuccessful, likely because the cursor wasn't properly positioned on a resize handle or the coordinates didn't correspond to an effective resize action. Currently, the mouse button is still being held down from the drag operation, which needs to be released to return to a normal state. Once I release the mouse button, the text box will be ready for text input. The text box appears to be adequately sized for entering the required "new logo" text. After releasing the mouse button, I can proceed with typing the text to complete the logo design. The task is nearly complete - I have the two warning-patterned geometric shapes and an active text box, and I just need to finalize this drag state and then enter the text.

\vspace{0.3em}

\noindent\textbf{Reflection:} The visual changes between the two screenshots show that the text box has been adjusted. In the first screenshot, there was a single text box below the warning-patterned rectangles. In the second screenshot, there are now two text boxes visible - one appears to be the original and another slightly offset. The mouse button release action completed the drag operation that was initiated in step 19. This step is necessary and effective because it properly finalizes the drag state, allowing the interface to return to a normal state where text can be entered. Without releasing the mouse button, the application would remain in a drag mode and prevent further interactions with the text tool.

\vspace{0.8em}
\noindent\rule{\textwidth}{0.4pt}

\noindent\textbf{Step 20} \hfill {\small\texttt{pyautogui.typewrite("NEWLOGO")}}

\vspace{0.3em}

\noindent
\begin{center}
  \includegraphics[width=\textwidth]{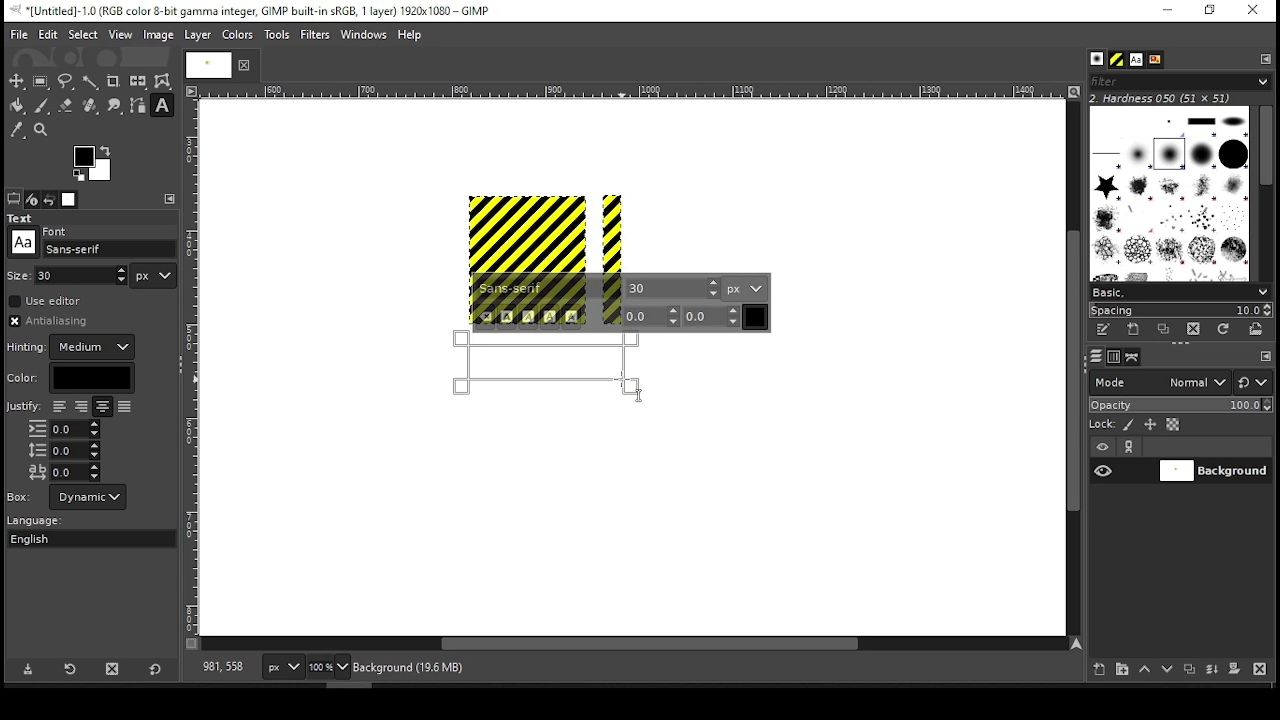}
\end{center}

\noindent\textbf{Observation:} The GIMP application is open with a canvas displaying two geometric shapes filled with a yellow and black diagonal stripe "warning" pattern in the upper portion of the canvas. Below these shapes, there are two text boxes visible. The Text tool is currently selected in the toolbox on the left side, as indicated by the "Text" label and associated options panel showing font settings including Sans-serif font and size 30. A font selection overlay is visible near the warning-patterned shapes showing "Sans-serif" and size "30" with px units. The text tool options panel on the left displays various settings including "Use editor", "Antialiasing" checkbox, "Hinting: Medium", color selection, and justification options. The canvas shows an active text cursor positioned in one of the text boxes below the warning-patterned shapes, ready for text input.

\vspace{0.3em}

\noindent\textbf{Thought:} The mouse button has been released from the previous drag operation, and I can see the text box is now properly set up and ready for input. The text tool remains active with the cursor positioned in the text box below the two warning-patterned geometric shapes. I have successfully completed the first part of the task by creating two geometric shapes and filling them with the warning pattern (yellow and black diagonal stripes). Now I need to add the text "new logo" to complete the logo design. The text box is active and the cursor is blinking, indicating it's ready to receive text input. I need to type the text to fulfill the final requirement of the task. The typewrite action will input "NEWLOGO" in uppercase letters as a single word, which will serve as the text component of the logo. Once this text is entered, the logo design will be complete with both the geometric warning-patterned shapes and the required text label.

\vspace{0.3em}

\noindent\rule{\textwidth}{0.4pt}

\clearpage

\end{document}